%% file: g3m_arxiv.tex
\documentclass[a4paper,11pt]{article}
\usepackage{authblk}
\usepackage[british]{babel}
\usepackage[mono=false]{libertine}
\usepackage{setspace}
\usepackage[utf8]{inputenc} % allow utf-8 input
\usepackage[T1]{fontenc}    % use 8-bit T1 fonts
\usepackage{url}            % simple URL typesetting
\usepackage{microtype}
\usepackage{graphicx}
\usepackage{amssymb}
\usepackage{subcaption}
\usepackage{booktabs} % for professional tables
\usepackage{amsfonts}       % blackboard math symbols
\usepackage{amsmath,amssymb,amsthm,mathrsfs,bbm,bm}
\usepackage{physics}
\usepackage{color}
\usepackage[usenames, dvipsnames]{xcolor}
\usepackage{enumitem}% http://ctan.org/pkg/enumitem
\usepackage{algorithm}
\usepackage{algorithmic}
\usepackage{balance}
\usepackage{microtype}
\usepackage{times}
\usepackage{wrapfig}
\usepackage{pythonhighlight}

\usepackage[top=2.5cm, bottom=2.5cm, left=2cm, right=2cm]{geometry}
\usepackage[pdfpagemode=UseNone,bookmarks=false, bookmarksopen=false,colorlinks=true,urlcolor=RoyalBlue,citecolor=YellowOrange,citebordercolor=RoyalBlue,linkcolor=RoyalBlue]{hyperref}

\newtheorem{theorem}{Theorem}
\newtheorem{proposition}{Proposition}
\newtheorem{lemma}{Lemma}

\newtheorem{definition}{Definition}
\newtheorem{conjecture}{Conjecture}

\input{settings_custom}

\numberwithin{equation}{section}

\begin{document}
%\newpage

\title{Learning curves of generic features maps for \\ realistic datasets with a teacher-student model}

\author[1]{Bruno Loureiro\thanks{bruno.loureiro@epfl.ch}}
\author[2]{C\'edric Gerbelot\thanks{cedric.gerbelot@ens.fr}}
\author[3]{Hugo Cui}
\author[4]{Sebastian Goldt}
\author[1]{\\Florent Krzakala}
\author[2]{Marc M\'ezard}
\author[3]{Lenka Zdeborov\'a}

\affil[1]{IdePHICS laboratory, \'Ecole F\'ed\'erale Polytechnique de Lausanne (EPFL), Switzerland}
\affil[2]{Laboratoire de Physique de l’Ecole Normale Supérieure, Université
  PSL, CNRS, \protect\\ Sorbonne Université, Université Paris-Diderot, Paris, France}
\affil[3]{SPOC laboratory, Ecole F\'ed\'erale Polytechnique de Lausanne (EPFL), Switzerland}
\affil[4]{International School of Advanced Studies (SISSA), Trieste, Italy}
\date{}

\maketitle

\begin{abstract}
 Teacher-student models provide a framework in which the typical-case
  performance of high-dimensional supervised learning can be described in
  closed form. 
  The assumptions of Gaussian i.i.d.~input data underlying the canonical teacher-student model may, however, be perceived as too restrictive to capture the behaviour of realistic data sets. 
  In this paper, we introduce a Gaussian covariate 
  generalisation of the model where the teacher and student can act on
  different spaces, generated with fixed, but generic feature maps. 
  While still solvable in a closed form, this generalization is able to capture the learning curves for a broad range of realistic data sets, thus redeeming the potential of the teacher-student framework.  
  Our contribution is then two-fold: First, we prove a rigorous formula for
  the asymptotic training loss and generalisation error. 
  Second, we present a number of situations
  where the learning curve of the model captures the one of a \emph{realistic
    data set} learned with kernel regression and classification, with
  out-of-the-box feature maps such as random projections or scattering
  transforms, or with pre-learned ones - such as the features learned by
  training multi-layer neural networks. 
  We discuss both the power and the limitations of the framework.
\end{abstract}

\newpage
\makeatletter
\def\l@subsubsection#1#2{}
\makeatother

{
  \hypersetup{linkcolor=black}
\begin{spacing}{0.9}
  \tableofcontents
  \end{spacing}
}

\newpage
%%%%%%%%%%%%%%%%%%%%%%%%%%%%%%%%%%%%%%%%%%%%
\section{Introduction}
\label{sec:main:intro}
\input{sections/introduction.tex}

%%%%%%%%%%%%%%%%%%%%%%%%%%%%%%%%%%%%%%%%%%%%
\section{Main technical results}
\label{sec:technical}
\input{sections/technical.tex}
%%%%%%%%%%%%%%%%%%%%%%%%%%%%%%%%%%%%%%%%%%%%
\section{Applications of the Gaussian model}
\label{sec:applications}
\input{sections/applications}
%%%%%%%%%%%%%%%%%%%%%%%%%%%%%%%%%%%%%%%%%%%%
\section*{Acknowledgements}
\input{sections/acknowledgements}
%%%%%%%%%%%%%%%%%%%%%%%%%%%%%%%%%%%%%%%%%%%%

\bibliographystyle{unsrt}
\bibliography{references}

\newpage
%%%%%%%%%%%
\appendix
%%%%%%%%%%
%%%%%%%%%%%%%%%%%%%%%%%%%%%%%%%%%%%%%%%%%%%%%
\section{Main result from the replica method}
\label{sec:app:replicas}
%%%%%%%%%%%%%%%%%%%%%%%%%%%%%%%%%%%%%%%%%%%%%
\input{./sections/appendix/replicas}

%%%%%%%%%%%%%%%%%%%%%%%%%%%%%%%%%%%%%%%%%%%%%%%%%
%\newpage
%\section{Power law scaling for kernel regression}
%\label{scaling}
%\input{sections/appendix/scaling_Ridge}
%%%%%%%%%%%%%%%%%%%%%%%%%%%%%%%%%%%%%%%%%%%%%%%%%%
\newpage
\section{Rigorous proof of the main result}
\label{main-proof}
\input{./sections/appendix/proof}
%%%%%%%%%%%%%%%%%%%%%%%%%%%%%%%%%%%%%%%%%%%%%%%%%%
\newpage
\section{Equivalence replica-Gordon}
\label{app-mapping}
\input{./sections/appendix/l2_matching}
%%%%%%%%%%%%%%%%%%%%%%%%%%%%%%%%%%%%%%%%%%%%%%%%%%
\newpage
\section{Details on the simulations}
\label{details}
\input{./sections/appendix/details}
%%%%%%%%%%%%%%%%%%%%%%%%%%%%%%%%%%%%%%%%%%%%%%%%%%
\newpage
\section{Ridge regression with linear teachers}
\label{RidgeUniversality}
\input{./sections/appendix/Universality}
%%%%%%%%%%%%%%%%%%%%%%%%%%%%%%%%%%%%%%%%%%%%%%%%%%
\newpage
%%%%%%%%%%%%%%%%%%%%%%%%%%%%%%%%%%%%%%%%%%%%%%%%%%

\end{document}

%% file: settings_custom.tex
\def\new{\text{new}}
\def\tot{\text{tot}}
\def\sign{\text{sign}}

\newcommand{\extr}{\textrm{\textbf{extr}}}

\def\dd{\text{d}}

\def\sign{\text{sign}}
\def\erf{\text{erf}}

\def\mat#1{\text{#1}}
\renewcommand{\vec}[1]{\boldsymbol{#1}}

\DeclareMathOperator*{\argmin}{arg\,min}

\def\prox{\mbox{prox}}

\newcommand\samples{n}
\newcommand\sdim{d}
\newcommand\tdim{p}
\newcommand\ddim{D}
\newcommand\loss{g}
\newcommand\ar{\gamma}

\newcommand\reg{r}

%% file: sections/introduction.tex
Teacher-student models are a popular framework to study the high-dimensional asymptotic
performance of learning problems with synthetic data, and have been the subject
of intense investigations spanning three decades~\cite{seung1992statistical,
  watkin1993statistical, engel2001statistical, donoho2009message, el2013robust,
  zdeborova2016statistical, donoho2016high}.  In the wake of understanding the
limitations of classical statistical learning approaches~\cite{zhang2016understanding,
  belkin2019reconciling, belkin2020two}, this direction is witnessing a renewal of
interest~\cite{mei2019generalization, hastie2019surprises, belkin2020two,
  candes2020phase, aubin2020generalization, salehi2020performance}. However, this framework is often assuming the input data to be Gaussian i.i.d., which is arguably too simplistic to be able to capture properties of realistic data.   In this paper, we redeem this line of work by defining a Gaussian covariate model  where the teacher and student act on different Gaussian correlated spaces with arbitrary covariance. We derive a rigorous asymptotic solution  of this model generalizing the formulas found in the above mentioned classical works.

We then put forward a theory, supported by universality arguments and numerical experiments, that this model captures learning curves, i.e.~the dependence of the training and test errors on the number of samples, for a generic class of feature maps applied to realistic datasets. These maps can be deterministic, random, or even learnt from the data. This analysis thus gives a unified framework to describe the learning curves of, for example, kernel regression and classification, the analysis of feature maps -- random
projections~\cite{rahimi2008random}, neural tangent kernels~\cite{jacot2018neural},
scattering transforms~\cite{andreux2020kymatio} -- as well as the analysis of transfer
learning performance on data generated by generative adversarial
networks~\cite{goodfellow2014generative}. 
We also discuss limits of applicability of our results, by showing concrete situations where the learning curves of the Gaussian covariate model differ from the actual ones.

\paragraph{Model definition ---} 
The Gaussian covariate teacher-student model is defined via two vectors $\vec{u} \in \mathbb{R}^{p}$ and $\vec{v} \in \mathbb{R}^{d}$, with correlation matrices $\Psi\in\mathbb{R}^{\tdim\times\tdim},\Omega\in\mathbb{R}^{\sdim\times\sdim}$ and $\Phi\in\mathbb{R}^{\tdim\times\sdim}$, from which we draw $\samples$ independent samples:
\begin{align}
    \begin{bmatrix}
    \vec{u}^{\mu} \\
    \vec{v}^{\mu}
    \end{bmatrix} \in \mathbb{R}^{p+d} \underset{\text{i.i.d.}}{\sim} \mathcal{N}\left(0, \begin{bmatrix}\Psi & \Phi \\ \Phi^{\top} & \Omega \end{bmatrix}\right), && \mu=1,\cdots, n.
    \label{def:GM3}
\end{align}
The {\it labels}
$y^{\mu}$ are generated by a \textbf{teacher} function that is only using the vectors
$\vec{u}^{\mu}$:
\begin{equation}
  \label{teacher}
  {y}^\mu = f_{0}\left(\frac{1}{\sqrt{p}} \vec{\theta}_{0}^{\top}\vec{u}^\mu \right)\, ,
\end{equation}
where $f_{0}:\mathbb{R} \to \mathbb{R}$ is a function that may include
randomness such as, for instance, an additive Gaussian noise, and
$\vec{\theta}_{0}\in\mathbb{R}^{\tdim}$ is a vector of teacher-weights with finite norm which can
be either random or deterministic. Learning is performed by the
\textbf{student} with weights $\vec{w}$ via empirical risk minimization that has
access only to the features $\vec{v}^{\mu}$:
\begin{equation}
  \label{eq:student}
  \hat{\vec{w}} = \argmin_{\vec{w}\in \mathbb{R}^{d}}
  \left[\sum\limits_{\mu=1}^{n}
    g\left(\frac{\vec{w}^{\top}\vec{v}^{\mu}}{\sqrt{d}}, y^{\mu}\right)+r(\vec{w})
  \right]\, ,
\end{equation}
where $r$ and $g$ are proper, convex, lower-semicontinuous functions of
${\vec{w}}\in\mathbb{R}^{\sdim}$ (e.g.~$g$ can be a logistic or a square loss and $r$
a $\ell_{p}$ $(p\!=\!1,2)$ regularization). The key quantities we want to compute in this model are the \emph{averaged training and generalisation errors} for the estimator $\vec{w}$,
\begin{align}
  \label{main-eq:errors}
    \mathcal{E}_{\rm train.}(\vec{w}) \equiv 
    \frac{1}{n}\sum\limits_{\mu=1}^{n} g\left(\frac{\vec{w}^{\top}\vec{v}^{\mu}}{\sqrt{d}}, y^{\mu}\right) \qq{and} \mathcal{E}_{\rm gen.}(\vec{w}) 
    \equiv    \mathbb{E}\left[\hat{g}\left(\hat{f}\left(\frac{\vec{v}_{\rm{new}}^{\top}\vec{w}}{\sqrt{d}}\right), f_{0}\left(\frac{\vec{u}_{\rm{new}}^{\top}\vec{\theta}_{0}}{\sqrt{p}}\right)\right)\right].
\end{align}
where $g$ is the loss function in eq.~\eqref{eq:student}, $\hat{f}$ is a prediction function (e.g.~$\hat{f} = \rm{sign}$ for a classification task), $\hat{g}$ is a performance measure (e.g.~$\hat{g}(\hat{y}, y) = (\hat{y}-y)^2$ for regression or $\hat{g}(\hat{y}, y) = \mathbb{P}(\hat{y}\neq y)$ for classification) and $(\vec{u}_{\new}, \vec{v}_{\new})$ is a fresh sample from the joint distribution of $\vec{u}$ and $\vec{v}$.

Our two \textbf{main technical contributions} are: 
\vspace{-2mm}
\begin{itemize}[wide = 1pt,noitemsep]
\item[(C1)] In Theorems \ref{thm:corollary:factorised} \& \ref{main-th-main}, we give a rigorous closed-form characterisation of the properties of the estimator $\hat{\vec{w}}$ for the Gaussian covariate model~\eqref{def:GM3}, and the corresponding training and generalisation errors in the high-dimensional limit. We prove our result using Gaussian comparison inequalities~\cite{gordon1985some};
\item[(C2)] We show how the same expression can be obtained using the replica method from statistical physics~\cite{mezard1987spin}. This is of additional interest given the wide range of applications of the replica approach in machine learning and computer science~\cite{mezard2009information}. In particular, this allows to put on a rigorous basis many results previously derived with the replica method.
\end{itemize}

\begin{figure}
\centering
\begin{subfigure}{.45\textwidth}
  \centering
  \vspace{-0.5cm}
  \includegraphics[width=1.0\linewidth]{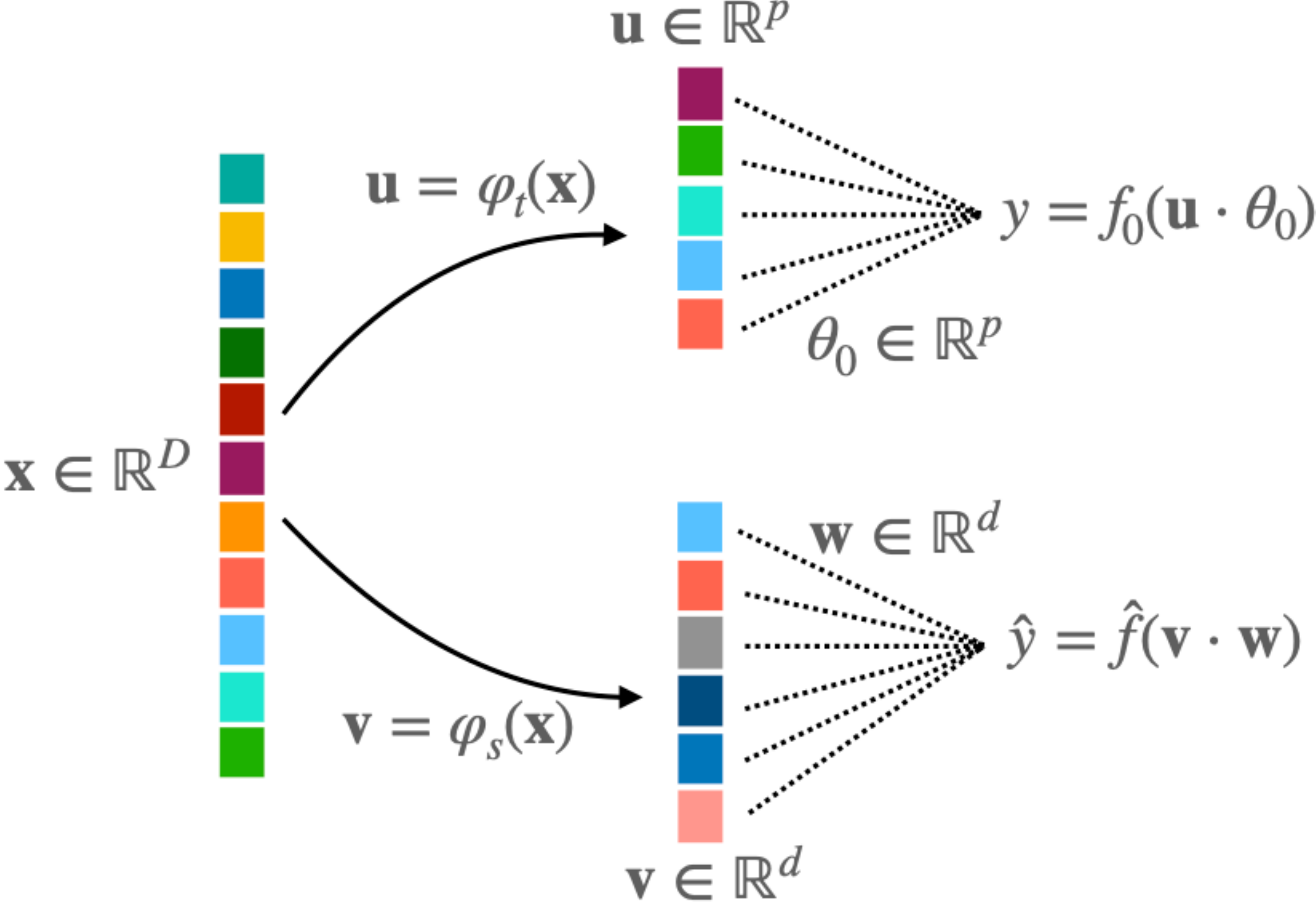}
\end{subfigure}%
\begin{subfigure}{.45\textwidth}
  \centering
  \includegraphics[width=0.81\linewidth]{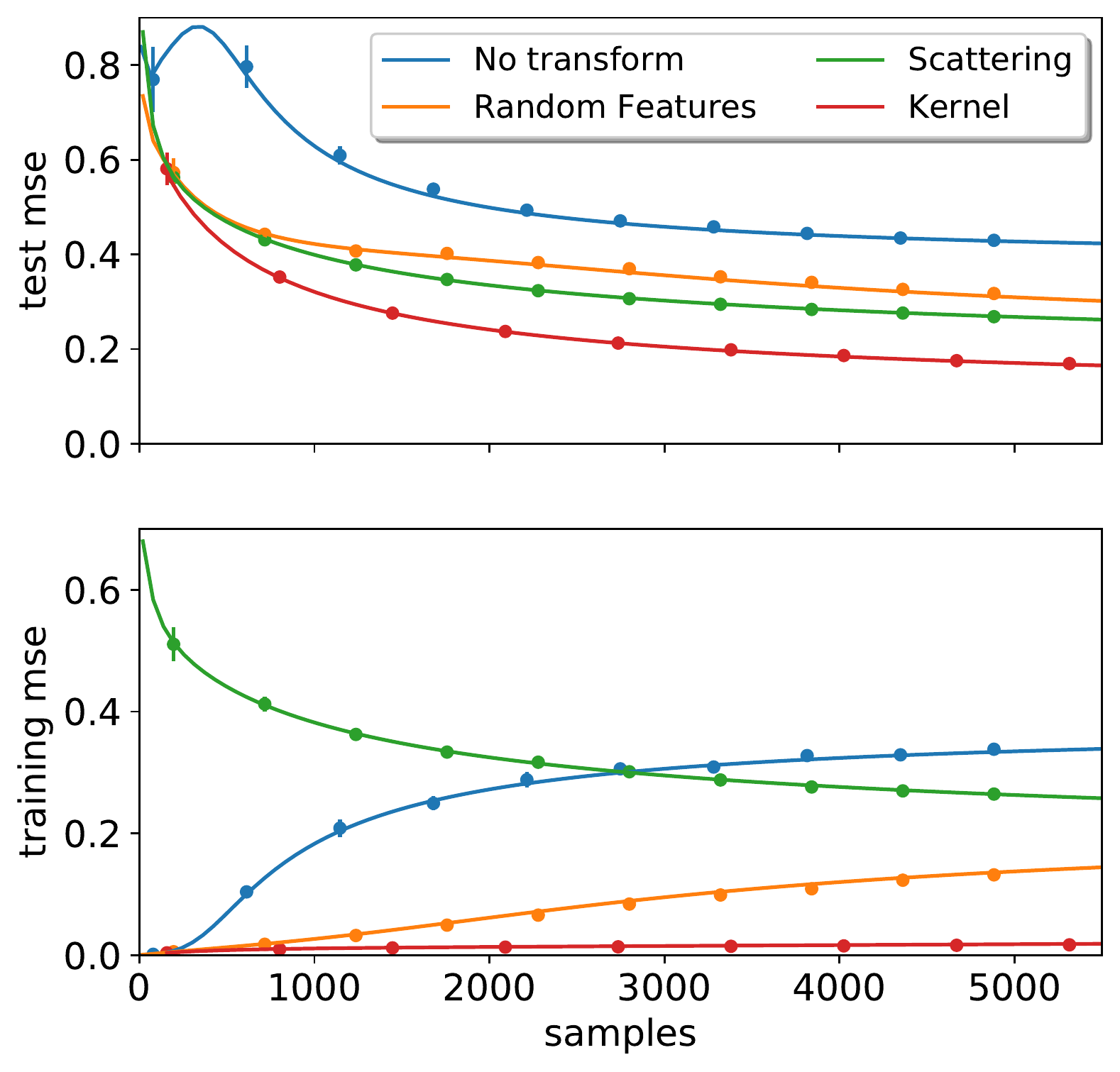}
\end{subfigure}
\vspace{-3mm}
\caption{\label{fig:setup} {\bf Left:} 
      Given a data set $\{\vec{x}^{\mu}\}_{\mu=1}^{n}$, teacher
    $\vec{u}= \vec{\varphi}_{t}(\vec{x})$ and student maps
    $\vec{v} = \vec{\varphi}_{t}(\vec{x})$, we assume $[\vec{u},\vec{v}]$ to be jointly
    Gaussian random variables and apply the results of the Gaussian covariate model
    (\ref{def:GM3}). {\bf Right:}  Illustration on real data, here ridge regression on even vs odd MNIST digits, with regularisation $\lambda\!=\!10^{-2}$. Full line is theory, points are simulations. We show the performance with no feature map (blue), random feature map with $\sigma=\rm{erf}$ \& Gaussian projection (orange), the scattering transform with parameters $J=3, L=8$ \cite{andreux2020kymatio} (green), and of the limiting kernel of the random map \cite{williams96} (red).   The covariance $\Omega$ is empirically estimated from the full data set, while the other quantities appearing in the Theorem~\ref{thm:corollary:factorised} are expressed directly as a function of the labels, see Section \ref{sec:regression:real}. Simulations are averaged over $10$ independent runs.
}
\vspace{-6mm}
\end{figure}

\vspace{-4mm}
\paragraph{Towards realistic data ---} 

In the second part of our paper, we argue that the above Gaussian covariate model~\eqref{def:GM3} is generic enough to capture the learning behaviour of a broad range of realistic data.  
Let $\{\vec{x}^{\mu}\}_{\mu=1}^{n}$ denote a data set with
$n$ independent samples on $\mathcal{X}\subset \mathbb{R}^{\ddim}$. Based on this input,
the \textbf{features} $\vec{u}, \vec{v}$ are given by (potentially) elaborated
transformations of $\vec{x}$, i.e.
\begin{equation}
  \label{eq:features}
  \vec{u} = \vec{\varphi}_{t}(\vec{x}) \in \mathbb{R}^p \qq{and} \vec{v} = \vec{\varphi}_{s}(\vec{x})\in\mathbb{R}^d
\end{equation}
for given centred feature maps $\vec{\varphi}_{t}:\mathcal{X}\to\mathbb{R}^{\tdim}$ and
$\vec{\varphi}_{s}:\mathcal{X}\to\mathbb{R}^{\sdim}$, see Fig.~\ref{fig:setup}. Uncentered features can be taken into account by shifting the covariances, but we focus on the centred case to lighten notation.

The Gaussian covariate model~\eqref{def:GM3} is exact in the
case where $\vec{x}$ are Gaussian variables and the feature maps
$(\vec{\varphi}_{s}, \vec{\varphi}_{s})$ preserve the Gaussianity, for example linear
features. In particular, this is the case for $\vec{u}\!=\!\vec{v}\!=\!\vec{x}$, which
is the widely-studied vanilla teacher-student model~\cite{gardner1989three}. The
interest of the model~\eqref{def:GM3} is that it also captures a range of cases in which
the feature maps $\vec{\varphi}_{t}$ and
$\vec{\varphi}_{s}$ are deterministic, or even
learnt from the data. 
The covariance matrices $\Psi$, $\Phi$, and $\Omega$ then represent different aspects of the data-generative process and learning model. The student (\ref{eq:student}) then corresponds to the last
layer of the learning model. These observation can be distilled into the following conjecture:
\begin{conjecture}(Gaussian equivalent model)
  \label{conjecture_one}
  For a wide class of data distributions $\{\vec{x}^{\mu}\}_{\mu=1}^{n}$, and features
  maps $\vec{u} =\vec{\varphi}_{t}(\vec{x}), \vec{v} =\vec{\varphi}_{s}(\vec{x})$, the
  generalisation and training errors of estimator (\ref{eq:student}) are asymptotically captured
  by the equivalent Gaussian model (\ref{def:GM3}), where $[\vec{u},\vec{v}]$ are
  jointly Gaussian variables, and thus by the closed-form expressions of
  Theorem~\ref{thm:corollary:factorised}.
\end{conjecture}

The second part of our \textbf{main contributions} are:
\vspace{-2mm}
\begin{itemize}[wide = 0pt,noitemsep]
\item[(C3)] In Sec.~\ref{sec:main:gan} we show that the theoretical predictions from (C1) captures the learning
  curves in non-trivial cases, e.g.~when input data are generated using a trained
  generative adversarial network, while extracting both the feature maps from a neural
  network trained on real data.
\item[(C4)] In Sec.~\ref{sec:regression:real}, we show empirically that for ridge regression the asymptotic formula of Theorem~\ref{thm:corollary:factorised} can
  be applied {\it directly} to real data sets, even though the Gaussian hypothesis is
  not satisfied. This universality-like property is a consequence of Theorem~\ref{thm:main:universality} and is illustrated in Fig.~\ref{fig:setup} (right) where the real learning curve of several features maps learning the odd-versus-even digit task on MNIST is compared to the theoretical prediction. 
\end{itemize}
\vspace{-4mm}

\paragraph{Related work ---} 
Rigorous results for teacher-student models: The Gaussian covariate
model (\ref{def:GM3}) contains the vanilla teacher-student model as a special
case where one takes $\vec{u}$ and $\vec{v}$ {\it identical}, with unique
covariance matrix $\Omega$. This special case has been extensively studied in
the statistical physics community using the heuristic replica
method~\cite{gardner1989three, opper1996statistical, seung1992statistical,
  watkin1993statistical, engel2001statistical}. Many recent rigorous results for
such models can be rederived as a special case of our formula,
e.g.~refs.~\cite{mei2019generalization, hastie2019surprises, ghorbani2020neural,
  belkin2020two, candes2020phase, thrampoulidis2018precise,
  montanari2019generalization, aubin2020generalization, salehi2020performance,
  celentano2020lasso}. Numerous of these results are based on the same proof
technique as we employed here: the Gordon's Gaussian min-max inequalities
\cite{gordon1985some, stojnic2013framework, oymak2013squared}. The asymptotic
analysis of kernel ridge regression \cite{bordelon2020}, of margin-based
classification \cite{huang2020large} also follow from our theorem.
See also Appendix \ref{sec:app:connection} for the details on these connections.
Other examples include models of the double descent phenomenon
\cite{mitra2019understanding}.  Closer to our work is the recent work of
\cite{dhifallah2020precise} on the random feature model. For ridge regression,
there are also precise predictions thanks to random matrix theory
\cite{dobriban2018high, hastie2019surprises, wu2020optimal, liao2020random, liu2020kernel, Bartlett30063,jacot2020kernel}. A related set of results was obtained
in~\cite{gerbelot2020colt} for orthogonal random matrix models. The main
technical novelty of our proof is the handling of a generic loss and
regularisation, not only ridge, representing convex empirical risk minimization,
for both classification and regression, with the generic correlation structure
of the model~(\ref{def:GM3}).

{Gaussian equivalence:} A similar Gaussian conjecture has been
discussed in a series of recent works, and some authors proved partial results
in this direction \cite{hastie2019surprises, mei2019generalization,
  montanari2019generalization, gerace2020generalisation, goldt2020modelling,
  goldt2020gaussian, dhifallah2020precise,
  hu2020universality}. Ref.~\cite{goldt2020gaussian} analyses a special case of
the Gaussian model (corresponding to $\vec{\varphi}_{t} = \rm{id}$ here), and
proves a Gaussian equivalence theorem (GET) for feature maps $\vec{\varphi}_{s}$
given by single-layer neural networks with fixed weights. They also show that
for Gaussian data $\vec{x}\sim\mathcal{N}(\vec{0}, \mat{I}_{\ddim})$, feature
maps of the form $\vec{v} = \sigma(\mat{W} \vec{x})$ (with some technical
restriction on the weights) led to the jointly-Gaussian property for the two
scalars $(\vec{v}\cdot \vec{w},\vec{u}\cdot \vec{\theta}_0)$ for \emph{almost}
any vector $\vec{w}$. However, their stringent assumptions on random teacher
weights limited the scope of applications to unrealistic label
models.  A related line of work discussed
similar universality through the lens of random matrix
theory~\cite{el2010spectrum, pennington2017nonlinear,
  louart2018concentration}. In particular, Seddik et al.~\cite{seddik2020random}
showed that, in our notations, vectors $[\vec{u},\vec{v}]$ obtained from
Gaussian inputs $\vec{x}\sim\mathcal{N}(\vec{0},\mat{I}_{\ddim})$ with Lipschitz
feature maps satisfy a concentration property. In this case, again, one can expect the two scalars
$\left(\vec{v}\cdot \vec{w},\vec{u}\cdot \vec{\theta}_0\right)$ to be jointly
Gaussian with high-probability on $\vec{w}$. 
Remarkably, in the case of random feature maps, \cite{hu2020universality} could go beyond this central-limit-like behavior and established the universality of the Gaussian covariate model (\ref{def:GM3}) for the actual learned weights $\hat{\vec{w}}$.

%% file: sections/technical.tex
Our main technical result is a closed-form expression for the asymptotic training and generalisation errors~\eqref{main-eq:errors} of the Gaussian covariate model introduced above. 
We start by presenting our result in the most relevant setting for the applications of interest in Section \ref{sec:applications}, which is the case of the $\ell_2$ regularization. Next, we briefly present our result in larger generality, which includes non-asymptotic results for non-separable losses and regularizations.

We start by defining key quantities that we will use to characterize the estimator $\hat{\vec{w}}$.
Let $\Omega = \mat{S}^{\top}\text{diag}(\omega_{i})\mat{S}$ be the spectral decomposition of $\Omega$. Let:
\begin{align}
    \rho \equiv \frac{1}{d}\vec{\theta}_{0}^{\top}\Psi\vec{\theta}_{0} \in\mathbb{R}, && \bar{\vec{\theta}} \equiv \frac{\mat{S}\Phi^{\top}\vec{\theta}_{0}}{\sqrt{\rho}} \in\mathbb{R}^{d} \label{eq:rho}
\end{align}
\noindent and define the joint empirical density $\hat{\mu}_{d}$ between $(\omega_{i}, \bar{\theta}_{i})$:
\begin{align}
\label{eq:main:specdensity}
    \hat{\mu}_{d}(\omega, \bar{\theta}) \equiv \frac{1}{d}\sum\limits_{i=1}^{d}\delta(\omega-\omega_{i})\delta(\bar{\theta} - \bar{\theta}_{i}).
\end{align}
Note that $\Phi^{\top}\vec{\theta}_{0}$ is the projection of the teacher weights on the student space, and therefore $\bar{\vec{\theta}}$ is the rotated projection on the basis of the student covariance, rescaled by the teacher variance. Together with the student eigenvalues $\omega_i$, these are relevant statistics of the model, encoded here in the joint distribution $\hat{\mu}_{d}$. 
\paragraph{Assumptions ---} Consider the \emph{high-dimensional} limit in which the number of samples $\samples$ and the dimensions $\tdim,\sdim$ go to infinity with fixed ratios:
\begin{equation}
\alpha \equiv \frac \samples\sdim\, ,  {\rm ~~and~~} \gamma \equiv \frac \tdim\sdim\, .
\end{equation}
Assume that the covariance matrices $\Psi,\Omega$ are positive-definite and that the Schur complement of the block covariance in equation (\ref{def:GM3}) is positive semi-definite.
Additionally, the spectral distributions of the matrices $\Phi, \Psi$ and $\Omega$ converge to distributions such that the limiting joint distribution $\mu$ is well-defined, and their maximum singular values are bounded with high probability as $n,p,d \to \infty$. Finally, regularity assumptions are made on the loss and regularization functions mainly to ensure feasibility of the minimization problem. We assume that the cost function $\reg+\loss$ is coercive, i.e.
$ \lim_{\norm{\vec{w}}_{2} \to +\infty} (\reg+\loss)(\vec{w}) = +\infty$
and that the following scaling condition holds : for all $\samples,\sdim \in \mathbb{N}, \vec{z} \in \mathbb{R}^{\samples}$ and any constant $c>0$, there exist a finite, positive constant $C$, such that, for any standard normal random vectors $\vec{h} \in \mathbb{R}^{d}$ and $\vec{g} \in \mathbb{R}^{n}$:
\begin{align}
\norm{\vec{z}}_{2} \leqslant c\sqrt{n} \implies \sup_{\vec{x} \in \partial g(\vec{z})} \norm{\vec{x}}_{2} \leqslant C\sqrt{n}, && \frac{1}{\sdim}\mathbb{E}\left[\reg(\vec{h})\right] <+\infty, && \frac{1}{n}\mathbb{E}\left[g(\vec{g})\right] < +\infty
 \end{align}
 The relevance of these assumptions in a supervised machine learning context is discussed in Appendix \ref{main-assumptions}. We are now in a position to state our result.
\begin{theorem}(Closed-form asymptotics for $\ell_2$ regularization)
\label{thm:corollary:factorised}
In the asymptotic limit defined above, the training and generalisation errors \eqref{main-eq:errors} of the estimator $\hat{\vec{w}}\in\mathbb{R}^{d}$ solving the empirical risk minimisation problem in eq.~\eqref{eq:student} with $\ell_2$ regularization $r(\vec{w}) = \frac{\lambda}{2}||\vec{w}||^2_{2}$ verify:
\begin{align}
    &\mathcal{E}_{\rm{train.}}(\hat{\vec{w}}) \xrightarrow[d \to \infty]{P} \mathbb{E}_{s,h\sim\mathcal{N}(0,1)}\left[g\left(\mbox{prox}_{V^{\star} g(.,f_{0}(\sqrt{\rho}s))}\left(\frac{m^{\star}}{\sqrt{\rho}}s+\sqrt{q^{\star}-\frac{{m^{\star}}^{2}}{\rho}} h\right),f_{0}(\sqrt{\rho}s)\right)\right] \notag\\ 
    &\mathcal{E}_{\rm{gen.}}(\hat{\vec{w}})  \xrightarrow[d \to \infty]{P}\mathbb{E}_{(\nu,\lambda)}\left[\hat{g}\left(\hat{f}(\lambda), f_{0}(\nu)\right)\right]
\end{align}
\noindent where $\mbox{prox}$ stands for the proximal operator defined as
\begin{equation}
    \mbox{prox}_{Vg(.,y)}(x) = \argmin_{z} \{g(z,y)+\frac{1}{2V}(x-z)^{2}\}
\end{equation}
and where $(\nu, \lambda)$ are jointly Gaussian scalar variables:
\begin{align}
    (\nu,\lambda)\sim\mathcal{N}\left(0, \begin{bmatrix}\rho & m^{\star} \\ m^{\star} & q^{\star}\end{bmatrix}\right),
\end{align}
\noindent and the overlap parameters $(V^{\star}, q^{\star}, m^{\star})$ are prescribed by the unique fixed point of the following set of self-consistent equations:
\begin{align}
\label{eq:main:sp}
\!\!\!\!
\begin{cases}
   V = \mathbb{E}_{(\omega, \bar{\theta})\sim \mu}\left[\frac{\omega}{\lambda+\hat{V}\omega}\right] \\
    m =\frac{\hat{m}}{\sqrt{\gamma}}\mathbb{E}_{(\omega, \bar{\theta})\sim \mu}\left[\frac{\bar{\theta}^2}{\lambda+\hat{V}\omega}\right]\\
   q = \mathbb{E}_{(\omega, \bar{\theta})\sim \mu}\left[\frac{\hat{m}^2\bar{\theta}^2\omega+\hat{q}\omega^2}{\left(\lambda+\hat{V}\omega\right)^2}\right] \\
\end{cases} \!\!\!\!\!\!, \!\! && \!\! \begin{cases}
   \hat{V} = \frac{\alpha}{V}(1-\mathbb{E}_{s,h\sim\mathcal{N}(0,1)}[f'_{g}(V,m,q)]) \\
    \hat{m}=\frac{1}{\sqrt{\rho \gamma}}\frac{\alpha}{V}\mathbb{E}_{s,h\sim\mathcal{N}(0,1)}\left[sf_{g}(V,m,q)\!-
    \! \frac{m}{\sqrt{\rho}}f'_{g}(V,m,q)\right] \\
  \hat{q} =  \frac{\alpha}{V^2}\mathbb{E}_{s,h\sim\mathcal{N}(0,1)}\!\left[\!\left(\frac{m}{\sqrt{\rho}}s+\!\sqrt{q\!-\!\frac{m^{2}}{\rho}} h\!-\!f_{g}(V,m,q)\right)^{2}\!\right]
  \end{cases}
\end{align}
\noindent where we defined the scalar random functions $f_{g}(V,m,q)=\mbox{prox}_{V g(.,f_{0}(\sqrt{\rho}s))}(\rho^{-1/2}ms+\sqrt{q-\rho^{-1} m^{2}} h)$ and $f'_{g}(V,m,h)=\mbox{prox}'_{V g(.,f_{0}(\sqrt{\rho}s))}(\rho^{-1/2} ms+\sqrt{q-\rho^{-1}m^{2}}h)$ as the first derivative of the proximal operator.
\end{theorem}
\emph{Proof}: This result is a consequence of Theorem \ref{main-th-main}, whose proof can be found in appendix \ref{main-proof}. 

The parameters of the model $(\vec{\theta}_{0}, \Omega, \Phi, \Psi)$ only appear trough~$\rho$, eq.~(\ref{eq:rho}), and the asymptotic limit $\mu$ of the joint distribution eq.~(\ref{eq:main:specdensity}) and $(f_{0},\hat{f}, \loss, \lambda)$. One can easily iterate the above equations to find their fixed point, and extract $(q^{*},m^{*})$ which appear in the expressions for the training and generalisation errors $(\mathcal{E}^{\star}_{\rm{train}}, \mathcal{E}^{\star}_{\rm{gen}})$, see eq.~\eqref{main-eq:errors}. Note that $(q^{\star}, m^{\star})$ have an intuitive interpretation in terms of the estimator $\hat{\vec{w}}\in\mathbb{R}^{d}$:
\begin{align}
    q^{\star} \equiv \frac{1}{d}\hat{\vec{w}}^{\top}\Omega\hat{\vec{w}}, && m^{\star} \equiv \frac{1}{\sqrt{dp}}\vec{\theta}_{0}^{\top}\Phi\hat{\vec{w}}
\end{align}
Or in words: $m^{\star}$ is the correlation between the estimator projected in the teacher space, while $q^{\star}$ is the reweighted norm of the estimator by the covariance $\Omega$. The parameter $V^{*}$ also has a concrete interpretation : it parametrizes the deformation that must be applied to a Gaussian field specified by the solution of the fixed point equations to obtain the asymptotic behaviour of $\hat{\mathbf{z}}$. It prescribes the degree of non-linearity given to the linear output by the chosen loss function. This is coherent with the robust regression viewpoint, where one introduces non-square losses to deal with the potential non-linearity of the generative model. $\hat{V}^{*}$ plays a similar role for the estimator $\hat{\mathbf{w}}$ through the proximal operator of the regularisation, see Theorem \ref{Train_gen} and \ref{main-th} in the Appendix. Two cases are of particular relevance for the experiments that follow. The first is the case of \emph{ridge regression}, in which $f_{0}(x) = \hat{f}(x)$ and both the loss $\loss$ and the performance measure $\hat{g}$ are taken to be the \emph{mean-squared error} $\text{mse}(y, \hat{y}) = \frac{1}{2}(y-\hat{y})^2$, and the asymptotic errors are given by the simple closed-form expression:
\begin{align}
    \mathcal{E}^{\star}_{\rm{gen}} = \rho + q^{\star} - 2m^{\star}, && \!\!\! \mathcal{E}^{\star}_{\rm{train}} = \frac{\mathcal{E}^{\star}_{\rm{gen}}}{(1+V^\star)^2}\, ,
    \label{eq:mse}
\end{align}
The second case of interest is the one of a binary classification task, for which $f_{0}(x) = \hat{f}(x) = \sign(x)$, and we choose the performance measure to be the \emph{classification error} $\hat{g}(y,\hat{y}) = \mathbb{P}(y\neq\hat{y})$. In the same notation as before, the asymptotic generalisation error in this case reads:
\begin{align}
    \label{eq:claloss}
    \mathcal{E}^{\star}_{\rm{gen}} = \frac{1}{\pi}\cos^{-1}\left(\frac{m^{\star}}{\sqrt{\rho q^{\star}}}\right),
\end{align}
\noindent while the training error $\mathcal{E}^{\star}_{\rm{train}}$ depends on the choice of $\loss$ - which we will take to be the logistic loss $\loss(y,x) = \log\left(1+e^{-xy}\right)$ in all of the binary classification experiments. 

As mentioned above, this paper includes stronger technical results including finite size corrections and precise characterization of the distribution of the estimator $\hat{\vec{w}}$, for generic, non-separable loss and regularization $g$ and $r$. This type of distributional statement is encountered for special cases of the model in related works such as \cite{miolane2018distribution,celentano2020lasso,montanari2019generalization}. Define $\mathcal{V} \in \mathbb{R}^{n\times d}$ as the matrix of concatenated samples used by the student. Informally, in high-dimension, the estimator $\hat{\vec{w}}$ and $\hat{\vec{z}} = \frac{1}{\sqrt{d}}\mathcal{V}\hat{\vec{w}}$ roughly behave as non-linear transforms of Gaussian random variables centered around the teacher vector $\boldsymbol{\theta}_{0}$ (or its projection on the covariance spaces) as follows:
\begin{align}
    &\vec{w}^{*} = \Omega^{-1/2}
     \substack{\mbox{\rm prox}\\\frac{1}{\hat{V}^{*}}\reg(\Omega^{-1/2}.)}
    \left(\frac{1}{\hat{V}^{*}}(\hat{m}^{*}\vec{t}+\sqrt{\hat{q}^{*}}\vec{g})\right) ,\,
    \vec{z}^{*} =
    \substack{\mbox{\rm prox}\\V^{*}g(.,\vec{z})}
    \left(\frac{m^{*}}{\sqrt{\rho}}\vec{s}+\sqrt{q^{*}-\frac{(m^{*})^{2}}{\rho}}\vec{h}\right)\, . \nonumber
\end{align}
where $\vec{s},\vec{h}\sim\mathcal{N}(0,\mat{I}_{\samples})$ and $\vec{g}\sim \mathcal{N}(0,\mat{I}_{\sdim})$ are random vectors independent of the other quantities, $\vec{t} = \Omega^{-1/2}\Phi^{\top}\vec{\theta}_{0}$, $\vec{y} = \vec{f}_{0}\left(\sqrt{\rho}\vec{s}\right)$, and $(V^{*},\hat{V}^{*},q^{*},\hat{q}^{*},m^{*},\hat{m}^{*})$ is the unique solution to the fixed point equations presented in Lemma \ref{lemma:fixed-point-eq} of appendix \ref{main-proof}. Those fixed point equations are the generalization of (\ref{eq:main:sp}) to generic, non-separable loss function and regularization. The formal concentration of measure result can then be stated in the following way:
\begin{theorem}(Non-asymptotic version, generic loss and regularization)
\label{main-th-main}
Under Assumption (\ref{main-assumptions}), consider any optimal solution $\hat{\vec{w}}$ to \ref{eq:student}. Then, there exist constants $C,c,c'>0$ such that, for any Lipschitz function $\phi_{1}:\mathbb{R}^{\sdim}\to \mathbb{R}$, and separable, pseudo-Lipschitz function $\phi_{2}:\mathbb{R}^{\samples}\to\mathbb{R}$ and any $0<\epsilon<c'$:
\begin{align}
    &\mathbb{P}\left(\abs{\phi_{1}\!\left(\!\frac{\hat{\vec{w}}}{\sqrt{d}}\right)\!-\mathbb{E}\left[
    \phi_{1}\!\left(\!\frac{\vec{w}^{*}}{\sqrt{d}}\right)\right]
    }\geqslant\epsilon\right) \leqslant\frac{C}{\epsilon^{2}}e^{-cn\epsilon^{4}} \,,
   \mathbb{P}\left(\abs{\phi_{2}\!\left(\!\frac{\hat{\vec{z}}}{\sqrt{n}}\right)\!-\mathbb{E}\left[
   \phi_{2}\!\left(\frac{\vec{z}^{*}}{\sqrt{n}}\right)\right]
   }\geqslant\epsilon\right)\leqslant \frac{C}{\epsilon^{2}}e^{-cn\epsilon^{4}}  .
   \nonumber
\end{align}
\end{theorem}
Note that in this form, the dimensions $n,p,d$ still appear explicitly, as we are characterizing the convergence of the estimator's distribution for large but finite dimension. The clearer, one-dimensional statements are recovered by taking the $n,p,d \to \infty$ limit with separable functions and an $\ell_{2}$ regularization. Other simplified formulas can also be obtained from our general result in the case of an $\ell_1$ penalty, but since this breaks rotational invariance, they do look more involved than the $\ell_2$ case.
From Theorem \ref{main-th-main}, one can deduce the expressions of a number of observables, represented by the test functions $\phi_{1},\phi_{2}$, characterizing the performance of $\hat{\vec{w}}$, for instance the training and generalization error. A more detailed statement, along with the proof, is given in appendix \ref{main-proof}.

%% file: sections/applications.tex
We now discuss how the theorems above are applied to characterise the learning curves for a range of concrete cases. We present a number of cases -- some rather surprising -- for which Conjecture~\ref{conjecture_one} seems valid, and point out some where it is not.  
An out-of-the-box iterator for all the cases studied hereafter is provided in the GitHub repository for this manuscript at~\url{https://github.com/IdePHICS/GCMProject}.

%%%%%%%%%%%%%%%%%%%%%%%%%%%%%%%%%%%%%%%%%%%%%%%%%%%%%%%%%%%%%%%%%
\subsection{Random kitchen sink with Gaussian data}
\label{sec:main:rks}
%%%%%%%%%%%%%%%%%%%%%%%%%%%%%%%%%%%%%%%%%%%%%%%%%%%%%%%%%%%%%%%%%

If we choose random feature maps $\vec{\varphi}_{s}(\vec{x}) = \sigma\left(\mat{F}\vec{x}\right)$ for a random matrix $\mat{F}$ and a chosen scalar function $\sigma$ acting component-wise, we obtain the random kitchen sink model \cite{rahimi2008random}.  This model has seen a surge of interest recently, and a sharp asymptotic analysis was provided in the particular case of uncorrelated Gaussian data $\vec{x}\sim\mathcal{N}(\vec{0}, \mat{I}_{\ddim})$ and $\vec{\varphi}_{t}(\vec{x})=\vec{x}$ in \cite{mei2019generalization,hastie2019surprises} for ridge regression and generalised by \cite{gerace2020generalisation,hu2020universality} for generic convex losses. Both results can be framed as a Gaussian covariate model with:
\begin{align}
    \Psi = \mat{I}_{\tdim}, && \Phi = \kappa_{1}\mat{F}^{\top}, && \Omega = \kappa_{0}^2\vec{1}_{\sdim}\vec{1}_{\sdim}^{\top}+\kappa_{1}^2\frac{\mat{F}\mat{F}^{\top}}{d} + \kappa_{\star}^{2}\mat{I}_{\sdim},  
\end{align}
\noindent where $\vec{1}_{\sdim}\in\mathbb{R}^{\sdim}$ is the all-one vector and the constants $(\kappa_{0},\kappa_{1}, \kappa_{\star})$ are related to the non-linearity $\sigma$:
\begin{align}
    \kappa_{0}\! = \mathbb{E}_{z\sim\mathcal{N}(0,1)}\left[\sigma(z)\right], &&  \kappa_{1}\! = \mathbb{E}_{z\sim\mathcal{N}(0,1)}\left[z\sigma(z)\right], && \kappa_{\star} \! = \sqrt{ \mathbb{E}_{z\sim\mathcal{N}(0,1)}\left[\sigma(z)^2\right]-\kappa_{0}^2 -\kappa_{1}^2}\, .
\end{align}
In this case, the averages over $\mu$ in eq.~\eqref{eq:main:sp} can be directly expressed in terms of the Stieltjes transform associated with the spectral density of $\mat{F}\mat{F}^{\top}$. Note, however, that our present framework can accommodate more involved random sinks models, such as when the teacher features are also a random feature model or multi-layer random architectures. 

%%%%%%%%%%%%%%%%%%%%%%%%%%%%%%%%%%%%%%%%%%%%%%%%%%%%%%%%%%%%%%%%%
\subsection{Kernel methods with Gaussian data}\label{sec:main:kernel}
%%%%%%%%%%%%%%%%%%%%%%%%%%%%%%%%%%%%%%%%%%%%%%%%%%%%%%%%%%%%%%%%

\begin{wrapfigure}{L}{0.52\textwidth}
  \begin{center}
  \vspace{-0.2cm}
  \centerline{\includegraphics[width=0.45\columnwidth]{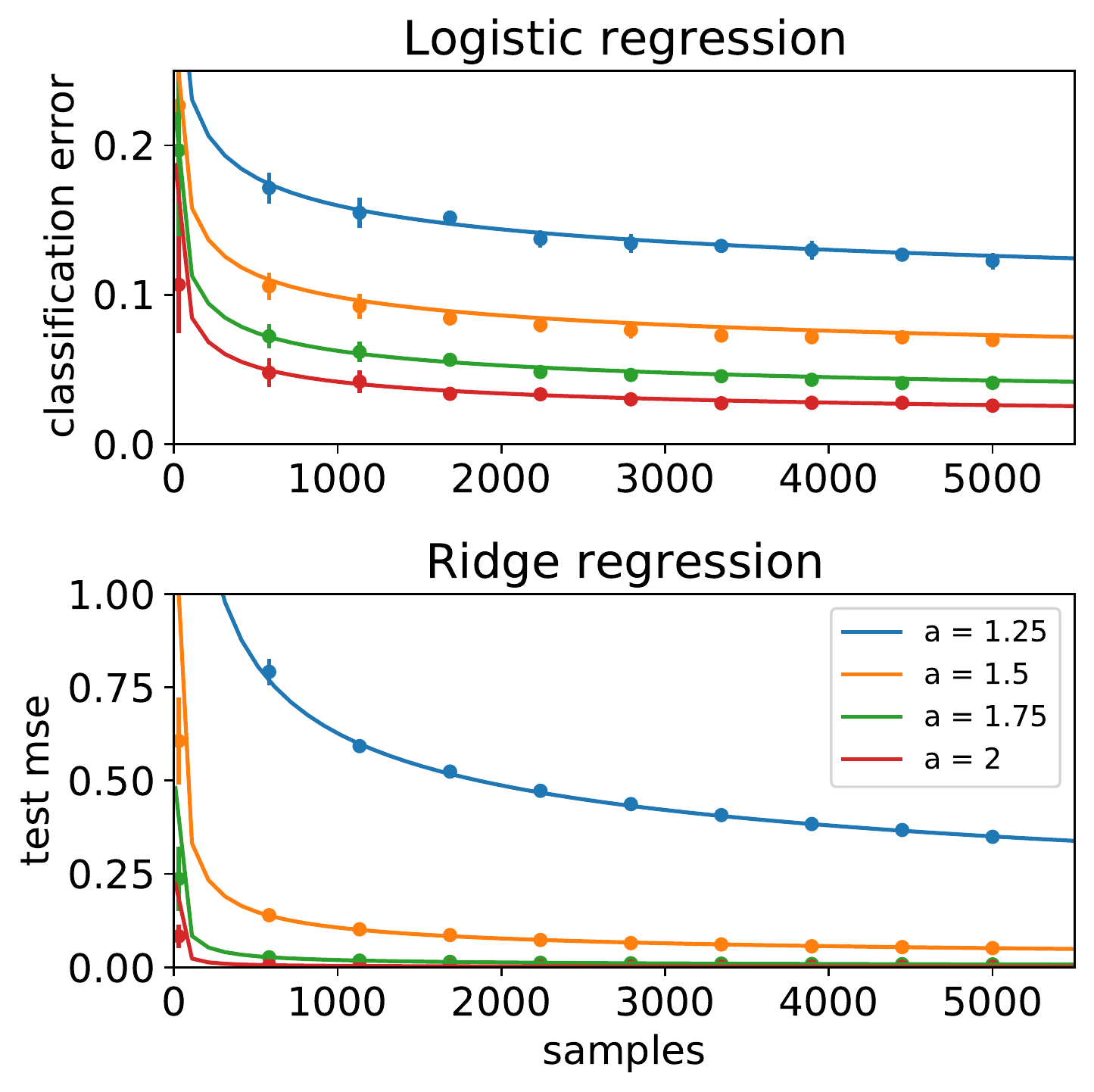}}
  \caption{Learning in kernel space: Teacher and student live in the same (Hilbert) feature space 
  $\vec{v}=\vec{u}\in\mathbb R^{\sdim}$ with $d\gg n$, 
  and the performance only depends on the relative decay between the student spectrum $\omega_{i} = d~i^{-2}$ (the capacity) and the teacher weights in feature space $\theta^2_{0i}\omega_{i} = d~i^{-a}$ (the source). Top: a task with sign teacher (in kernel space), fitted with a max-margin support vector machine (logistic regression with vanishing regularisation \cite{rosset2003margin}). Bottom: a  task with linear teacher (in kernel space) fitted via kernel ridge regression with vanishing regularisation. Points are simulation that matches the theory (lines). Simulations are averaged over $10$ independent runs.}
\label{fig:kernel}
\end{center}
\vspace{-1.cm}
\end{wrapfigure}

Another direct application of our formalism is to kernel methods. Kernel methods admit a dual representation in terms of optimization  over feature space \cite{scholkopf2018learning}. The connection is given by Mercer's theorem, which provides an eigen-decomposition of the kernel and of the target function in the feature basis, effectively mapping kernel regression to a teacher-student problem on feature space. The classical way of studying the performance of kernel methods \cite{steinwart2009optimal, caponnetto2007optimal} is then to directly analyse the performance of convex learning in this space. In our notation, the teacher and student feature maps are equal, and we thus set $\tdim=\sdim, \Psi=\Phi=\Omega=\rm{diag}(\omega_{i})$ where $\omega_{i}$ are the eigenvalues of the kernel and we take the teacher weights $\vec{\theta}_{0}$ to be the decomposition of the target function in the kernel feature basis. 

There are many results in classical learning theory on this problem for the case of ridge regression (where the teacher is usually called "the source" and the eigenvalues of the kernel matrix the "capacity", see e.g.~\cite{steinwart2009optimal,pillaud2018statistical}). However, these are worst case approaches, where no assumption is made on the true distribution of the data. 
In contrast, here we follow a {\it typical case} analysis, assuming Gaussianity in feature space. Through Theorem \ref{thm:corollary:factorised}, this allows us to go beyond the restriction of the ridge loss. An example for logistic loss is in Fig.~\ref{fig:kernel}. 

For the particular case of kernel ridge regression,  Th.~\ref{thm:corollary:factorised} provides a rigorous proof of the formula conjectured in \cite{bordelon2020}.  App.~\ref{sec:app:connection} presents an explicit mapping to their results. Hard-margin Support Vector Machines (SVMs) have also been studied using the heuristic replica method from statistical physics in \cite{opper99, opper01}. In our framework, this corresponds to the \emph{hinge loss} $\loss(x, y) = \text{max}(0, 1-yx)$ when $\lambda\to 0^{+}$.  Our theorem thus puts also these works on rigorous grounds, and extends them to more general losses and regularization.

%%%%%%%%%%%%%%%%%%%%%%%%%%%%%%%%%%%%%%%%%%%%%%%%%%%%%%%%%%%%%%%%%%%%%%%
\subsection{GAN-generated data and learned teachers}\label{sec:main:gan}
%%%%%%%%%%%%%%%%%%%%%%%%%%%%%%%%%%%%%%%%%%%%%%%%%%%%%%%%%%%%%%%%%%%%%%%%

To approach more realistic data sets, we now consider the case in which the input data $\vec{x}\in\mathcal{X}$ is given by a generative neural network $\vec{x} = \mathcal{G}(\vec{z})$, where $\vec{z}$ is a Gaussian i.i.d. latent vector. Therefore, the covariates $[\vec{u},\vec{v}]$ are the result of the following Markov chain:
\begin{align}
    \vec{z} \underset{\mathcal{G}}{\mapsto} \vec{x}\in\mathcal{X}\underset{\vec{\varphi}_{t}}{\mapsto} \vec{u} \in \mathbb{R}^{\tdim}, &&
    \vec{z} \underset{\mathcal{G}}{\mapsto} \vec{x}\in\mathcal{X} \underset{\vec{\varphi}_{s}}{\mapsto}  \vec{v}\in\mathbb{R}^{\sdim}.
    \label{eq:genmodel}
\end{align}
With a model for the covariates, the missing ingredient is the teacher weights $\vec{\theta}_{0}\in\mathbb{R}^{\tdim}$, which determine the label assignment: $y=f_{0}(\vec{u}^{\top}\vec{\theta}_{0})$. In the experiments that follow, we fit the teacher weights \emph{from the original data set in which the generative model $\mathcal{G}$ was trained}. Different choices for the fitting yield different teacher weights, and the quality of label assignment can be accessed by the performance of the fit on the test set. The set $(\vec{\varphi}_{t}, \vec{\varphi}_{s}, \mathcal{G}, \vec{\theta}_{0})$ defines the data generative process. For predicting the learning curves from the iterative eqs.~(\ref{eq:main:sp}) we need to sample from the spectral measure $\mu$, which amounts to estimating the \emph{population} covariances $(\Psi, \Phi, \Omega)$. This is done from the generative process in eq.~(\ref{eq:genmodel}) with a Monte Carlo sampling algorithm. This pipeline is explained in detail in Appendix \ref{details}. An open source implementation of the algorithms used in the experiments is available online at~\url{https://github.com/IdePHICS/GCMProject}.

\begin{figure}[ht]
\begin{center}
\begin{subfigure}[t]{0.45\textwidth}                
    \includegraphics[width=\textwidth]
    {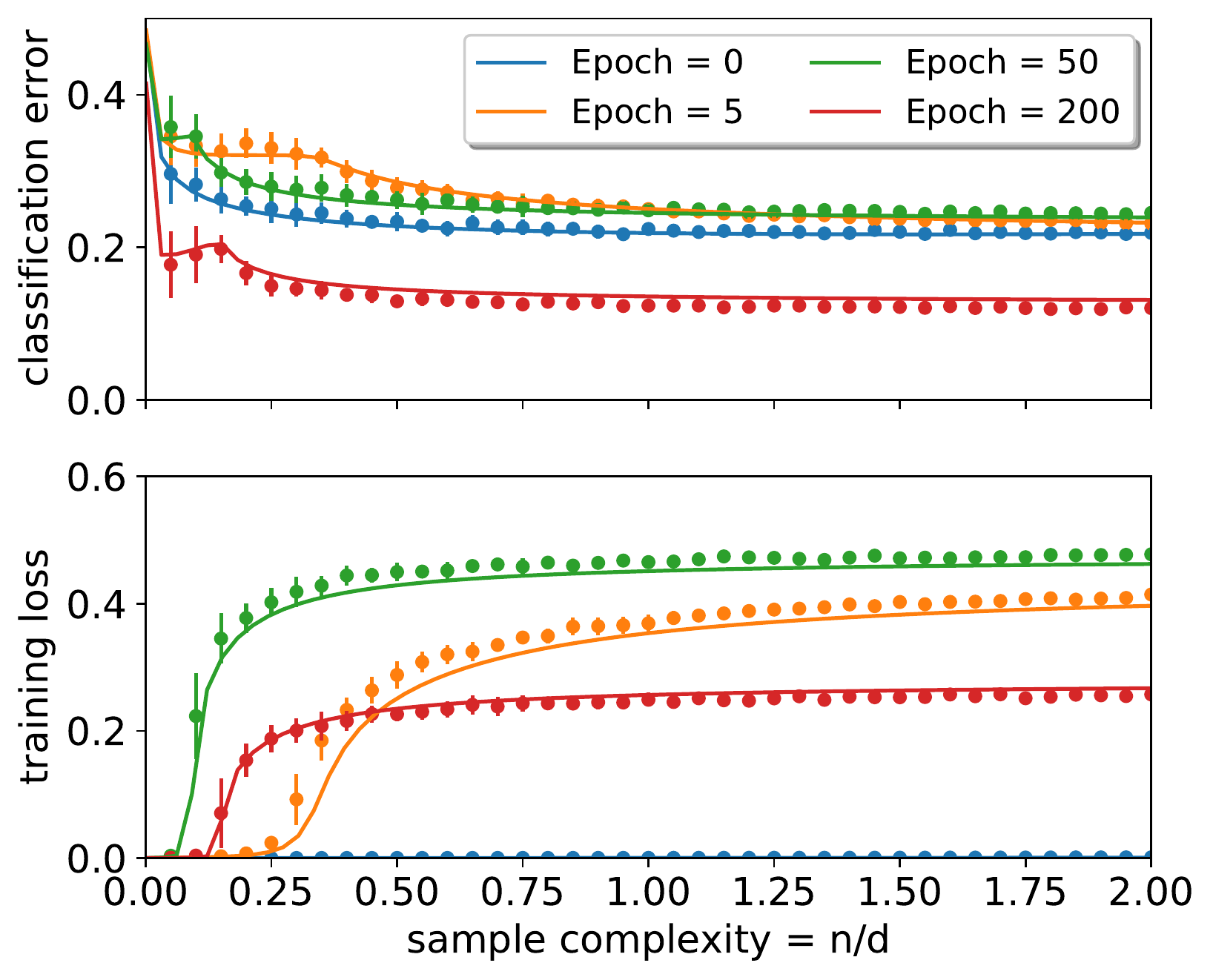}
  \end{subfigure}
\begin{subfigure}[t]{0.51\textwidth}                
\includegraphics[width=\textwidth]
{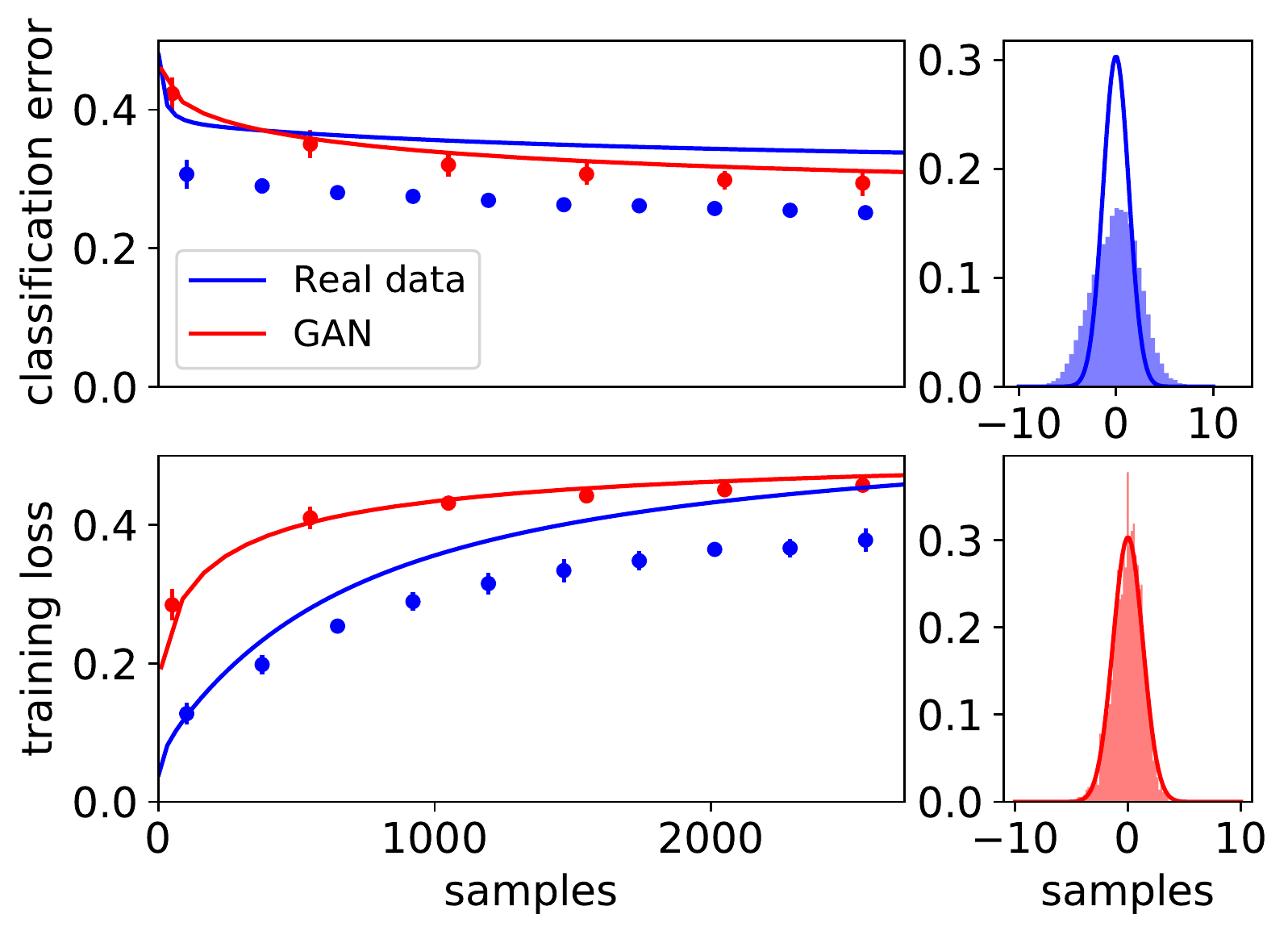}
  \end{subfigure}
  \caption{{\bf Left:} generalisation classification error  
   (top) and (unregularised) training loss
  (bottom) vs the sample complexity $\alpha=\samples/\sdim$ for logistic regression on a learned feature map trained on dcGAN-generated CIFAR10-like images labelled by a teacher fully-connected neural network (see Appendix \ref{sec:app:realdata} for architecture details), with vanishing $\ell_2$
   regularisation.  The different curves compare featured maps at different epochs of training. The theoretical predictions based on the Gaussian covariate model (full lines) are in very good agreement with the actual performance (points).
   {\bf Right:} Test classification error (top) and (unregularised) training loss,
   (bottom) for logistic regression as a function of the number of samples $n$ for an animal vs not-animal binary classification task with $\ell_2$ regularization $\lambda = 10^{-2}$, comparing real CIFAR10 grey-scale images (blue) with dcGAN-generated CIFAR10-like gray-scale images (red). The real-data learning curve was estimated, just as in Figs.~\ref{fig:realdata} from the population covariances on the full data set, and it is not in agreement with the theory in this case. On the very right we depict the histograms of the variable $\frac{1}{\sqrt{\sdim}} \vec{v}^{\top}\hat{\vec{w}}$ for a fixed number of samples $n=2\sdim = 2048$ and the respective theoretical predictions (solid line). Simulations are averaged over $10$ independent runs.
  }
\label{fig:logistic:epochs}
\label{fig:logistic:break}
\end{center}
\end{figure}

Fig.~\ref{fig:logistic:epochs} shows an example of the learning curves resulting from the pipeline discussed above in a logistic regression task on data generated by a GAN trained on CIFAR10 images.
More concretely, we used a pre-trained five-layer deep convolutional GAN (dcGAN) from \cite{radford2015unsupervised}, which maps $100$ dimensional i.i.d. Gaussian noise into $k=32\times 32\times 3$ realistic looking CIFAR10-like images: $\mathcal{G}: \vec{z}\in\mathbb{R}^{100}\mapsto \vec{x}\in\mathbb{R}^{32\times 32\times 3}$. To generate labels, we trained a simple fully-connected four-layer neural network on the \emph{real} CIFAR10 data set, on a odd ($y = +1$) vs. even ($y=-1$) task, achieving $\sim 75\%$ classification accuracy on the test set. The teacher weights $\vec{\theta}_{0}\in\mathbb{R}^{\tdim}$ were taken from the last layer of the network, and the teacher feature map $\vec{\varphi}_{t}$ from the three previous layers. For the student model, we trained a completely independent fully connected $3$-layer neural network on the dcGAN-generated CIFAR10-like images and took snapshots of the feature maps $\vec{\varphi}^{i}_{s}$ induced by the $2$-first layers during the first $i\in\{0, 5, 50, 200\}$ epochs of training. Finally, once $\left(\mathcal{G}, \vec{\varphi}_{t}, \vec{\varphi}^{i}_{s}, \vec{\theta}_{0}\right)$ have been fixed, we estimated the covariances $(\Psi, \Phi, \Omega)$ with a Monte Carlo algorithm. Details of the architectures used and of the training procedure can be found in Appendix.~\ref{sec:app:realdata}.

Fig.~\ref{fig:logistic:epochs} depicts the resulting learning curves obtained by training the last layer of the student. Interestingly, the performance of the feature map at epoch $0$ (random initialisation) beats the performance of the learned features during early phases of training in this experiment. Another interesting behaviour is given by the separability threshold of the learned features, i.e. the number of samples for which the training loss becomes larger than $0$ in logistic regression. At epoch $50$ the learned features are separable at lower sample complexity $\alpha = \samples/\sdim$ than at epoch $200$ - even though in the later the training and generalisation performances are better.

%%%%%%%%%%%%%%%%%%%%%%%%%%%%%%%%%%%%%%%%%%%%%%%%%%%%%%%%%%%%%%%%
\subsection{Learning from real data sets}
\label{sec:regression:real}
%%%%%%%%%%%%%%%%%%%%%%%%%%%%%%%%%%%%%%%%%%%%%%%%%%%%%%%%%%%%%%%%
\paragraph{Applying teacher/students to a real data set ---} 
Given that the learning curves of realistic-looking inputs can be captured by the Gaussian covariate model, it is fair to ask whether the same might be true for \emph{real data sets}. To test this idea, we first need to cast the real data set into the teacher-student formalism, and then compute the covariance matrices $\Omega,\Psi,\Phi$ and teacher vector $\vec{\theta}_0$ required by model (\ref{def:GM3}).

Let $\{\vec{x}^{\mu}, y^{\mu}\}_{\mu=1}^{n_{\tot}}$ denote a real data set, e.g.~MNIST or Fashion-MNIST for concreteness, where $n_{\tot} = 7\times 10^{4}$, $\vec{x}^{\mu}\in\mathbb{R}^{\ddim}$ with $\ddim =784$. Without loss of generality, we can assume the data is centred. To generate the teacher, let $\vec{u}^{\mu} = \vec{\varphi}_{t}(\vec{x}^{\mu})\in\mathbb{R}^{\tdim}$ be a feature map such that data is invertible in feature space, i.e. that $y^{\mu} = \vec{\theta}_{0}^{\top}\vec{u}^{\mu}$ for some teacher weights $\vec{\theta}_{0}\in\mathbb{R}^{\tdim}$, which should be computed from the samples. Similarly, let $\vec{v}^{\mu}=\vec{\varphi}_{s}(\vec{x}^{\mu})\in\mathbb{R}^{\sdim}$ be a feature map we are interested in studying. 
Then, we can estimate the population covariances $(\Psi, \Phi,\Omega)$ empirically from the \emph{entire} data set as:
\begin{align}
\Psi = \sum\limits_{\mu=1}^{n_{\tot}}
\frac {\vec{u}^{\mu}{\vec{u}^{\mu}}^{\top}}{n_{\tot}}, && \Phi =  \sum\limits_{\mu=1}^{n_{\tot}} \frac{\vec{u}^{\mu}{\vec{v}^{\mu}}^{\top}}{n_{\tot}}, &&
\Omega = \sum\limits_{\mu=1}^{n_{\tot}}\!
\frac {\vec{v}^{\mu}{\vec{v}^{\mu}}^{\top}}{n_{\tot}}.
\label{eq:main:empcov}
\end{align}
At this point, we have all we need to run the self-consistent equations (\ref{eq:main:sp}). The issue with this approach is that there is not a unique teacher map $ \vec{\varphi}_{t}$ and teacher vector~$\vec{\theta}_0$ that fit the true labels.
However, we can show that \emph{all interpolating linear teachers are equivalent}:
\begin{theorem}(Universality of linear teachers)
    For any teacher feature map $\vec{\varphi}_{t}$, and for any $\vec{\theta}_0$ that interpolates the data so that $y^{\mu} = \vec{\theta}_{0}^{\top}\vec{u}^{\mu}\, \forall \mu$, the asymptotic predictions of model (\ref{def:GM3}) are equivalent.
    \label{thm:main:universality}
\end{theorem}
\vspace{-0.6cm}
\begin{proof} It follows from the fact that the teacher weights and covariances only appear in eq.~\eqref{eq:main:sp} through $\rho = \frac{1}{\tdim}\vec{\theta}_{0}^{\top}\Psi\vec{\theta}_{0}$ and the projection $\Phi^{\top}\vec{\theta}_{0}$. Using the estimation \eqref{eq:main:empcov} and the assumption that it exists  $y^{\mu}=\vec{\theta}_{0}^{\top}\vec{u}^{\mu}$, one can write these quantities directly from the labels $y^{\mu}$:
\begin{align}
    \rho = \frac{1}{n_{\tot}}\sum\limits_{\mu=1}^{n_{\tot}}\left(y^{\mu}\right)^2, && \Phi^{\top}\vec{\theta}_{0} =\frac{1}{n_{\tot}} \sum\limits_{\mu=1}^{n_{\tot}}y^{\mu}\vec{v}^{\mu}\, .
\end{align}
For linear interpolating teachers, results are thus independent of the choice of the teacher. 
\end{proof}
Although this result might seen surprising at first sight, it is quite intuitive. Indeed, the information about the teacher model only enters the Gaussian covariate model~\eqref{def:GM3} through the statistics of $\vec{u}^{\top}\vec{\theta}_{0}$. For a linear teacher $f_{0}(x)=x$, this is precisely given by the labels.

\paragraph{Ridge Regression with linear teachers ---} 

\begin{wrapfigure}{L}{0.5\textwidth}
    \includegraphics[width=0.45\textwidth]{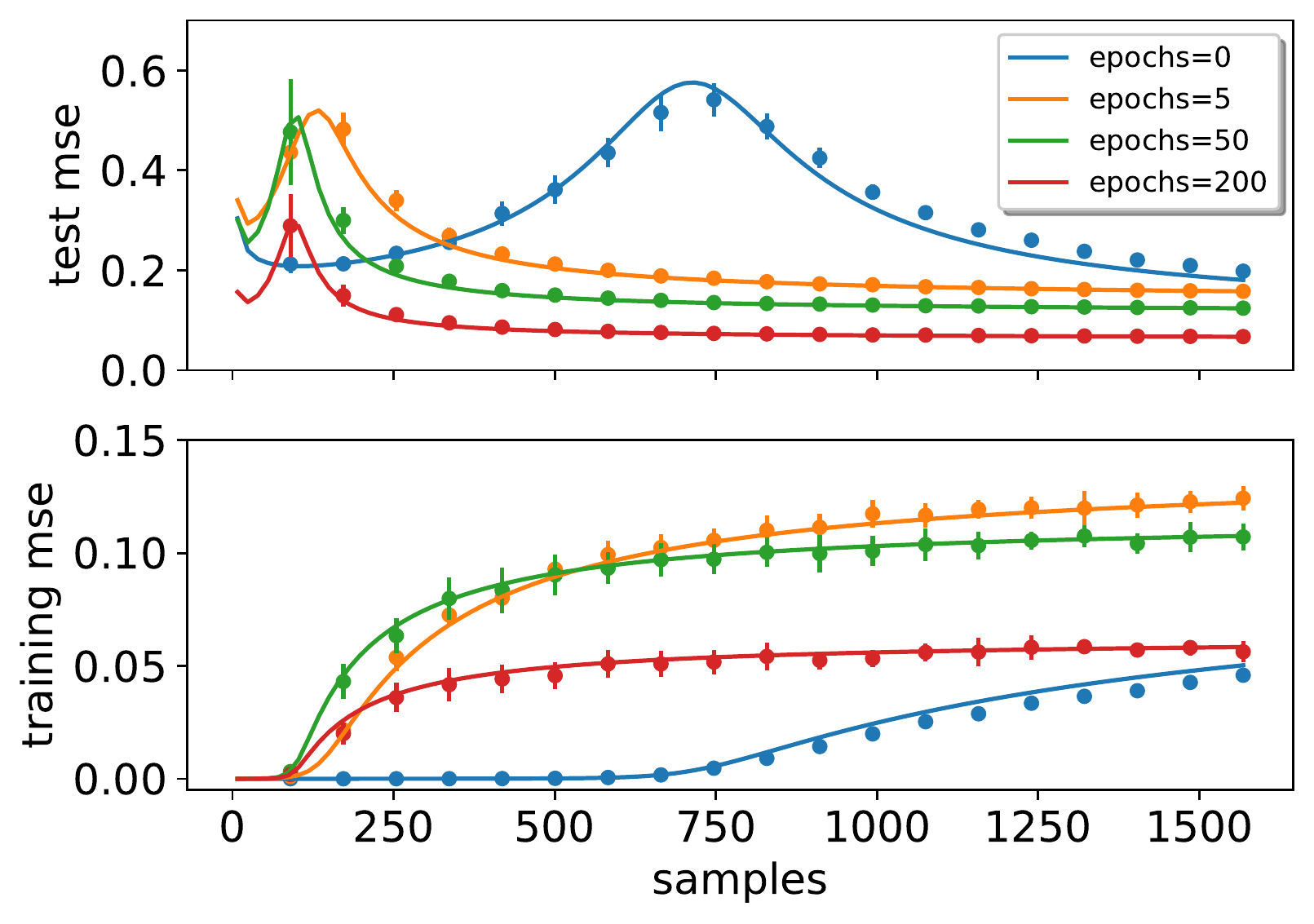}
  \caption{Test and training mean-squared errors eqs.~\eqref{eq:mse} as a function of the number of samples $\samples$ for ridge regression. 
  The Fashion-MNIST data set, with vanishing regularisation $\lambda = 10^{-5}$. In this plot, the student feature map $\vec{\varphi}_{s}$ is a 3-layer fully-connected neural network with $\sdim=2352$ hidden neurons trained on the full data set with the square loss. Different curves correspond to the feature map obtained at different stages of training. Simulations are averaged over $10$ independent runs.
  Further details on the simulations are described in Appendix \ref{sec:app:realdata}}
\label{fig:realdata}
\end{wrapfigure}

We now test the prediction of model (\ref{def:GM3}) on real data sets, and show that it is surprisingly effective in predicting the learning curves, at least for the ridge regression task. We have trained a 3-layer fully connected neural network with ReLU activations on the full Fashion-MNIST data set to distinguish clothing used above vs. below the waist \cite{xiao2017}. The student feature map $\vec{\varphi}_{s}:\mathbb{R}^{784}\to \mathbb{R}^{\sdim}$ is obtained by removing the last layer, see Appendix \ref{sec:app:realdata} for a detailed description. In Fig.~\ref{fig:realdata} we show the test and training errors of the ridge estimator on a sub-sample of $n<n_{\rm{tot}}$ on the Fashion-MNIST images. We observe remarkable agreement between the learning curve obtained from simulations and the theoretical prediction by the matching Gaussian covariate model. Note that for the square loss and for $\lambda\ll 1$, the worst performance peak is located at the point in which the linear system becomes invertible. Curiously, Fig. ~\ref{fig:realdata} shows that the fully-connected network progressively learns a low-rank representation of the data as training proceeds. This can be directly verified by counting the number of zero eigenvalues of $\Omega$, which go from a full-rank matrix to a matrix of rank $380$ after 200 epochs of training.

Fig.~\ref{fig:setup} (right) shows a similar experiment on the MNIST data set, but for different out-of-the-box feature maps, such as random features and the scattering transform \cite{bruna2012}, and we chose the number of random features $\sdim = 1953$ to match the number of features from the scattering transform. Note the characteristic double-descent behaviour \cite{opper1996statistical,spigler2019jamming,belkin2019reconciling}, and the accurate prediction of the peak where the interpolation transition occurs. 
We note in Appendix~\ref{sec:app:realdata} that for both Figs.~\ref{fig:realdata} and \ref{fig:setup}, for a number of samples $n$ closer to $n_{\rm tot}$ we start to see deviations between the real learning curve and the theory. This is to be expected since in the teacher-student framework the student can, in principle, express the same function as the teacher if it recovers its weights exactly. Recovering the teacher weights becomes possible with a large training set. In that case, its test error will be zero. However, in our setup the test error on real data remains finite even if more training data is added, leading to the discrepancy between teacher-student learning curve and real data, see Appendix~\ref{sec:app:realdata} for further discussion.

Why is the Gaussian model so effective for describing learning with data that are {\it not} Gaussian? The point is that ridge regression is sensitive only to second order statistics, and not to the full distribution of the data. It is a classical property (see Appendix \ref{RidgeUniversality}) that the training and generalisation errors are only a function of the spectrum of the \emph{empirical} and \emph{population} covariances, and of their products. Random matrix theory teaches us that such quantities are very robust, and their asymptotic behaviour is universal for a broad class of distributions of $[\vec{u},\vec{v}]$ \cite{bai2008large,ledoit2011eigenvectors,el2009concentration,louart2018concentration}. The asymptotic behavior of kernel matrices has indeed been the subject of intense scrutiny \cite{el2010spectrum,cheng2013spectrum,pennington2017nonlinear,mei2019generalization,fan2019spectral,seddik2020random}. Indeed, a universality result akin to Theorem \ref{thm:main:universality} was noted in \cite{jacot2020kernel} in the specific case of kernel methods. We thus expect the validity of model (\ref{def:GM3}) for ridge regression, with a linear teacher, to go way beyond the Gaussian assumption.

\vspace{-0.2cm}

\paragraph{Beyond ridge regression ---} 
The same strategy fails beyond ridge regression and mean-squared test
error. This suggests a limit in the application of model (\ref{def:GM3}) to real
(non-Gaussian) data to the universal linear teacher.  To illustrate this,
consider the setting of Figs.~\ref{fig:realdata}, and
compare the model predictions for the binary classification error instead of the
$\ell_2$ one. There is a clear mismatch between the simulated performance and
prediction given by the theory (see Appendix \ref{sec:app:realdata}) due to the
fact that the classification error does not depends only on the first two
moments.

We present an additional experiment in Fig.~\ref{fig:logistic:break}. We compare
the learning curves of logistic regression on a classification task on the
\emph{real} CIFAR10 images with the real labels versus the one on
dcGAN-generated CIFAR10-like images and teacher generated labels from
Sec.~\ref{sec:main:gan}. While the Gaussian theory captures well the behaviour
of the later, it fails on the former. A histogram of the distribution of the
product $\vec{u}^{\top}\hat{\vec{w}}$ for a fixed number of samples illustrates
well the deviation from the prediction of the theory with the real case, in
particular on the tails of the distribution. The difference between GAN
generated data (that fits the Gaussian theory) and real data is clear.  Given
that for classification problems there exists a number of choices of "sign"
teachers and feature maps that give the exact same labels as in the data set, an
interesting open question is: \emph{is there a teacher that allows to reproduce the
  learning curves more accurately}? This question is left for future works.
  
  \vspace{-0.2cm}
  

%% file: sections/acknowledgements.tex
We thank Romain Couillet, Cosme Louart,  Loucas Pillaud-Vivien, Matthieu Wyart, Federica Gerace, Luca Saglietti and Yue Lu for discussions. We are grateful to Kabir Aladin Chandrasekher, Ashwin Pananjady and Christos Thrampoulidis for pointing out discrepancies in the finite size rates and insightful related discussions. We acknowledge funding from the ERC under the European Union's Horizon 2020 Research and Innovation Programme Grant Agreement 714608-SMiLe, and from the French
National Research Agency grants ANR-17-CE23-0023-01 PAIL.

%% file: sections/appendix/replicas.tex
In this appendix we derive the formula for the performance of the Gaussian covariate model from a heuristic replica analysis. The computation closely follows the recent developments in  \cite{gerace2020generalisation,aubin2020generalization}. We refer to \cite{mezard1987spin,engel2001statistical,mezard2009information} for an introduction to this remarkable heuristic (but seemingly never failing) approach.

\paragraph{The data: } First, let's recall the definition of our model. Consider synthetic labelled data $(\vec{v}, y)\in\mathbb{R}^{\sdim}\times \mathbb{R}$ drawn independently from a joint distribution with density:
\begin{align}
    p_{\vec{\theta}_{0}}(\vec{v}, y) = \int_{\mathbb{R}^{\tdim}}\dd\vec{u}~P_{0}(y|\vec{u}^{\top}\vec{\theta}_{0}) \mathcal{N}(\vec{u},\vec{v}; \vec{0}, \Sigma)
\end{align}
\noindent where $P_{0}$ is a given likelihood on $\mathbb{R}$, $\vec{\theta}_{0}\in\mathbb{R}^{\tdim}$ is a fixed vector of parameters and $\Sigma$ is a correlation matrix given by:
\begin{align}
    \Sigma = \begin{bmatrix}\Psi & \Phi \\ \Phi^{\top} & \Omega \end{bmatrix}\in\mathbb{R}^{(\tdim+\sdim)\times (\tdim+\sdim)}
\end{align}
\noindent for symmetric positive semi-definite matrices $\Psi$ and $\Omega$ and $\Phi \in\mathbb{R}^{\tdim\times\sdim}$. In its simplest form, which we will mostly be using in the applications, we take the likelihood $P_{0}(y|x) = \delta(y-f_{0}(x))$ to be a deterministic function with $f_{0}:\mathbb{R}\to\mathbb{R}$ a non-linearity, e.g. $f_{0}(x)=\sign(x)$ to generate binary labels.

\paragraph{The task: } In our analysis, we are interested in the training and generalisation performance of a linear classifier $\hat{y} = f_{\vec{w}}(\vec{v}) = \hat{f}\left(\vec{w}^{\top}\vec{v}\right)$ trained on $\samples$ independent samples $\mathcal{D} = \{(\vec{v}^{\mu}, y^{\mu})\}_{\mu=1}^{\samples}$ from $p_{\vec{\theta}_{0}}$ by minimising the regularised empirical risk:
\begin{align}
\hat{\vec{w}} = \underset{\vec{w}\in\mathbb{R}^{\sdim}}{\argmin}\left[\sum\limits_{\mu=1}^{n}\loss\left(y^{\mu},\vec{w}^{\top}\vec{v}^{\mu}\right)+\frac{\lambda}{2}||\vec{w}||_{2}^{2}\right],
\label{eq:app:argmin}
\end{align}
\noindent where $\lambda>0$ is the regularisation strength. We define the \emph{sample complexity} $\alpha = \samples/\sdim$ and the \emph{aspect ratio} $\ar = \tdim/\sdim$.

\paragraph{Gibbs minimisation:} As it was proven in Theorem \ref{Train_gen} of the main manuscript, the asymptotic performance of the estimator in eq.~\eqref{eq:app:argmin} is fully characterised by the following scalar parameters:
\begin{align}
    \rho = \frac{1}{\tdim}\vec{\theta}_{0}^{\top}\Psi\vec{\theta}_{0}, && m^{\star} = \frac{1}{\sqrt{\tdim\sdim}}\vec{\theta}_{0}^{\top}\Phi\hat{\vec{w}}, && q^{\star} = \frac{1}{\sdim}\hat{\vec{w}}^{\top}\Omega\hat{\vec{w}}
\end{align}
The replica method is precisely a heuristic tool allowing us to circumvent the high-dimensional estimation problem defined in eq.~\eqref{eq:app:argmin} and giving us direct access to $(m^{\star}, q^{\star})$.

The starting point is to define the following Gibbs measure over weights $\vec{w}\in\mathbb{R}^{\sdim}$:
\begin{align}
\mu_{\beta}(\dd\vec{w}) = \frac{1}{\mathcal{Z}_{\beta}} e^{-\beta\left[\sum\limits_{\mu=1}^{\samples}\loss\left(y^{\mu},\vec{w}^{\top}\vec{v}^{\mu}\right)+\frac{\lambda}{2}\sum\limits_{i=1}^{\sdim}w_{i}^2\right]}\dd\vec{w}	= \frac{1}{\mathcal{Z}_{\beta}}\underbrace{\prod\limits_{\mu=1}^{\samples} e^{-\beta\loss\left(y^{\mu},\vec{w}^{\top}\vec{v}^{\mu}\right)}}_{P_{\loss}}\underbrace{\prod\limits_{i=1}^{\sdim}e^{-\frac{\beta\lambda}{2}w_{i}^2}\dd w_{i}}_{P_{w}}
\end{align}
where $\mathcal{Z}_{\beta}$, known as the \emph{partition function}, is a constant normalising the Gibbs measure $\mu_{\beta}$:
\begin{align}
    \mathcal{Z}_{\beta} = \int_{\mathbb{R}^{d}}\left(\prod\limits_{i=1}^{\sdim}\dd w_{i}\right)~ e^{-\frac{\beta\lambda}{2}w_{i}^2}\prod\limits_{\mu=1}^{\samples} e^{-\beta\loss\left(y^{\mu},\vec{w}^{\top}\vec{v}^{\mu}\right)}
\end{align}
Note that $P_{\loss}$ and $P_{w}$ can be interpreted as a (unormalised) likelihood and prior distribution respectively. In the limit $\beta\to\infty$, the measure $\mu_{\beta}$ concentrates around solutions of the minimisation in eq.~\eqref{eq:app:argmin}. The aim in the replica method is to compute the free energy density, defined as:
\begin{align}
\beta f_{\beta} = -\lim\limits_{\sdim\to\infty}\frac{1}{\sdim}\mathbb{E}_{\mathcal{D}}\log\mathcal{Z}_{\beta}.
\label{eq:app:freeen}
\end{align}

%%%%%%%%%%%%%%%%%%%%%%%%%%%%%%%%%%%%%%%%%%%%%%%%%%
\subsection{Replica computation of the free energy}
%%%%%%%%%%%%%%%%%%%%%%%%%%%%%%%%%%%%%%%%%%%%%%%%%%
The average in eq.~\eqref{eq:app:freeen} is not straightforward due to the logarithm term. The replica method consists of computing it using the following trick to get rid of the logarithm: 
\begin{align}
	\log\mathcal{Z}_{\beta}	= \lim\limits_{r\to 0^{+}}\frac{1}{r}\partial_{r}\mathcal{Z}^{r}_{\beta}
\end{align}

%%%%%%%%%%%%%%%%%%%%%%%%
\subsection*{Averaging}
%%%%%%%%%%%%%%%%%%%%%%%%
Applying the trick above, the computation of the free energy density boils down to the evaluation of the averaged replicated partition function:
\begin{align}
\mathbb{E}_{\mathcal{D}}\mathcal{Z}^{r}_{\beta} &= \prod\limits_{\mu=1}^{n}\mathbb{E}_{(\vec{v}^{\mu}, y^{\mu})}	\prod\limits_{a=1}^{r}\int_{\mathbb{R}^{\sdim}}P_{w}(\dd\vec{w}^{a})P_{\loss}\left(y^{\mu}\Big|\frac{\vec{v}^{\mu}\cdot \vec{w}^{a}}{\sqrt{\sdim}}\right)\notag\\
&=\prod\limits_{\mu=1}^{n}\int_{\mathbb{R}}\dd y^{\mu}\int_{\mathbb{R}^{\tdim}}P_{\vec{\theta}_{0}}(\dd\vec{\theta}_{0})\int_{\mathbb{R}^{\sdim\times r}}\left(\prod\limits_{a=1}^{r}P_{w}(\dd\vec{w}^{a})\right)\underbrace{\mathbb{E}_{\vec{u}^{\mu},\vec{v}^{\mu}}\left[P_{0}\left(y^{\mu}|\frac{\vec{u}^{\mu}\cdot\vec{\theta}_{0}}{\sqrt{\tdim}}\right)\prod\limits_{a=1}^{r}P_{\loss}\left(y^{\mu}|\frac{\vec{v}^{\mu}\cdot\vec{w}^{a}}{\sqrt{\sdim}}\right)\right]}_{(\star)}
\end{align}
Note that in the above we included an average over the parameters $\vec{\theta}_{0}\in\mathbb{R}^{\tdim}$. The case in which $\vec{\theta}_{0}$ is a fixed vector can be recovered by choosing a point mass $P_{\vec{\theta}_{0}} = \delta_{\vec{\theta}_{0}}$. Focusing on the average term in brackets:
\begin{align}
	(\star) &= \mathbb{E}_{(\vec{u},\vec{v})}\left[P_{0}\left(y^{\mu}\Big|\frac{\vec{u}^{\mu}\cdot\vec{\theta}_{0}}{\sqrt{\tdim}}\right)\prod\limits_{a=1}^{r}P_{\loss}\left(y^{\mu}\Big|\frac{\vec{v}^{\mu}\cdot\vec{w}^{a}}{\sqrt{\sdim}}\right)\right]\notag\\ 
	&= \int_{\mathbb{R}}\dd\nu_{\mu}P_{0}\left(y|\nu_{\mu}\right)\int_{\mathbb{R}^{r}}\left(\prod\limits_{a=1}^{r}\dd\lambda^{a}_{\mu}P_{\loss}(y^{\mu}|\lambda_{\mu}^{a})\right)~\underbrace{\mathbb{E}_{(\vec{u}^{\mu},\vec{v}^{\mu})}\left[\delta\left(\nu_{\mu}-\frac{\vec{u}^{\mu}\cdot\vec{\theta}_{0}}{\sqrt{\tdim}}\right)\prod\limits_{a=1}^{r}\delta\left(\lambda^{a}_{\mu}-\frac{\vec{v}^{\mu}\cdot\vec{w}^{a}}{\sqrt{\sdim}}\right)\right]}_{P(\nu,\lambda)}\notag
\end{align}
Note that the term in brackets defines the joint density over $(\nu_{\mu},\lambda^{a}_{\mu})$. It is easy to check that these are Gaussian random variables with zero mean and covariance matrix given by:
\begin{align}
\Sigma^{ab} = 
\begin{pmatrix}
 	\rho & m^{a}\\
 	m^{a} & Q^{ab}
\end{pmatrix}.
\end{align}
\noindent where the so-called overlap parameters $(\rho, m^{a}, Q^{ab})$ are related to the weights $\vec{\theta}_{0}, \vec{w}$:
\begin{align}
\rho &\equiv \mathbb{E}\left[\nu_{\mu}^2\right] = \frac{1}{\tdim}{\vec{\theta}_{0}}^{\top}\Psi\vec{\theta}_{0}, && m^{a} \equiv \mathbb{E}\left[\lambda_{\mu}^{a}\nu_{\mu}\right]= \frac{1}{\sqrt{\tdim\sdim}}\vec{\theta}_{0}^{\top}\Phi{\vec{w}^{a}}, && Q^{ab} \equiv \mathbb{E}\left[\lambda_{\mu}^{a}\lambda_{\mu}^{b}\right]= \frac{1}{\sdim}{\vec{w}^{a}}^{\top}\Omega \vec{w}^{b}	\notag
\end{align}
We can therefore write the averaged replicated partition function as:
\begin{align}
\mathbb{E}_{\mathcal{D}}\mathcal{Z}^{r}_{\beta} &=\prod\limits_{\mu=1}^{n}\int\dd y^{\mu}\int_{\mathbb{R}^{\tdim}}P_{\vec{\theta}_{0}}(\dd\vec{\theta}_{0})	\int_{\mathbb{R}^{\sdim\times r}}\left(\prod\limits_{a=1}^{r}P_{w}(\dd\vec{w}^{a})\right)\int_{\mathbb{R}}\dd\nu_{\mu}P_{0}(y^{\mu}|\nu_{\mu})\times\notag\\
&\quad\times\int_{\mathbb{R}^{r}}\left(\prod\limits\dd\lambda_{\mu}^{a}P_{\loss}\left(y^{\mu}|\lambda_{\mu}^{a}\right)\right)\mathcal{N}(\nu_{\mu}, \lambda^{a}_{\mu};\vec{0},\Sigma^{ab})
\label{eq:avgZr:2}
\end{align}

%%%%%%%%%%%%%%%%%%%%%%%%%%%%%%%%%%%%%%%%%%%%%%%%%
\subsection*{Rewriting as a saddle-point problem}
%%%%%%%%%%%%%%%%%%%%%%%%%%%%%%%%%%%%%%%%%%%%%%%%%
The next step is to free the overlap parameters by introducing delta functions:
\begin{align}
1 &\propto \int_{\mathbb{R}}\dd\rho~\delta\left(\tdim\rho - {\vec{\theta}_{0}}^{\top}\Psi\vec{\theta}_{0}\right)\int_{\mathbb{R}^{r}}\prod\limits_{a=1}^{r} \dd m^{a}~\delta\left(\sqrt{\tdim\sdim} m^{a}-\vec{\theta}_{0}^{\top}\Phi\vec{w}^{a}\right)\notag\\
&\qquad\times\int_{\mathbb{R}^{r\times r}}\prod\limits_{1\leq a\leq b\leq r}\dd Q^{ab}~\delta\left(\sdim Q^{ab}-{\vec{w}^{a}}^{\top}\Omega\vec{w}^{b}\right)\notag\\
&=\int_{\mathbb{R}}\frac{\dd\rho\dd\hat{\rho}}{2\pi}~e^{i\hat{\rho}\left(\tdim\rho - {\vec{\theta}_{0}}^{\top}\Psi\vec{\theta}_{0}\right)}\int_{\mathbb{R}^{r}}\prod\limits_{a=1}^{r} \frac{\dd m^{a}\dd\hat{m}^{a}}{2\pi}~e^{i\sum\limits_{a=1}^{r}\hat{m}^{a}\left(\sqrt{\tdim\sdim} m^{a}-\vec{\theta}_{0}^{\top}\Phi\vec{w}^{a}\right)}\times\notag\\
&\qquad\times\int_{\mathbb{R}^{r\times r}}\prod\limits_{1\leq a\leq b\leq r}\frac{\dd Q^{ab}\dd\hat{Q}^{ab}}{2\pi}~e^{i\sum\limits_{1\leq a\leq b\leq r}\hat{Q}^{ab}\left(\sdim Q^{ab}-{\vec{w}^{a}}^{\top}\Omega\vec{w}^{b}\right)}
\end{align}
Inserting this in eq.~\eqref{eq:avgZr:2} allow us to rewrite:
\begin{align}
\mathbb{E}_{\mathcal{D}}\mathcal{Z}_{\beta}^{r} = 	\int_{\mathbb{R}}\frac{\dd\rho\dd\hat{\rho}}{2\pi}\int_{\mathbb{R}^{r}}\prod\limits_{a=1}^{r} \frac{\dd m^{a}\dd\hat{m}^{a}}{2\pi}\int_{\mathbb{R}^{r\times r}}\prod\limits_{1\leq a\leq b\leq r}\frac{\dd Q^{ab}\dd\hat{Q}^{ab}}{2\pi} e^{\sdim\Phi^{(r)}}
\label{eq:avgZr:3}
\end{align}
\noindent where we have absorbed a $-i$ factor in the integrals (this won't matter since we will look to the saddle-point) and defined the potential:
\begin{align}
\Phi^{(r)} = -\ar \rho\hat{\rho}	-\sqrt{\ar}\sum\limits_{a=1}^{r}m^{a}\hat{m}^{a}-\sum\limits_{1\leq a\leq b\leq r}Q^{ab}\hat{Q}^{ab}+\alpha\Psi^{(r)}_{y}(\rho,m^{a}, Q^{ab}) + \Psi^{(r)}_{w}(\hat{\rho},\hat{m}^{a},\hat{Q}^{ab})
\label{eq:replica:Phir}
\end{align}
\noindent where we recall that $\alpha = \samples/\sdim$, $\ar = \tdim/\sdim$ and:
\begin{align}
\Psi_{w}^{(r)} &= \frac{1}{\sdim}\log\int_{\mathbb{R}^{\tdim}}P_{\vec{\theta}_{0}}\left(\dd\vec{\theta}_{0}\right)\int_{\mathbb{R}^{\sdim\times r}}\prod\limits_{a=1}^{r}P_{w}\left(\dd\vec{w}^{a}\right) e^{\hat{\rho}{\vec{\theta}_{0}}^{\top}\Psi\vec{\theta}_{0}+\sum\limits_{a=1}^{r}\hat{m}^{a}\vec{\theta}_{0}^{\top}\Phi\vec{w}^{a}+\sum\limits_{1\leq a\leq b\leq r}\hat{Q}^{ab}{\vec{w}^{a}}^{\top}\Omega\vec{w}^{b}}\\
\Psi_{y}^{(r)} &= \log\int_{\mathbb{R}}\dd y\int_{\mathbb{R}}\dd\nu~P_{0}(y|\nu)\int\prod\limits_{a=1}^{r}\dd\lambda^{a}P_{\loss}(y|\lambda^{a})~ \mathcal{N}(\nu,\lambda^{a};\vec{0},\Sigma^{ab})
\end{align}
In the high-dimensional limit where $\sdim\to\infty$ while $\alpha = \samples/\sdim$ and $\ar = \tdim/\sdim$ stay finite, the integral in eq.~\eqref{eq:avgZr:3} concentrate around the values of the overlaps that extremise $\Phi^{(r)}$, and therefore we can write:
\begin{align}
\beta f_{\beta} = -\lim\limits_{r\to 0^{+}}\frac{1}{r}\extr~ \Phi^{(r)}\left(\hat{\rho},\hat{m}^{a},\hat{Q}^{ab};\rho,m^{a},Q^{ab}\right)
\end{align}
%%%%%%%%%%%%%%%%%%%%%%%%%%%%%%%%%%%%%%%
\subsection*{Replica symmetric ansatz}
%%%%%%%%%%%%%%%%%%%%%%%%%%%%%%%%%%%%%%%
In order to proceed with the $r\to 0^{+}$ limit, we restrict the extremisation above to the following replica symmetric ansatz:
\begin{align}
m^{a} = m, && \hat{m}^{a} = \hat{m}, &&\text{ for } a=1,\dots, r	\notag\\
Q^{aa} = r, && \hat{Q}^{aa} = -\frac{1}{2}\hat{r}, &&\text{ for } a=1,\dots, r	\notag\\
Q^{ab} = q, && \hat{Q}^{ab} = \hat{q}, &&\text{ for } 1\leq a<b\leq r
\end{align}
Inserting this ansatz in eq.~\eqref{eq:replica:Phir} allows us to explicitly take the $r\to 0^{+}$ limit for each term. The first three terms are straightforward to obtain. The limit of $\Psi_{y}^{(r)}$ is cumbersome, but it common to many replica computations for the generalised linear likelihood $P_{\loss}$. We refer the curious reader to Appendix C of \cite{gerace2020generalisation} or to Appendix IV of \cite{aubin2020generalization} for details, and write the final result here:
\begin{align}
\Psi_{y}\equiv\lim\limits_{r\to 0^{+}}\frac{1}{r}\Psi^{(r)}_{w} = \mathbb{E}_{\xi\sim\mathcal{N}(0,1)}\left[\int_{\mathbb{R}}\dd y~\mathcal{Z}_{0}\left(y,\frac{m}{\sqrt{q}}\xi, \rho-\frac{m^2}{q}\right)\log\mathcal{Z}_{\loss}(y,\sqrt{q}\xi,V)\right]
\end{align}
\noindent where we have defined $V=r-q$ and:
\begin{align}
\mathcal{Z}_{g/0}(y,\omega,V) = \mathbb{E}_{x\sim\mathcal{N}(\omega, V)} \left[P_{g/0}(y|x)\right].
\end{align}
Note that as in \cite{gerace2020generalisation}, the consistency condition of the zeroth order term in the free energy fix $\rho = \mathbb{E}_{\vec{\theta}_{0}}\left[\frac{1}{\tdim}{\vec{\theta}_{0}}^{\top}\Psi\vec{\theta}_{0}\right]$ and $\hat{\rho} = 0$. On the other hand, the limit of the prior term here is exactly as the one discussed in Appendix C of \cite{goldt2020gaussian}, and is given by:
\begin{align}
\Psi_{w}\equiv \lim\limits_{r\to 0^{+}}\frac{1}{r}\Psi_{w}^{(r)} = \frac{1}{\sdim}\mathbb{E}_{\xi,\vec{\theta}_{0}}\log\int_{\mathbb{R}^{\sdim}}P_{w}\left(\dd\vec{w}\right) e^{-\frac{\hat{V}}{2}\vec{w}^{\top}\Omega\vec{w}+\vec{w}^{\top}\left(\hat{m}\Phi^{\top}\vec{\theta}_{0}+\hat{q}\Omega^{1/2}\vec{\xi}\right)}.
\end{align}

%%%%%%%%%%%%%%%%%%%%%%%%
\subsection*{Summary}
%%%%%%%%%%%%%%%%%%%%%%%%
The replica symmetric free energy density is simply given by:
\begin{align}
f_{\beta} = \underset{q,m,\hat{q},\hat{m}}{\extr}~\left\{-\frac{1}{2}r\hat{r}-\frac{1}{2}q\hat{q}+\sqrt{\ar}~m\hat{m}-\alpha \Psi_{y}(r, m,q) -	 \Psi_{w}(\hat{r}, 	\hat{m},\hat{q})\right\}
\label{eq:app:freeen:final}
\end{align}
\noindent where
\begin{align}
\Psi_{w} &= \lim\limits_{\sdim\to\infty} \frac{1}{\sdim}\mathbb{E}_{\xi,\vec{\theta}_{0}}\log\int_{\mathbb{R}^{\sdim}}P_{w}\left(\dd\vec{w}\right) e^{-\frac{\hat{V}}{2}\vec{w}^{\top}\Omega\vec{w}+\vec{w}^{\top}\left(\hat{m}\Phi^{\top}\vec{\theta}_{0}+\hat{q}\Omega^{1/2}\vec{\xi}\right)}\notag\\
\Psi_{y} &= \mathbb{E}_{\xi\sim\mathcal{N}(0,1)}\left[\int_{\mathbb{R}}\dd y~\mathcal{Z}_{0}\left(y,\frac{m}{\sqrt{q}}\xi, \rho-\frac{m^2}{q}\right)\log\mathcal{Z}_{\loss}(y,\sqrt{q}\xi,V)\right]
\end{align}

%%%%%%%%%%%%%%%%%%%%%%%%%%%%%%%%%%%%%%%%%%%%%%%%
\subsection{Ridge regression and fixed weights}
%%%%%%%%%%%%%%%%%%%%%%%%%%%%%%%%%%%%%%%%%%%%%%%%
For an $\ell_{2}$-regularisation term, we have:
\begin{align}
P_{w}(\dd\vec{w}) = \frac{1}{(2\pi)^{\sdim/2}}e^{-\frac{\beta \lambda}{2}||\vec{w}||_{2}^2}	\dd\vec{w}
\end{align}
\noindent where we have included a convenient constant, and therefore:
\begin{align}
\int_{\mathbb{R}^{\sdim}}P_{w}(\dd\vec{w})&e^{-\frac{\hat{V}}{2}\vec{w}^{\top}\Omega\vec{w}+\vec{w}^{\top}\left(\hat{m}\Phi^{\top}\vec{\theta}_{0}+\sqrt{\hat{q}}\Omega^{1/2}\vec{\xi}\right)} =\int_{\mathbb{R}^{\sdim}}\frac{\dd\vec{w}}{(2\pi)^{p/2}}e^{-\frac{1}{2}\vec{w}^{\top}\left(\beta\lambda\mat{I}_{\sdim}+\hat{V}\Omega\right)\vec{w}+\vec{w}^{\top}\left(\hat{m}\Phi^{\top}\vec{\theta}_{0}+\sqrt{\hat{q}}\Omega^{1/2}\vec{\xi}\right)}\notag\\
&=\frac{\exp\left(\frac{1}{2}\left(\hat{m}\Phi^{\top}\vec{\theta}_{0}+\sqrt{\hat{q}}\Omega^{1/2}\vec{\xi}\right)^{\top}\left(\beta\lambda\mat{I}_{\sdim}+\hat{V}\Omega\right)^{-1}\left(\hat{m}\Phi^{\top}\vec{\theta}_{0}+\sqrt{\hat{q}}\Omega^{1/2}\vec{\xi}\right)^{\top}\right)}{\sqrt{\det\left(\beta\lambda\mat{I}_{\sdim}+\hat{V}\Omega\right)}}
\end{align}
\noindent taking the log and using $\log\det = \tr\log$, up to the limit:
\begin{align}
\Psi_{w} &= \frac{1}{2\sdim}\mathbb{E}_{\xi,\vec{\theta}_{0}}\left[\left(\hat{m}\Phi^{\top}\vec{\theta}_{0}+\sqrt{\hat{q}}\Omega^{1/2}\vec{\xi}\right)^{\top}\left(\beta\lambda\mat{I}_{\sdim}+\hat{V}\Omega\right)^{-1}\left(\hat{m}\Phi^{\top}\vec{\theta}_{0}+\sqrt{\hat{q}}\Omega^{1/2}\vec{\xi}\right)\right]\notag\\
&\qquad-\frac{1}{2\sdim}	\tr\log \left(\beta\lambda\mat{I}_{\sdim}+\hat{V}\Omega\right)
\end{align}
Defining the shorthand $\mat{A} = \left(\beta\lambda\mat{I}_{\sdim}+\hat{V}\Omega\right)^{-1}$, we can now take the averages over $\xi$ explicitly:
\begin{align}
	\mathbb{E}_{\vec{\xi}}\left[\left(\hat{m}\Phi^{\top}\vec{\theta}_{0}+\sqrt{\hat{q}}\Omega^{1/2}\vec{\xi}\right)^{\top}\mat{A}\left(\hat{m}\Phi^{\top}\vec{\theta}_{0}+\hat{q}\Omega^{1/2}\vec{\xi}\right)\right] = \hat{m}^{2}{\vec{\theta}_{0}}^{\top}\Phi\mat{A}\Phi^{\top}\vec{\theta}_{0}+\hat{q}\tr\Omega^{1/2}\mat{A} \Omega^{1/2}
\end{align}
Putting together, up to the limit:
\begin{align}
\Psi_{w} &=	-\frac{1}{2\sdim}	\tr\log \left(\beta\lambda\mat{I}_{\sdim}+\hat{V}\Omega\right)+\frac{1}{2\sdim}\tr\left(\hat{m}^{2}\Phi^{\top}\vec{\theta}_{0}\vec{\theta}_{0}^{\top}\Phi+\hat{q}\Omega\right)\left(\beta\lambda\mat{I}_{\sdim}+\hat{V}\Omega\right)^{-1} 
\end{align}

%%%%%%%%%%%%%%%%%%%%%%%%%%%%%%%%%%%%%%%%%%%%%%%
\subsection[Taking the zero temperature limit]{Taking the $\beta\to\infty$ limit}
%%%%%%%%%%%%%%%%%%%%%%%%%%%%%%%%%%%%%%%%%%%%%%%%
Finally, in order to take the $\beta\to\infty$ limit explicitly, we note that under the rescaling 
\begin{align}
V\to \beta^{-1} V && q\to q && m\to m\notag\\
\hat{V}\to \beta\hat{V} && \hat{q}\to \beta^2\hat{q} && \hat{m}\to \beta\hat{m}.
\end{align}
The potential $\Psi_{w}$ has a trivial limit:
\begin{align}
\lim\limits_{\beta\to\infty}\frac{1}{\beta}\Psi_{w} &=	-\frac{1}{2\sdim}	\tr\log \left(\lambda\mat{I}_{\sdim}+\hat{V}\Omega\right)+\frac{1}{2\sdim}\tr\left(\hat{m}^{2}\Phi^{\top}\vec{\theta}_{0}\vec{\theta}_{0}^{\top}\Phi+\hat{q}\Omega\right)\left(\lambda\mat{I}_{\sdim}+\hat{V}\Omega\right)^{-1} 
\end{align}
\noindent while $\Psi_{y}$ requires more attention. Since $\mathcal{Z}_{0}$ only depends on $(q,m)$, it is invariant under the rescaling. On the other hand, we have that:
\begin{align}
    \mathcal{Z}_{g}(y,\sqrt{q}\xi,V) = \sqrt{\beta}\int\frac{\dd x}{\sqrt{2\pi V}}e^{-\beta\left[\frac{(x-\sqrt{q}\xi)^2}{2V}+\loss(y,x)\right]}\underset{\beta\to\infty}{=} e^{-\beta \mathcal{M}_{V\loss(y,\cdot)}(\sqrt{q}\xi)}
\end{align}
\noindent where $\mathcal{M}$ is the Moreau envelope associated to the loss $\loss$:
\begin{align}
    \mathcal{M}_{\tau\loss(y,\cdot)}(x) = \underset{z\in\mathbb{R}}{\inf}\left[\frac{(z-x)^2}{2\tau}+\loss(y,z)\right]
\end{align}
\noindent and therefore:
\begin{align}
    \lim\limits_{\beta\to\infty}\frac{1}{\beta} \Psi_{y} = - \mathbb{E}_{\xi\sim\mathcal{N}(0,1)}\left[\int\dd y~\mathcal{Z}_{0}\left(y,\frac{m}{\sqrt{q}}\xi, \rho-\frac{m^2}{q}\right)~ \mathcal{M}_{V\loss(y,\cdot)}\left(\sqrt{q}\xi\right)\right]
\end{align}
The zero temperature therefore is simply given by:
\begin{align}
\lim\limits_{\beta\to\infty }f_{\beta} = \underset{V,q,m,\hat{V},\hat{q},\hat{m}}{\extr}&~\left\{-\frac{1}{2}\left(q\hat{V}-\hat{q}V\right)+\sqrt{\ar}~m\hat{m}+\alpha \mathbb{E}_{\xi\sim\mathcal{N}(0,1)}\left[\int\dd y~\mathcal{Z}_{0}~ \mathcal{M}_{V\loss(y,\cdot)}\right]\right.\notag\\
&\qquad\left.-\frac{1}{2\sdim}\tr\left(\hat{m}^{2}\Phi^{\top}\vec{\theta}_{0}\vec{\theta}_{0}^{\top}\Phi+\hat{q}\Omega\right)\left(\lambda\mat{I}_{\sdim}+\hat{V}\Omega\right)^{-1}\right\}\label{eq:app:freeen:zerot}
\end{align}
%%%%%%%%%%%%%%%%%%%%%%%%%%%%%%%%%%%%%%%%%%%%%%%%%%%%%%%%%%%%%
\subsection{Saddle-point equations}
%%%%%%%%%%%%%%%%%%%%%%%%%%%%%%%%%%%%%%%%%%%%%%%%%%%%%%%%%%%%%
To solve the extremisation problem defined by eq.~\eqref{eq:app:freeen:zerot}, we search for vanishing gradient points of the potential. This lead to a set of self-consistent \emph{saddle-point} equations:
\begin{align}
	\begin{cases}
		\hat{V} = -\alpha\mathbb{E}_{\xi}\left[\int_{\mathbb{R}}\dd y~\mathcal{Z}_{0} ~\partial_{\omega}f_{g}\right]\\
		\hat{q} = \alpha \mathbb{E}_{\xi}\left[\int_{\mathbb{R}}\dd y~\mathcal{Z}_{0} f_{g}^2\right]\\
		\hat{m} = \frac{\alpha}{\sqrt{\ar}} \mathbb{E}_{\xi}\left[\int_{\mathbb{R}}\dd y~\partial_{\omega}\mathcal{Z}_{0}~f_{g}  \right]
	\end{cases} && 
	\begin{cases}
		V =  \frac{1}{\sdim}\tr\left(\lambda\mat{I}_{\sdim}+\hat{V}\Omega\right)^{-1}\Omega\\
		q = \frac{1}{\sdim}\tr\left[\left(\hat{q}\Omega+\hat{m}^{2}\Phi^{\top}\vec{\theta}_{0}\vec{\theta}_{0}^{\top}\Phi\right)\Omega\left(\lambda\mat{I}_{\sdim}+\hat{V}\Omega\right)^{-2}\right]\\
		m= \frac{1}{\sqrt{\ar}}\frac{\hat{m}}{p}\tr \Phi^{\top}\vec{\theta}_{0}\vec{\theta}_{0}^{\top}\Phi\left(\lambda\mat{I}_{\sdim}+\hat{V}\Omega\right)^{-1}
	\end{cases}
	\label{eq:app:sp}
\end{align}
\noindent where $f_{g}(y,\omega,V) = -\partial_{\omega}\mathcal{M}_{V\loss(y,\cdot)}(\omega)$, which can also be obtained from the proximal operator
\begin{align}
    \prox_{V\loss(y, \cdot)}(\omega) = \underset{z\in\mathbb{R}}{\argmin}\left[\frac{(z-\omega)^2}{2V} +\loss\left(y,z\right)\right]
    \label{eq:app:prox}
\end{align}
\noindent using the envelope theorem $\mathcal{M}'_{\tau f}(x) = \tau^{-1}\left(x - \prox_{\tau f}(x)\right)$. A python implementation of the saddle-point equations for the losses discussed below is available in~\url{https://github.com/IdePHICS/GCMProject} %\cite{ourgit}. 

\subsection{Examples}
\label{sec:app:replicas:egs}
We now discuss a couple of examples in which the equations above simplify. 

\paragraph{Ridge regression: } Consider a ridge regression task with $f_{0}(x)=\hat{f}(x)=x$, loss $\loss(y,x) = \frac{1}{2}(y-x)^2$ and choose $\hat{g}(y,x) = \frac{1}{2}(y-x)^2$. In this case, our model is closely related to the mismatched models in \cite{hastie2019surprises} and \cite{ghorbani2020neural}. In the first, labels are generated in a higher-dimensional space which contains the features as a subspace, and can be mapped to our model in the case $p>d$ by defining the projection of the teacher weights in the student space $\Phi^{\top}\vec{\theta}_{0}\in\mathbb{R}^{\sdim}$ and its orthogonal complement $(\Phi^{\top}\vec{\theta}_{0})^{\perp}\in\mathbb{R}^{p-d}$. In the second, the teacher acts on an orthogonal subset of the features, and can be mapped with a similar construction to our model in the case $p<d$. These two cases were studied for specific linear tasks, such as ridge and random features regression, with the covariances modelling structure in the data. Conceptually, our model differs slightly in the sense that any additional fixed feature layer, e.g. random projections or a pre-trained feature map, is also contained in the convariances.

For the linear task, the asymptotic training and generalisation errors read:
\begin{align}
  \mathcal{E}^{\star}_{\rm{train.}} = \frac{\rho + q^{\star} - 2m^{\star}}{(1+V^{\star})^2},   && \mathcal{E}^{\star}_{\rm{gen.}} = \rho + q^{\star} - 2m^{\star}
\end{align}
\noindent where $\rho = \frac{1}{\tdim}\vec{\theta}_{0}^{\top}\Psi\vec{\theta}_{0}$ and $(V^{\star}, q^{\star}, m^{\star})$ are the fixed point of the following set of self-consistent equations:
\begin{align}
    \begin{cases}
        \hat{V} = \frac{\alpha}{1+V}\\
        \hat{q} = \alpha\frac{\rho+q-2m}{(1+V)^2}\\
        \hat{m} = \frac{1}{\sqrt{\gamma}}\frac{\alpha}{1+V}
    \end{cases}, && 
    \begin{cases}
		V =  \frac{1}{\sdim}\tr\left(\lambda\mat{I}_{\sdim}+\hat{V}\Omega\right)^{-1}\Omega\\
		q = \frac{1}{\sdim}\tr\left[\left(\hat{q}\Omega+\hat{m}^{2}\Phi^{\top}\vec{\theta}_{0}\vec{\theta}_{0}^{\top}\Phi\right)\Omega\left(\lambda\mat{I}_{\sdim}+\hat{V}\Omega\right)^{-2}\right]\\
		m=\frac{1}{\sqrt{\ar}}\frac{\hat{m}}{\sdim}\tr \Phi^{\top}\vec{\theta}_{0}\vec{\theta}_{0}^{\top}\Phi\left(\lambda\mat{I}_{\sdim}+\hat{V}\Omega\right)^{-1}
	\end{cases}.
	\label{eq:app:ridge}
\end{align}
Note that quite interestingly we have the following relationship between the training and generalisation error:
\begin{align}
    \mathcal{E}^{\star}_{\rm{train.}} = \frac{\mathcal{E}^{\star}_{\rm{gen.}}}{(1+V^{\star})^2}.
\end{align}
This give us an interesting interpretation of $V^{\star}$ as parametrising the variance gap between the generalisation and training error\footnote{We thank St\'ephane d'Ascoli for bringing this relation to our attention.}. 
In particular, note that $V^{\star}$ only depends on the spectrum of the population covariance, since it is the solution of:
\begin{align}
    V = \int\nu_{\Omega}(\dd\omega)\frac{\omega}{\lambda+\frac{\alpha \omega}{1+V}}
\end{align}
\noindent where $\nu_{\Omega}$ is the spectral density of $\Omega$.
\subsection*{Binary classification}
For a binary classification task, we tak $f_{0}(x) = \hat{f}(x) = \rm{sign}(x) \in\{-1,1\}$. Our equations generalise the ones derived \cite{huang2020large} in the specific case of $d=p$ and $\Psi = \mat{I}_{\sdim}$, $\vec{\theta}_{0}\sim\mathcal{N}(\vec{0},\mat{I}_{\sdim})$. For binary classification, the asymptotic classification error $\mathcal{E}_{\rm{gen.}}(\hat{w}) = \mathbb{P}\left(y\neq\rm{sign}(\hat{w}^{\top}\vec{u})\right)$ can be explicitly writen is terms of the overlaps as: 
\begin{align}
    \mathcal{E}^{\star}_{\rm{gen.}} = \frac{1}{\pi} \cos^{-1}\left(\frac{m^{\star}}{\sqrt{\rho q^{\star}}}\right).
\end{align}
\noindent where again $(q^{\star}, m^{\star})$ are solutions of the self-consistent saddle-point equations. The teacher measure is given by:
\begin{align}
    \mathcal{Z}_{0}(y,\omega,V) = \frac{\delta_{y,1}+\delta_{y,-1}}{2}\left(1+\erf\left(\frac{y\omega}{\sqrt{2V}}\right)\right)
\end{align}

The explicit form of the equation depends on the choice of the loss function, three of which are of particular interest:
\paragraph{Square-loss: } As in the ridge case, for $\loss(y,x) = \frac{1}{2}(y-x)^2$ the saddle-point equations simplify considerably:
\begin{align}
    \begin{cases}
        \hat{V} = \frac{\alpha}{1+V}\\
        \hat{q} = \alpha\frac{1+q-2m\sqrt{\frac{2}{\pi\rho}}}{(1+V)^2}\\
        \hat{m} = \sqrt{\frac{2}{\pi\rho}}\frac{\alpha}{1+V}
    \end{cases}, && 
    \begin{cases}
		V =  \frac{1}{\sdim}\tr\left(\lambda\mat{I}_{\sdim}+\hat{V}\Omega\right)^{-1}\Omega\\
		q = \frac{1}{\sdim}\tr\left[\left(\hat{q}\Omega+\hat{m}^{2}\Phi^{\top}\vec{\theta}_{0}\vec{\theta}_{0}^{\top}\Phi\right)\Omega\left(\lambda\mat{I}_{\sdim}+\hat{V}\Omega\right)^{-2}\right]\\
		m=\frac{1}{\sqrt{\ar}}\frac{\hat{m}\gamma}{\sdim}\tr \Phi^{\top}\vec{\theta}_{0}\vec{\theta}_{0}^{\top}\Phi\left(\lambda\mat{I}_{\sdim}+\hat{V}\Omega\right)^{-1}
	\end{cases}.
\end{align}
Similarly, the asymptotic training error also admits a simple expression:
\begin{align}
    \mathcal{E}^{\star}_{\rm{train.}} =\frac{1}{4} \frac{1+q^{\star}-2m^{\star}\sqrt{\frac{2}{\pi\rho}}}{(1+V^{\star})^{2}}
\end{align}
\paragraph{Logistic regression: } Different from the previous cases, for logistic loss $\loss(y,x) = \log\left(1+e^{-yx}\right)$ the equations for $(\hat{V},\hat{q},\hat{m})$ cannot be integrated explicitly, since the proximal operator doesn't admit a closed form solution. Instead, $f_{g}$ can be found by solving the following self-consistent equation:
\begin{align}
    f_{g} = \frac{y}{1+e^{y(Vf_{g}+\omega)}}.
\end{align}
% Note that this also allow us to express $\partial_{\omega}f_{g} = \left(2\cosh(\frac{1}{2}y(Vf_{g}+\omega))\right)^{-2}$. 

\paragraph{Soft-margin regression: } Another useful case in which the proximal operator has a closed form solution is for the hinge loss $\loss(y,x)=\rm{max}(0, 1-yx)$. In this case:
\begin{align}
f_{g}(y,\omega, V) = \begin{cases}
	y & \text{ if } \omega y < 1-V\\
	\frac{y-\omega}{V} & \text{ if } 1-V < \omega y < 1\\
	0 & \text{ otherwise }\\
\end{cases}, && 
\partial_{\omega}f_{g}(y,\omega, V) = 
\begin{cases}
	-\frac{1}{V} & \text{ if } 1-V < \omega y < 1\\
	0 & \text{ otherwise }\\	
\end{cases}	
\end{align}
Again, the equations cannot be integrated explicitly. Note that in the limit $\lambda \to 0$, both the logistic and soft-margin solutions converge to the max-margin estimator %\cite{NIPS2003_2433}.

\subsection{Relation to previous models}
\label{sec:app:connection}
\paragraph{Random features: } The feature map for random features learning can be written as:
\begin{align}
\Phi_{\mat{F}}: \vec{u}\in\mathbb{R}^{\tdim} \mapsto \vec{v} = \sigma\left(\frac{1}{\sqrt{k}}\mat{F}\vec{u}\right)\in\mathbb{R}^{\sdim}
\end{align}
\noindent where $\vec{u}\in\mathbb{R}^{\tdim}$ is the original data, $\mat{F}\in\mathbb{R}^{d\times k}$ is a chosen random projection matrix and $\sigma:\mathbb{R}\to\mathbb{R}$ is a chosen non-linearity acting component-wise in $\mathbb{R}^{\sdim}$, see \cite{rahimi2008random}. Random features learning has attracted a lot of interest recently, and has been studied in \cite{mei2019generalization, gerace2020generalisation, huang2020large, dhifallah2020precise} in the case of Gaussian data $\vec{u}\sim\mathcal{N}(\vec{0},\mat{I}_{\tdim})$. Our model encompasses all of these works, and in the case of Gaussian data the covariancec $(\Psi, \Omega, \Phi)$ can be explicitly related to the projection matrix $\mat{F}$:
\begin{align}
    \Psi = \mat{I}_{\tdim}, && \Phi = \kappa_{1}\mat{F}, && \Omega = \kappa_{0}^2\vec{1}_{\sdim}\vec{1}_{\sdim}^{\top}+\kappa_{1}^2\frac{\mat{F}\mat{F}^{\top}}{d}+\kappa_{\star}^{2}\mat{I}_{\sdim}
\end{align}
\noindent where $\vec{1}_{\sdim}\in\mathbb{R}^{\sdim}$ is the all-ones vector and the constants $(\kappa_{0},\kappa_{1}, \kappa_{\star})$ are related to $\sigma$ as:
\begin{align}
    \kappa_{0} = \mathbb{E}_{z\sim\mathcal{N}(0,1)}\left[\sigma(z)\right], && \kappa_{1} = \mathbb{E}_{z\sim\mathcal{N}(0,1)}\left[z\sigma(z)\right], && \kappa_{\star} = \sqrt{ \mathbb{E}_{z\sim\mathcal{N}(0,1)}\left[\sigma(z)^2\right]-\kappa_{0}^2 -\kappa_{1}^2}
\end{align}
These relations hold asymptotically, and rely on the \emph{Gaussian equivalence theorem} (GET), see \cite{goldt2020gaussian} for a proof.

\paragraph{Generative models: } In \cite{goldt2020gaussian}, a similar Gaussian covariate model was used to study the performance of random feature regression on data generated from pre-trained generative models:
\begin{align}
    \vec{v} = \mathcal{G}(\vec{u})\in\mathbb{R}^{\sdim}, && \vec{u}\sim\mathcal{N}(\vec{0},\mat{I}_{\tdim}).
\end{align}
\noindent where $\mathcal{G}:\mathbb{R}^{\tdim}\to\mathbb{R}^{\sdim}$ is a generative network mapping the latent space $\mathbb{R}^{\tdim}$ to the input space $\mathbb{R}^{\sdim}$ (e.g. a pre-trained GAN). Labels were generated directly in the latent space $\mathbb{R}^{\tdim}$ using a generalised linear model on random weights: $y=f_{0}\left(\vec{\theta}_{0}^{\top}\vec{u}\right)$ with $\vec{u}\sim\mathcal{N}(\vec{0},\mat{I}_{\tdim})$. A \emph{Gaussian Equivalence Principle} (GEP) stating that the asymptotic generalisation and training performances of this model are fully captured by second order statistics was conjectured and shown to hold numerically for different choices of generative models $\mathcal{G}$. Indeed, this model is a particular case of ours when $\Psi = \mat{I}_{\tdim}$ and $\vec{\theta}_{0}\sim\mathcal{N}(\vec{0},\mat{I}_{\tdim})$. Assuming that the GEP holds, our model therefore can be seen as a generalisation of \cite{goldt2020gaussian} to structured teachers. For instance, in Section \ref{sec:main:gan} of the main we show several cases in which the teacher $\vec{u} = \tilde{\mathcal{G}}(\vec{c})$ for a latent vector $\vec{c}\sim\mathcal{N}(\vec{0}, \mat{I}_{k})$ and a pre-trained map $\tilde{\mathcal{G}}$ that can include a generative model and a fixed feature map (e.g. random features, scattering transform, pre-learned neural network, etc.). Also, it is important to stress that our model also account for the case in which the teacher weights $\vec{\theta}_{0}\in\mathbb{R}^{\tdim}$ are fixed, and therefore can be also learned.

\paragraph{Kernel methods: } Let $\mathcal{H}$ be a Kernel Reproducing Hilbert space (RKHS) associated to a given kernel $K$ and $\mathcal{D} = \{\vec{x}^{\mu}, y^{\mu}\}_{\mu=1}^{n}$ be a labelled data set with $\vec{x}\sim p_{x}$ independently, and set $\mathcal{X} = \rm{supp} (p_{x})$. In Kernel regression, the aim is to solve:
\begin{align}
    \underset{f\in\mathcal{H}}{\min}\left[\frac{1}{2}\sum\limits_{\mu=1}^{n}\left(y^{\mu}-f(\vec{x}^{\mu})\right)^2+\frac{\lambda}{2}||f||^2_{\mathcal{H}}\right]\label{eq:app:kernelprob}
\end{align}
\noindent where $||\cdot||_{\mathcal{H}}$ is the norm induced by the scalar product in $\mathcal{H}$. An alternative representation of this problem is given by the feature decomposition of the kernel given by Mercer's theorem:
\begin{align}
    K(\vec{x}, \vec{x}') = \sum\limits_{i=1}^{\infty}\omega_{i} e_{i}(\vec{x}')e_{i}(\vec{x})
\end{align}
\noindent where $\omega_{i}$ and $e_{i}(\vec{x})$ are the eigenvalues and eigenvectors associated with the kernel:
\begin{align}
    \int_{\mathbb{R}^{k}}p_{x}(\dd\vec{x}')K(\vec{x}, \vec{x}')e_{i}(\vec{x}') = \omega_{i}e_{i}(\vec{x})
\end{align}
Note that $\{e_{i}(\vec{x})\}_{i}^{\infty}$ form an orthonormal basis of the space of square-integrable functions $L^{2}\left(\mathcal{X}\right)$ (with respect to the standard scalar product of $L^2$). It is also convenient to define the feature map $\varphi_{i}(\vec{x}) = \sqrt{\omega_{i}}e_{i}(\vec{x})$, which is an orthonormal basis of $\mathcal{H}\subset L^{2}(\mathcal{X})$ (with respect to the scalar product induced by $K$). Therefore, if we assume that the labels $y^{\mu} = f_{0}(\vec{x}^{\mu})$ are generated from a ground truth target function (not necessarely part of $\mathcal{H}$), we can expand both $f$ and $f_{0}$ the feature basis:
\begin{align}
    f(\vec{x}) = \sum\limits_{i=1}^{\infty}w_{i}\varphi_{i}(\vec{x}), && f_{0}(\vec{x}) = \sum\limits_{i=1}^{\infty}\theta_{0i}\varphi_{i}(\vec{x})\label{eq:app:featurebasis}
\end{align}
Note that $f\in\mathcal{H}$ implies that for this sum to make sense $w_{i}$ needs to decay fast enough with respect to $\sqrt{w_{i}}$, but in general we can have $f_{0}\notin \mathcal{H}$ meaning that $\theta_{0}^{i}$ decays slower than $\sqrt{\omega_{i}}$ but still fast enough such that $f_{0}\in L^{2}(\mathcal{X})$. If the number of features is finite ($\omega_{i} = 0$ for $i\geq d$) or if we introduce a cut-off $d\gg n, |\mathcal{X}|$, the representation in the feature basis in eq.~\eqref{eq:app:featurebasis} allow us to rewrite Kernel regression problem in eq.~\eqref{eq:app:kernelprob} simply as ridge regression in feature space:
\begin{align}
    \underset{\vec{w}\in\mathbb{R}^{d}}{\min}\left[\frac{1}{2}\sum\limits_{\mu=1}^{n}\left(\vec{\theta}^{\top}_{0}\vec{\varphi}(\vec{x})-\vec{w}^{\top}\vec{\varphi}(\vec{x})\right)^2+\frac{\lambda}{2}||\vec{w}||_{2}^2\right].
\end{align}
Letting $\vec{v}=\vec{\varphi}(\vec{x})\in\mathbb{R}^{\sdim}$, this formulation is equivalent to our model with $p=d$ and covariance matrices given by:
\begin{align}
    \Psi = \Phi = \Omega = \rm{diag}(\omega_{i}).
\end{align}
Indeed, inserting this expression equation \eqref{eq:app:ridge}:
\begin{align}
    \begin{cases}
        \hat{V} = \hat{m} = \frac{\alpha}{1+V}\\
        \hat{q} = \alpha\frac{\rho+q-2m}{(1+V)^2}\\
    \end{cases}, && 
    \begin{cases}
		V =  \frac{1}{\sdim}\sum\limits_{i=1}^{d}\frac{\omega_{i}}{\lambda + \hat{V}\omega_{i}}\\
		q = \frac{1}{\sdim}\sum\limits_{i=1}^{d}\frac{\hat{q}\omega_{i}^2+\theta_{0i}^2\omega_{i}^3 \hat{m}^2}{(\lambda+\hat{V}\omega_{i})^2}\\
		m=\frac{\hat{m}}{\sdim}\sum\limits_{i=1}^{d}\frac{\omega_{i}^2\theta_{0i}^2}{\lambda+\hat{V}\omega_{i}}
	\end{cases}.
	\label{eq:app:kernel}
\end{align}
\noindent and making a change of variables $\hat{q}\leftarrow\hat{q}\frac{\sdim^2}{n}$, $\hat{m}\leftarrow\hat{m}\frac{\sdim}{n}$, $\hat{V}\leftarrow\hat{q}\frac{\sdim}{n}$, $\rho\leftarrow\sdim \rho$, $m\leftarrow\sdim m$, $q\leftarrow\sdim q$, $\lambda\leftarrow\sdim \lambda$ we recover exactly the self-consistent equations of \cite{bordelon2020} for the performance of kernel ridge regression directly from our equations. 
%In Appendix \ref{scaling} we show how to recover the results of \cite{bordelon2020} and \cite{spigler2019asymptotic} for the power law scaling of the generalisation error in kernel ridge regression. 
Moreover, our model allow to generalise this discussion to more involved kernel tasks such as kernel logistic regression and support vector machines.

%% file: sections/appendix/proof.tex
This section presents the core technical result of this paper in its full generality, along with the required assumptions and its complete proof. For technical reasons, variables different than the ones appearing in the replica calculation are introduced. The proof is nonetheless presented in a self-contained way and the relation with the replica variables are given in appendix \ref{app-mapping}, eq.(\ref{match-rep-gordon}). We start by reminding the formulation of the problem. Consider the matrices $U \in \mathbb{R}^{n \times p}$ of concatenated vectors $\mathbf{u}$ used by the teacher and $\mathcal{V} \in \mathbb{R}^{n \times d}$ the corresponding one for the student.
The estimator may now be defined using potentially non-separable functions:
  \begin{equation}
\label{eq:full-student}
    \hat{\mathbf{w}} = \argmin_{\mathbf{w}\in \mathbb{R}^{d}}
    \left[
    g\left(\frac{1}{\sqrt{d}}\mathcal{V}\mathbf{w},\mathbf{y}\right)+r(\mathbf{w})
    \right]\, ,
\end{equation}
where the function $g:\mathbb{R}^{n}\to \mathbb{R}$.
The training and generalization errors are reminded as: 
\begin{align}
    \mathcal{E}_{\rm train}(\mathbf{w}) &\equiv 
    \frac{1}{n} \mathbb{E}\left[g\left( \frac{1}{\sqrt{d}}\mathcal{V}\mathbf{w},\mathbf{y}\right)+\reg\left(\mathbf{w}\right)\right] \\
    \mathcal{E}_{\rm gen}(\mathbf{w}) 
    &\equiv
    \mathbb{E}\left[\hat{g}(\hat{f}\left(\mathbf{v}_{\rm{new}}^{\top}\mathbf{w}), y_{\rm{new}}\right)\right] \equiv
    \mathbb{E}\left[\hat{g}\left(\hat{f}(\mathbf{v}_{\rm{new}}^{\top}\mathbf{w}), \vec{f}_{0}(\mathbf{u}_{\rm{new}}^{\top}\boldsymbol{\theta}_{0})\right)\right]\, .\label{app-train-gen-err}
\end{align} 
Intuitively, the variables $\mathbf{u}_{\rm{new}}^{\top}\boldsymbol{\theta}_{0}$ and $\mathbf{v}_{\rm{new}}^{\top}\mathbf{w}$ will play a key role in the analysis. Given an instance of $\boldsymbol{\theta}_{0}$ and $\mathbf{w}$, the tuple $\left(\frac{1}{\sqrt{p}}\mathbf{u}_{\rm{new}}^{\top}\boldsymbol{\theta}_{0},\frac{1}{\sqrt{d}}\mathbf{v}_{\rm{new}}^{\top}\mathbf{w}\right)$ is a bivariate Gaussian with covariance:
\begin{equation}
\begin{bmatrix}
\frac{1}{p}\boldsymbol{\theta}_{0}^{\top}\Psi\boldsymbol{\theta}_{0} & \frac{1}{\sqrt{dp}}(\Phi^{\top}\boldsymbol{\theta}_{0})^{\top}\mathbf{w} \\ \frac{1}{\sqrt{dp}}(\Phi^{\top}\boldsymbol{\theta}_{0})^{\top}\mathbf{w} & \frac{1}{d}\mathbf{w}^{\top}\Omega\mathbf{w}
\end{bmatrix}\, .
\label{def:cov_gauss}
\end{equation}
We thus define the following overlaps, that will play a fundamental role in the analysis:
\begin{align}
    \label{nice-overlaps}
    \rho = \frac{1}{p}\boldsymbol{\theta}_{0}^{\top}\Psi\boldsymbol{\theta}_{0}, && m = \frac{1}{\sqrt{dp}}(\Phi^{\top}\boldsymbol{\theta}_{0})^{\top}\mathbf{w} \, , && q = \frac{1}{d}\mathbf{w}^{\top}\Omega\mathbf{w}, && \chi =  \frac{1}{d}\boldsymbol{\theta}_{0}^{\top}\Phi\Omega^{-1}\Phi^{\top}\boldsymbol{\theta}_{0}\, .
\end{align}
Note that here, we will not introduce the spectral decomposition \ref{eq:main:specdensity} as it will not simplify the expressions as in the $l_{2}$ case. The representations are mathematically equivalent nonetheless.
Our main result is that the distribution of the estimator $\hat{\mathbf{w}}$ can be exactly computed in the weak sense from the solution to six scalar fixed point equations with a unique solution.
%%%%%%%%%%%%%%%%%%%%%%%%%%%%%%%%%%%
\subsection{Necessary assumptions}
\label{main-assumptions}
%%%%%%%%%%%%%%%%%%%%%%%%%%%%%%%%%%%
We start with a list of the necessary assumptions for the most generic version of the result to hold. We also briefly discuss how they are relevant in a supervised machine learning context.
\begin{enumerate}[font={\bfseries},label={(A\arabic*)}]
\item The vector $\vec{\theta}_{0}$ is pulled from any given distribution $p_{\vec{\theta}_{0}} \in \mathbb{R}^{\tdim}$ (this includes deterministic vectors with bounded norm), and is independent of the matrices U and $\mathcal{V}$. Additionally, the signal is non-vanishing and has finite squared norm, i.e. the following holds almost surely: 
\begin{equation}
    \lim_{p \to \infty} 0<\mathbb{E}\left[\frac{\vec{\theta}_{0}^{\top}\vec{\theta}_{0}}{p}\right] < +\infty
\end{equation} 
\item The covariance matrices verify:
\begin{equation}
\left(\Psi,\Omega\right) \in \mathbb{S}_{p}^{++}\times \mathbb{S}_{d}^{++}, \quad \Omega-\Phi^{\top}\Psi^{-1}\Phi \succeq 0
\end{equation}
The spectral distributions of the matrices $\Phi, \Psi$ and $\Omega$ converge to distributions such that the overlaps defined by equation (\ref{nice-overlaps}) are well-defined. Additionally, the maximum singular values of the covariance matrices are bounded with high probability when $n,p,d\to \infty$.
\item  The functions $\reg$ and $\loss$ are proper, lower semi-continuous, convex functions. Additionally, we assume that the cost function $\reg+\loss$ is coercive, i.e.:
\begin{equation}
    \lim_{\norm{\mathbf{w}}_{2} \to +\infty} (\reg+\loss)(\mathbf{w}) = +\infty
\end{equation}
and that the following scaling condition holds : for all $\samples,\sdim \in \mathbb{N}, \mathbf{z} \in \mathbb{R}^{\samples}$ and any constant $c>0$, there exist finite, positive constants $C_{1},C_{2},C_{3}$, such that, for any standard normal random vectors $\mathbf{h} \in \mathbb{R}^{d}$ and $\mathbf{g} \in \mathbb{R}^{n}$:
\begin{align}
\norm{\mathbf{z}}_{2} \leqslant c\sqrt{n} \implies \sup_{\mathbf{x} \in \partial g(\mathbf{z})} \norm{\mathbf{x}}_{2} \leqslant C_{1}\sqrt{n}, && \frac{1}{\sdim}\mathbb{E}\left[\reg(\mathbf{\mathbf{h}})\right] <+\infty, && \frac{1}{n}\mathbb{E}\left[g(\mathbf{g})\right] < +\infty
 \end{align}
\item The random elements of the function $f_{0}$ are independent of the matrices $U$ and $\mathcal{V}$. Additionally the following limit exists and is finite 
\begin{equation}
    \lim_{n \to \infty} \mathbb{E}\left[\frac{1}{n}f_{0}(U\vec{\theta}_{0})^{\top}f_{0}(U\vec{\theta}_{0})\right] < +\infty \notag
\end{equation}
\item When we send the dimensions $n,p,d$ to infinity, they grow with finite ratios $\alpha = \samples/\sdim$, $\gamma = \tdim/\sdim$.
\item \textbf{Additional assumptions for linear finite sample size rates} : the teacher vector $\boldsymbol{\theta}_{0}$ has sub-Gaussian one dimensional marginals. The functions $\reg,\loss,\phi_{1},\phi_{2}$ are pseudo-Lipschitz of finite order. The eigenvalues of the covariance matrices are bounded with probability one.
\item \textbf{Additional assumptions for exponential finite sample size rates}: all of the above, and the loss function $\loss$ is separable and pseudo-Lipschitz of order 2, the regularisation is either a ridge or a Lipschitz function, the functions $\phi_{1},\phi_{2}$ are respectively separable, pseudo-Lipschitz of order 2, and a square or Lipschitz function.
\end{enumerate}

The first assumption (A1) ensures that the teacher distribution is non-vanishing. The positive definiteness in (A2) means the covariance matrices of the blocks U and V are well-specified. Note that the cross-correlation matrix $\Phi$ can have singular values equal to zero. The assumption about the limiting spectral distribution is essentially a summability condition which is immediately verified if the limiting spectral distributions have compact support, a common case. The scaling assumptions from (A3) are natural as they imply that non-diverging inputs result in non-diverging outputs in the functions $f$ and $g$, as well as the sub-differentials. Similar scaling assumptions are encountered in proofs such as \cite{thrampoulidis2018precise}. They also allow to show Gaussian concentration of Moreau envelopes, as we will see in Lemma \ref{conc-Mor}. The \emph{coercivity} assumption is verified in most common machine learning setups : any convex loss with ridge regularisation, or any convex loss that is bounded below with a coercive regularisation (LASSO, elastic-net,...), see Corollary 11.15 from \cite{bauschke2011convex}. Assumption (A4) is a classical assumption of teacher-student setups, where any correlation between the teacher and the student is modeled by the covariance matrices and not by the label generating function $f_{0}$. The summability condition ensures generalization error is well-defined for squared performance measures. Finally, (A5) is the typical \emph{high-dimensional} limit used in statistical physics of learning, random matrix theory and a large recent body of work in high-dimensional statistical learning.
\subsection{Main theorem}
First, let's define quantities and a scalar optimization problem that will be used to state the asymptotic behaviour of (\ref{teacher}-\ref{eq:student}):
\begin{definition}(Scalar potentials/replica free energy)
\label{scalar-potentials}
Define the following functions of the scalar variables $\tau_{1}>0,\tau_{2}>0,\kappa\geqslant0,\eta\geqslant 0,\nu,m$:
\begin{align}
    \mathcal{L}_{g}(\tau_{1},\kappa,m,\eta) &= \frac{1}{n}\mathbb{E}\left[\mathcal{M}_{\frac{\tau_{1}}{\kappa}g(.,\mathbf{y})}\left(\frac{m}{\sqrt{\rho}}\mathbf{s}+\eta\mathbf{h}\right)\right] \label{inter-quant1}\, , \\
    \mathcal{L}_{\reg}(\tau_{2},\eta,\nu,\kappa) &= \frac{1}{d}\mathbb{E}\left[\mathcal{M}_{\frac{\eta}{\tau_{2}}\reg(\Omega^{-1/2}.)}\left(\frac{\eta}{\tau_{2}}(\nu\mathbf{t}+\kappa\mathbf{g})\right)\right]\, , \nonumber
    \end{align}
    where $\mathbf{s},\mathbf{h}\sim\mathcal{N}(0,\mat{I}_{\samples})$ and $\mathbf{g}\sim \mathcal{N}(0,\mat{I}_{\sdim})$ are random vectors independent of the other quantities, $\mathbf{t} = \Omega^{-1/2}\Phi^{\top}\vec{\theta}_{0}$, $\mathbf{y} = \vec{f}_{0}\left(\sqrt{\rho}\mathbf{s}\right)$, and $\mathcal{M}$ denotes the Moreau envelope of a target function. 
    
    From these quantities define the following potential:
    \begin{align}
    \label{free-energy}
    &\mathcal{E}(\tau_{1},\tau_{2},\kappa,\eta,\nu,m) =  \frac{\kappa\tau_{1}}{2}-\frac{\eta\tau_{2}}{2}+m\nu\sqrt{\gamma}-\frac{\tau_{2}}{2\eta}\frac{m^{2}}{\rho}\notag\\&-\frac{\eta}{2\tau_{2}}(\nu^{2}\chi+\kappa^{2})+\alpha\mathcal{L}_{g}(\tau_{1},\kappa,m,\eta)+\mathcal{L}_{\reg}(\tau_{2},\eta,\nu,\kappa)\, .
\end{align}
Under Assumption (\ref{main-assumptions}), the previously defined quantities all admit finite limits when $\samples,\tdim,\sdim \to \infty$.
\end{definition}
\emph{Proof}: This
follows directly from Lemma \ref{conc-Mor}.

The next lemma characterizes important properties of the "potential" function $\mathcal{E}(\tau_{1},\tau_{2},\kappa,\eta,\nu,m)$:
\begin{lemma}(Geometry and minimizers of $\mathcal{E}$)
\label{uniqueness}
 The function $\mathcal{E}(\tau_{1},\tau_{2},\kappa,\eta,\nu,m)$ is jointly convex in $(m,\eta,\tau_{1})$ and jointly concave in $(\nu,\kappa,\tau_{2})$, and the optimization problem 
 \begin{equation}
    \label{optminmax}
    \min_{m,\eta,\tau_{1}}\max_{\kappa,\nu,\tau_{2}} \mathcal{E}(\tau_{1},\tau_{2},\kappa,\eta,\nu,m)
\end{equation}
has a unique solution $(\tau_{1}^{*},\tau_{2}^{*},\kappa^{*},\eta^{*},\nu^{*},m^{*})$ on $\mbox{dom}(\mathcal{E})$.
\end{lemma}
\emph{Proof}: see Appendix \ref{proof-uniqueness}. 
The optimality condition of problem (\ref{optminmax}) yields the set of self-consistent fixed point equations given in Lemma \ref{lemma:fixed-point-eq} of Appendix \ref{main-proof}. Finally, define the following variables:
\begin{align}
\label{distrib-equiv}
    \mathbf{w}^{*} = \Omega^{-1/2}\mbox{\rm prox}_{\frac{\eta^{*}}{\tau_{2}^{*}}\reg(\Omega^{-1/2}.)}\left(\frac{\eta^{*}}{\tau_{2}^{*}}(\nu^{*}\mathbf{t}+\kappa^{*}\mathbf{g})\right) \, , && \mathbf{z}^{*} = \mbox{\rm prox}_{\frac{\tau_{1}^{*}}{\kappa^{*}}g(.,\mathbf{y})}\left(\frac{m^{*}}{\sqrt{\rho}}\mathbf{s}+\eta^{*}\mathbf{h}\right)\, .
\end{align}
where $\mbox{prox}$ denotes the proximal operator.
With these definitions, we can now state our main result:
\begin{theorem}(Training loss and generalisation error)
\label{Train_gen}
 Under Assumption (\ref{main-assumptions}), there exist constants $C,c,c'>0$ such that, for any optimal solution $\hat{\mathbf{w}}$ to (\ref{eq:student}), the training loss and generalisation error defined by equation  verify, for any $0<\epsilon<c'$:
\begin{align}
\label{eq:errors}
    &\mathbb{P}\left(\abs{\mathcal{E}_{\rm train}(\hat{\mathbf{w}})-\mathcal{E}_{\rm train}^{*}}\geqslant\epsilon\right) \leqslant \frac{C}{\epsilon^{2}}e^{-cn\epsilon^{4}} \, , \\
    &\mathbb{P}\left(\abs{\mathcal{E}_{\rm gen}(\hat{\mathbf{w}})-\mathbb{E}_{\omega,\xi}\left[\hat{g}(f_{0}(\omega),\hat{f}(\xi))\right]}\geqslant\epsilon \right)\leqslant\frac{C}{\epsilon^{2}}e^{-cn\epsilon^{4}}\, ,\notag
\end{align}
where $\mathcal{E}_{train}^{*}$ is defined as follows:
\begin{align}
\label{eq:train:asym}
    &\mathcal{E}_{\rm train}^{*} \!= \! \frac{1}{\samples}\mathbb{E}\left[g\left(\mathbf{z}^{*},\mathbf{y}\right)\right]+\frac{1}{\alpha \sdim}\mathbb{E}\left[\reg\left(\mathbf{w}^{*}\right)\right],
\end{align}
 and the random variables $(\omega,\xi)$ are jointly Gaussian with covariance
\begin{equation}
 (\omega,\xi) \sim \mathcal{N}\!\left(0,\begin{bmatrix}\rho&m^{*} \\m^{*} & q^{*}\end{bmatrix}\right),  \thickspace q^{*}\!=\!(\eta^{*})^{2}\!+\!\frac{(m^{*})^{2}}{\rho}\, .
 \label{eq:def:q}
 \end{equation}
\end{theorem}
\emph{Proof}: see Appendix \ref{proof-connect}. Note that the regularisation may be removed to evaluate the training loss.
A more generic result, aiming directly at the estimator $\hat{\mathbf{w}}$, can also be stated:
\begin{theorem}
\label{main-th}
Under Assumption (\ref{main-assumptions}), for any optimal solution $\hat{\mathbf{w}}$ to $(\ref{eq:student})$, denote $\hat{\mathbf{z}} =\frac{1}{\sqrt{\sdim}}\mathcal{V}\hat{\mathbf{w}}$. Then, there exist constants $C,c,c'>0$ such that, for any Lipschitz function $\phi_{1}:\mathbb{R}^{\sdim}\to \mathbb{R}$, and separable, pseudo-Lipschitz function $\phi_{2}:\mathbb{R}^{\samples}\to\mathbb{R}$ and any $0<\epsilon<c'$:
\begin{align}
    &\mathbb{P}\left(\abs{\phi_{1}(\frac{\hat{\mathbf{w}}}{\sqrt{d}})-\mathbb{E}\left[\phi_{1}\left(\frac{\mathbf{w}^{*}}{\sqrt{d}}\right)\right]}\geqslant\epsilon\right) \leqslant\frac{C}{\epsilon^{2}}e^{-cn\epsilon^{4}} \, ,\\
   &\mathbb{P}\left(\abs{\phi_{2}(\frac{\hat{\mathbf{z}}}{\sqrt{n}})-\mathbb{E}\left[\phi_{2}\left(\frac{\mathbf{z}^{*}}{\sqrt{n}}\right)\right]}\geqslant\epsilon\right)\leqslant \frac{C}{\epsilon^{2}}e^{-cn\epsilon^{4}} \, .
\end{align}
\end{theorem}
\emph{Proof}: see Appendix \ref{proof-connect}. Concentration still holds for a larger class of functions $\phi_{1,2}$, but exponential rates are lost. This is discussed in Appendix \ref{main-assumptions}.
\subsection{Theoretical toolbox}
\label{toolbox-app}
Here we remind a few known results that are used throughout the proof. We also provide proofs of useful, straightforward consequences of theses results that do not appear explicitly in the literature for completeness.
\subsubsection{A Gaussian comparison theorem}
We start with the Convex Gaussian Min-max Theorem, as presented in \cite{thrampoulidis2018precise}, which is a tight version of an inequality initially derived in \cite{gordon1985some}. 
\begin{theorem}(CGMT)
\label{CGMT}
Let $\mathbf{G} \in \mathbb{R}^{m\times n}$ be an i.i.d. standard normal matrix and $\mathbf{g} \in \mathbb{R}^{m}$, $\mathbf{h} \in \mathbb{R}^{\samples}$ two i.i.d. standard normal vectors independent of one another. Let $\mathcal{S}_{\mathbf{w}},\mathcal{S}_{\mathbf{u}}$ be two compact sets such that $\mathcal{S}_{\mathbf{w}} \subset \mathbb{R}^{\samples}$ and $\mathcal{S}_{\mathbf{u}} \subset \mathbb{R}^{m}$. Consider the two following optimization problems for any continuous $\psi$ on $\mathcal{S}_{\mathbf{w}}\times\mathcal{S}_{\mathbf{u}}$ :
\begin{align}
    \mathbf{C}(\mathbf{G}) &:= \min_{\mathbf{w}\in \mathcal{S}_{\mathbf{w}}} \max_{\mathbf{u}\in \mathcal{S}_{\mathbf{u}}}\mathbf{u}^{\top}\mathbf{G}\mathbf{w}+\psi(\mathbf{w},\mathbf{u}), \label{PO}\\
    \mathcal{C}(\mathbf{g},\mathbf{h}) &:= \min_{\mathbf{w}\in \mathcal{S}_{\mathbf{w}}} \max_{\mathbf{u}\in \mathcal{S}_{\mathbf{u}}} \norm{\mathbf{w}}_{2}\mathbf{g}^{\top}\mathbf{u}+\norm{\mathbf{u}}_{2}\mathbf{h}^{\top}\mathbf{w}+\psi(\mathbf{w},\mathbf{u})\label{AO}
\end{align}
then the following holds:
\begin{enumerate}
    \item For all $c\in \mathbb{R}$:
    \begin{equation*}
        \mathbb{P}(\mathbf{C}(\mathbf{G})<c)\leqslant 2\mathbb{P}(\mathcal{C}(\mathbf{g},\mathbf{h})\leqslant c)
    \end{equation*}
    \item Further assume that $\mathcal{S}_{\mathbf{w}},\mathcal{S}_{\mathbf{u}}$ are convex sets and $\psi$ is convex-concave on $\mathcal{S}_{\mathbf{w}}\times \mathcal{S}_{\mathbf{u}}$. Then, for all $c\in \mathbb{R}$,
    \begin{equation*}
        \mathbb{P}(\mathbf{C}(\mathbf{G})>c)\leqslant 2\mathbb{P}(\mathcal{C}(\mathbf{g},\mathbf{h})\geqslant c)
    \end{equation*}
    In particular, for all $\mu\in \mathbb{R},t>0,\mathbb{P}(\abs{\mathbf{C}(\mathbf{G})-\mu}>t)\leqslant 2\mathbb{P}(\abs{\mathcal{C}(\mathbf{g},\mathbf{h})-\mu}\geqslant t)$.
    \end{enumerate}
\end{theorem}
Following \cite{thrampoulidis2018precise}, we will say that any reformulation of a target problem matching the form of (\ref{PO}) is an acceptable primary optimization problem (PO), and the corresponding form (\ref{AO}) is an acceptable auxiliary problem (AO). The main idea of this approach is to study the asymptotic properties of the (PO) by studying the simpler (AO). 
\subsubsection{Proximal operators and Moreau envelopes : differentials and useful functions}
Here we remind the definition and some important properties of Moreau envelopes and proximal operators, key elements of convex analysis. Other properties will be used throughout the proof but at less crucial stages, thus we don't remind them explicitly. Our main reference for these properties will be \cite{bauschke2011convex}. \newline
Consider a closed, proper function $f$ such that dom(f)$\subset \mathbb{R}^{\samples}$. Its Moreau envelope and proximal operator are respectively defined by :
\begin{align}
    \mathcal{M}_{\tau f}(\mathbf{x}) = \min_{\mathbf{z} \in \mbox{dom}(f)} \{f(\mathbf{z})+\frac{1}{2\tau}\norm{\mathbf{x}-\mathbf{z}}_{2}^{2}\}, &&\mbox{prox}_{\tau f}(\mathbf{x}) = \argmin_{\mathbf{z} \in \rm{dom}(f)} \{f(\mathbf{z})+\frac{1}{2\tau}\norm{\mathbf{x}-\mathbf{z}}_{2}^{2}\}
\end{align}
As reminded in \cite{thrampoulidis2018precise}, the Moreau envelope is jointly convex in $(\tau,\mathbf{x})$ and differentiable almost everywhere, with gradients:
\begin{align}
    \nabla_{\mathbf{x}} \mathcal{M}_{\tau f}(\mathbf{x}) &= \frac{1}{\tau}(\mathbf{x}-\mbox{prox}_{\tau f}(\mathbf{x})) \label{grad-mor}\\ 
    \frac{\partial}{\partial \tau} \mathcal{M}_{\tau f}(\mathbf{x}) &= -\frac{1}{2\tau^{2}}\norm{\mathbf{x}-\mbox{prox}_{\tau f}(\mathbf{x})}_{2}^{2} \label{part1-mor}
\end{align}
We remind that $\mbox{prox}_{\tau f}(\mathbf{x})$ is the unique point which solves the strongly convex optimization problem defining the Moreau envelope, i.e.:
\begin{equation}
    \label{link-prox-mor}
    \mathcal{M}_{\tau f}(\mathbf{x}) = f(\mbox{prox}_{\tau f}(\mathbf{x}))+\frac{1}{2\tau}\norm{\mathbf{x}-\mbox{prox}_{\tau f}(\mathbf{x})}_{2}^{2}
\end{equation}
We also remind the definition of order k pseudo-Lipschitz function.
\begin{definition}{Pseudo-Lipschitz function}
For $k \in \mathbb{N}^{*}$ and any $n,m \in \mathbb{N}^{*}$, a function $\phi : \mathbb{R}^{\samples} \to \mathbb{R}^{m}$ is called a \emph{pseudo-Lipschitz of order k} if there exists a constant L(k) such that for any $\mathbf{x},\mathbf{y} \in \mathbb{R}^{\samples}$, 
\begin{equation}
    \norm{\phi(\mathbf{x})-\phi(\mathbf{y})}_{2} \leqslant L(k) \left(1+\left(\norm{\mathbf{x}}_{2}\right)^{k-1}+\left(\norm{\mathbf{y}}_{2}\right)^{k-1}\right)\norm{\mathbf{x}-\mathbf{y}}_{2}
\end{equation}
\end{definition}
We now give some further properties that will be helpful throughout the proof.
\begin{lemma}(Moreau envelope of pseudo-Lipschitz function)
\label{Mor-pseudo-Lip}
Consider a proper, lower-semicontinuous, convex, pseudo-Lipschitz function $f:\mathbb{R}^{\samples}\to \mathbb{R}$ of order $k$. Then its Moreau envelope is also pseudo-Lipschitz of order k.
\end{lemma}
\emph{Proof of Lemma \ref{Mor-pseudo-Lip}}:
For any $\mathbf{x},\mathbf{y}$ in $\mbox{dom}(f)$, we have, using the pseudo-Lipschitz property:
\begin{align}
    \abs{f(\mbox{prox}_{\tau f}(\mathbf{x}))-f(\mbox{prox}_{\tau f}(\mathbf{y}))} &\leqslant L(k) \left(1+\left(\norm{\mbox{prox}_{\tau f}(\mathbf{x})}_{2}\right)^{k-1}+\left(\norm{\mbox{prox}_{\tau f}(\mathbf{y})}_{2}\right)^{k-1}\right) \notag\\
    &\hspace{4cm}\norm{\mbox{prox}_{\tau f}(\mathbf{x})-\mbox{prox}_{\tau f}(\mathbf{y})}_{2} \notag\\
    &\leqslant L(k) \left(1+\left(\norm{\mathbf{x}}_{2}\right)^{k-1}+\left(\norm{\mathbf{y}}_{2}\right)^{k-1}\right)\norm{\mathbf{x}-\mathbf{y}}_{2}
\end{align}
where the second line follows immediately with the same constant $L(k)$ owing to the firm-nonexpansiveness of the proximal operator. Furthermore
\begin{align}
    &\norm{\mathbf{x}-\mbox{prox}_{\tau f}(\mathbf{x})}_{2}^{2}-\norm{\mathbf{y}-\mbox{prox}_{\tau f}(\mathbf{y})}_{2}^{2} =\notag \\ 
    &\tau\abs{\partial f(\mbox{prox}_{\tau f}(\mathbf{x}))+\partial f(\mbox{prox}_{\tau f}(\mathbf{y}))}\abs{\left(\mathbf{x}-\mbox{prox}_{\tau f}(\mathbf{x})-\mathbf{y}+\mbox{prox}_{\tau f}(\mathbf{y})\right)} \notag \\
    &\leqslant \tau\norm{\partial f(\mbox{prox}_{\tau f}(\mathbf{x}))+\partial f(\mbox{prox}_{\tau f}(\mathbf{y}))}_{2}\norm{\left(\mathbf{x}-\mbox{prox}_{\tau f}(\mathbf{x})-\mathbf{y}+\mbox{prox}_{\tau f}(\mathbf{y})\right)}_{2}
\end{align}
due to the pseudo-Lipschitz property, one has 
\begin{equation}
    \partial f(\mbox{prox}_{\tau f}(\mathbf{x})) \leqslant L(k)\left(1+2\norm{\mbox{prox}_{\tau f}(\mathbf{x})}_{2}^{k-1}\right)
\end{equation}
This, along with the firm-nonexpansiveness of $\rm{Id}-\mbox{prox}$, concludes the proof.
\qed
\begin{lemma}(Useful functions)
\label{useful-functions}
For any $\mathbf{x} \in \mathbb{R}^{\samples}, \tau >0, \theta \in \mathbb{R}$ and any proper, convex lower semi-continuous function $f$, define the following functions:
\begin{align}
    h_{1} : \mathbb{R} &\to \mathbb{R} \notag \\
    \theta &\mapsto \mathbf{x}^{T}\mbox{prox}_{\tau f(.)}(\theta\mathbf{x}) \\
    h_{2} : \mathbb{R} &\to \mathbb{R} \notag \\
    \tau &\mapsto \frac{1}{2\tau^{2}}\norm{\mathbf{x}-\mbox{prox}_{\tau f(.)}(\mathbf{x})}_{2}^{2} \\
    h_{3} : \mathbb{R} &\to \mathbb{R} \notag \\
    \tau &\mapsto \norm{\mbox{prox}_{\frac{ f}{\tau}(.)}(\frac{\mathbf{x}}{\tau})}_{2}^{2} \\
    h_{4} : \mathbb{R} &\to \mathbb{R} \notag \\
    \tau &\mapsto \norm{\mathbf{x}-\mbox{prox}_{\tau f}(\mathbf{x})}_{2}^{2}
\end{align}
$h_{1}$ is nondecreasing, and $h_{2},h_{3},h_{4}$ are nonincreasing. \newline
\end{lemma}
\emph{Proof of Lemma \ref{useful-functions}}:
For any $\theta,\tilde{\theta}\in\mathbb{R}$:
\begin{align}
    (\theta-\tilde{\theta})(h_{1}(\theta)-h_{1}(\tilde{\theta})) &= (\theta\mathbf{x}-\tilde{\theta}\mathbf{x})^{\top}\left(\mbox{prox}_{\tau f(.)}(\theta\mathbf{x})-\mbox{prox}_{\tau f(.)}(\tilde{\theta}\mathbf{x})\right) \notag\\
    &\geqslant \norm{\mbox{prox}_{\tau f(.)}(\theta\mathbf{x})-\mbox{prox}_{\tau f(.)}(\tilde{\theta}\mathbf{x})}_{2}^{2} \notag\\ 
    &\geqslant 0
\end{align}
where the inequality comes from the firm non-expansiveness of the proximal operator. Thus $h_{1}$ is nondecreasing. \newline
Since the Moreau envelope $\mathcal{M}_{\tau f}(\mathbf{x})$ is convex in $\tau$, we have, for any $\tau,\tilde{\tau}$ in $\mathbb{R}_{++}$
\begin{align}
    \left(\tau-\tilde{\tau}\right)\left(\frac{\partial}{\partial \tau} \mathcal{M}_{\tau f}(\mathbf{x})-\frac{\partial}{\partial \tilde{\tau}} \mathcal{M}_{\tilde{\tau} f}(\mathbf{x})\right) \geqslant 0, &&\iff&& \left(\tau-\tilde{\tau}\right)\left(h_{2}(\tilde{\tau})-h_{2}(\tau)\right) \geqslant 0
\end{align}
which implies that $h_{2}$ is non-increasing. \newline
Using the Moreau decomposition, see e.g. \cite{bauschke2011convex}, we have:
\begin{align}
    h_{2}(\tau) &= \frac{1}{2\tau^{2}}\norm{\mathbf{x}-\left(\mathbf{x}-\tau\mbox{prox}_{\frac{f^{*}}{\tau}}\left(\frac{\mathbf{x}}{\tau}\right)\right)}_{2}^{2} = \norm{\mbox{prox}_{\frac{f^{*}}{\tau}}\left(\frac{\mathbf{x}}{\tau}\right)}_{2}^{2}
\end{align}
which is a nonincreasing function of $\tau$. Since $f$ is convex, we can restart this short process with the conjugate of $f$ to obtain the desired result. Thus $h_{3}$ is nonincreasing and $(\tau-\tilde{\tau})(h_{3}(\tau)-h_{3}(\tilde{\tau})) \leqslant 0$. \newline
Moving to $h_{4}$, proving that it is nonincreasing is equivalent to proving that the following function is increasing
\begin{align}
    h_{5}(\tau) &= \mbox{prox}_{\tau f}(\mathbf{x})^{\top}\left(2\mathbf{x}-\mbox{prox}_{\tau f}(\mathbf{x})\right) 
\end{align}
using the Moreau decomposition again
\begin{equation}
    h_{5}(\tau) = \left(\mathbf{x}-\tau\mbox{prox}_{\frac{ f^{*}}{\tau}}\left(\frac{\mathbf{x}}{\tau}\right)\right)^{\top}\left(\mathbf{x}+\tau\mbox{prox}_{\frac{ f^{*}}{\tau}}\left(\frac{\mathbf{x}}{\tau}\right)\right)
\end{equation}
then, for any $\tau, \tilde{\tau}$ in $\mathbb{R}_{++}$:
\begin{align}
   (\tau-\tilde{\tau})(h_{5}(\tau)-h_{5}(\tilde{\tau})) = (\tau-\tilde{\tau})\left(\tilde{\tau}^{2}\norm{\mbox{prox}_{\frac{ f^{*}}{\tilde{\tau}}}\left(\frac{\mathbf{x}}{\tilde{\tau}}\right)}_{2}^{2}-\tau^{2}\norm{\mbox{prox}_{\frac{ f^{*}}{\tau}}\left(\frac{\mathbf{x}}{\tau}\right)}_{2}^{2}\right)
\end{align}
separating the cases $\tau\leqslant\tilde{\tau}$ and $\tau\geqslant\tilde{\tau}$, and using the result on $h_{3}$ then gives the desired result. \qed \newline
The following inequality is similar to one that appeared in one-dimensional form in \cite{thrampoulidis2018precise}.
\begin{lemma}(A useful inequality)
\label{useful-ineq1}
For any proper, lower semi-continuous convex function $f$, any $\mathbf{x},\tilde{\mathbf{x}}$ in $\mbox{dom}(f)$, and any $\gamma,\tilde{\gamma}\in \mathbb{R}_{++}$, the following holds:
\begin{align}
\left(\mbox{prox}_{\tilde{\gamma} f}(\tilde{\mathbf{x}})-\mbox{prox}_{\gamma f}(\mathbf{x})\right)^{\top}&\left(\frac{\tilde{\mathbf{x}}}{\tilde{\gamma}}-\frac{\mathbf{x}}{\gamma}-\frac{1}{2}\left(\frac{1}{\tilde{\gamma}}-\frac{1}{\gamma}\right)\left(\mbox{prox}_{\tilde{\gamma} f}(\tilde{\mathbf{x}})+\mbox{prox}_{\gamma f}(\mathbf{x})\right)\right)\notag \\
&\geqslant \left(\frac{1}{2\tilde{\gamma}}+\frac{1}{2\gamma}\right)\norm{\left(\mbox{prox}_{\tilde{\gamma} f}(\tilde{\mathbf{x}})-\mbox{prox}_{\gamma f}(\mathbf{x})\right)}_{2}^{2}
\end{align}
\end{lemma}
\emph{Proof of Lemma \ref{useful-ineq1}} : the subdifferential of a proper convex function is a monotone operator, thus:
\begin{align}
    \left(\mbox{prox}_{\tilde{\gamma} f}(\tilde{\mathbf{x}})-\mbox{prox}_{\gamma f}(\mathbf{x})\right)^{\top}\left(\partial f(\mbox{prox}_{\tilde{\gamma} f}(\tilde{\mathbf{x}}))-\partial f(\mbox{prox}_{\gamma f}(\mathbf{x}))\right) \geqslant 0
\end{align}
additionally, $\mbox{prox}_{\gamma f}(\mathbf{x}) = \left(\rm{Id}+\gamma\partial f\right)^{-1}(\mathbf{x})$, hence:
\begin{align}
    &\partial f(\mbox{prox}_{\tilde{\gamma} f}(\tilde{\mathbf{x}}))-\partial f(\mbox{prox}_{\gamma f}(\mathbf{x})) = \left(\frac{\tilde{\mathbf{x}}}{\tilde{\gamma}}-\frac{\mathbf{x}}{\gamma}-\frac{1}{\tilde{\gamma}}\mbox{prox}_{\tilde{\gamma}}(\tilde{\mathbf{x}})+\frac{1}{\gamma}\mbox{prox}_{\gamma f}(\mathbf{x})\right) \notag \\
    &=\frac{\tilde{\mathbf{x}}}{\tilde{\gamma}}-\frac{\mathbf{x}}{\gamma}-\frac{1}{\tilde{\gamma}}\mbox{prox}_{\tilde{\gamma}}(\tilde{\mathbf{x}})+\frac{1}{\gamma}\mbox{prox}_{\gamma f}(\mathbf{x})-\frac{1}{2}\left(\frac{1}{\tilde{\gamma}}-\frac{1}{\gamma}\right)\left(\mbox{prox}_{\tilde{\gamma} f}(\tilde{\mathbf{x}})+\mbox{prox}_{\gamma f}(\mathbf{x})\right) \notag \\
    &+\frac{1}{2}\left(\frac{1}{\tilde{\gamma}}-\frac{1}{\gamma}\right)\left(\mbox{prox}_{\tilde{\gamma} f}(\tilde{\mathbf{x}})+\mbox{prox}_{\gamma f}(\mathbf{x})\right) \notag \\
    &=\left(\frac{\tilde{\mathbf{x}}}{\tilde{\gamma}}-\frac{\mathbf{x}}{\gamma}-\frac{1}{2}\left(\frac{1}{\tilde{\gamma}}-\frac{1}{\gamma}\right)\left(\mbox{prox}_{\tilde{\gamma} f}(\tilde{\mathbf{x}})+\mbox{prox}_{\gamma f}(\mathbf{x})\right)\right)-\left(\frac{1}{2\tilde{\gamma}}+\frac{1}{2\gamma}\right)\left(\mbox{prox}_{\tilde{\gamma} f}(\tilde{\mathbf{x}})-\mbox{prox}_{\gamma f}(\mathbf{x})\right)
\end{align}
which gives the desired inequality. \qed

\subsubsection{Useful concentration of measure elements}
We begin by reminding the Gaussian-Poincaré inequality, see e.g. \cite{boucheron2013concentration}.
\begin{proposition}(Gaussian Poincaré inequality) \\
    \label{Gauss-poinc}
Let $\mathbf{g} \in \mathbb{R}^{\samples}$ be a $\mathcal{N}(0, \mat{I}_{n})$ random vector. Then for any continuous, weakly differentiable $\varphi$, there exists a constant c such that:
\begin{equation}
    \mbox{Var}[\varphi(\mathbf{g})] \leqslant c~\mathbb{E}\left[\norm{\nabla \varphi(\mathbf{g})}_{2}^{2}\right]
\end{equation}
\end{proposition}
We now use this previous result to show Gaussian concentration of Moreau envelopes of appropriately scaled convex functions.
\begin{lemma}(Gaussian concentration of Moreau envelopes)\\
    \label{conc-Mor}
Consider a proper, convex function $f : \mathbb{R}^{\samples} \to \mathbb{R}$ verifying the scaling conditions of Assumptions \ref{main-assumptions} and let $\mathbf{g} \in \mathbb{R}^{\samples}$ be a standard normal 
random vector. Then, for any parameter $\tau>0$ and any $\epsilon>0$, there exists a constant $c$ such that the following holds:
\begin{equation}
    \mathbb{P}\left(\abs{\frac{1}{n}\mathcal{M}_{\tau f(.)}(\mathbf{g})- \mathbb{E}\left[\frac{1}{n}\mathcal{M}_{\tau f(.)}(\mathbf{g})\right]}\geqslant \epsilon\right)\leqslant \frac{c}{n\tau^{2}\epsilon^{2}}
\end{equation}
\end{lemma}
\emph{Proof of Lemma \ref{conc-Mor}}: \quad \\
\quad \\
 We start by showing that the Moreau envelope of a proper, convex function $f:\mathbb{R}^{\samples} \to \mathbb{R}$ verifying the scaling conditions of Assumptions \ref{main-assumptions} is integrable with respect to the Gaussian measure. Using the convexity of the optimization problem defining the Moreau envelope, and the fact that $f$ is proper, there exists $\mathbf{z}_{0} \in \mathbb{R}^{\samples}$ and a finite constant $\mathcal{K}$ such that :
\begin{align}
    \frac{1}{n}\mathcal{M}_{\tau f(.)}(\mathbf{g}) &\leqslant \frac{1}{n}f(\mathbf{z}_{0})+\frac{1}{2n\tau}\norm{\mathbf{z}_{0}-\mathbf{g}}_{2}^{2}\notag \\
    &\leqslant \mathcal{K}+\frac{1}{2n\tau}\norm{\mathbf{z}_{0}-\mathbf{g}}_{2}^{2}
\end{align}
where the second line is integrable under a multivariate Gaussian measure.
Then, using Proposition \ref{Gauss-poinc}, we get:
\begin{align}
    \mbox{Var}\left[\frac{1}{n}\mathcal{M}_{\tau f(.)}(\mathbf{g})\right] &\leqslant \frac{c}{n^{2}}\mathbb{E}\left[\norm{\nabla_{\mathbf{z}}\mathcal{M}_{\tau f(.)}(\mathbf{g})}^{2}_{2}\right] \\
    &=\frac{c}{n^{2}}\mathbb{E}\left[\norm{\frac{1}{\tau}\left(\mathbf{z}-\mbox{prox}_{\tau f}(\mathbf{g})\right)}_{2}^{2}\right]
\end{align}
Using Proposition 12.27 and Corollary 4.3 from \cite{bauschke2011convex}, $\mathbf{g} \to \mathbf{z}-\mbox{prox}_{\tau f}(\mathbf{g})$ is firmly non-expansive and:
\begin{align}
    \norm{\mathbf{g}-\mbox{prox}_{\tau f}(\mathbf{g})}_{2}^{2} &\leqslant \langle \mathbf{g}\vert \mathbf{g}-\mbox{prox}_{\tau f}(\mathbf{g})\rangle \quad \mbox{which implies} \\
    \norm{\mathbf{g}-\mbox{prox}_{\tau f}(\mathbf{g})}_{2}^{2} &\leqslant \norm{\mathbf{g}}_{2}^{2} \quad \mbox{using the Cauchy-Schwarz inequality}
\end{align}
then 
\begin{align}
    \mbox{Var}\left[\frac{1}{n}\mathcal{M}_{\tau f(.)}(\mathbf{g})\right] \leqslant \frac{c}{n^{2}\tau^{2}}\mathbb{E}\left[\norm{\mathbf{g}}_{2}^{2}\right] = \frac{c}{n\tau^{2}}
\end{align}
Chebyshev's inequality then gives, for any $\epsilon>0$:
\begin{equation}
    \mathbb{P}\left(\abs{\frac{1}{n}\mathcal{M}_{\tau f(.)}(\mathbf{g})- \mathbb{E}\left[\frac{1}{n}\mathcal{M}_{\tau f(.)}(\mathbf{g})\right]}\geqslant \epsilon\right)\leqslant \frac{c}{n\tau^{2}\epsilon^{2}}
\end{equation} \qed \\

Gaussian concentration of pseudo-Lipschitz functions of finite order can also be proven using the Gaussian Poincaré inequality to yield a bound similar to the one obtained for Moreau envelopes. We thus give the result without proof:
\begin{lemma}(Concentration of pseudo-Lipschitz functions)
Consider a pseudo-Lipschitz function of finite order k, $f:\mathbb{R}^{\samples}\to \mathbb{R}$. Then for any vector $\mathbf{g} \sim \mathcal{N}(0,\mat{I}_{\samples})$ and any $\epsilon>0$, there exists a constant $C(k)>0$ such that
\begin{equation}
    \mathbb{P}\left(\abs{f(\frac{\mathbf{g}}{\sqrt{n}})-\mathbb{E}\left[f(\frac{\mathbf{g}}{\sqrt{n}})\right]}\geqslant \epsilon\right)\leqslant \frac{L^{2}(k)C(k)}{n\epsilon^{2}}
\end{equation}
\end{lemma}
We now cite an exponential concentration lemma for separable, pseudo-Lipschitz functions of order 2, taken from \cite{ma2017analysis}.
\begin{lemma}(Lemma B.5 from \cite{ma2017analysis})
\label{conc-pseudo-lip-2}
Consider a separable, pseudo-Lipschitz function of order 2, $f:\mathbb{R}^{\samples}\to \mathbb{R}$. Then for any vector $\mathbf{g} \sim \mathcal{N}(0,\mat{I}_{\samples})$ and any $\epsilon>0$, there exists constants $C,c,c'>0$ such that
\begin{equation}
    \mathbb{P}\left(\abs{\frac{1}{n}f(\mathbf{g})-\mathbb{E}\left[\frac{1}{n}f(\mathbf{g})\right]}\geqslant c'\epsilon\right)\leqslant Ce^{-cn\epsilon^{2}}
\end{equation}
where it is understood that $f(\mathbf{g}) = \sum_{i=1}^{n}f(g_{i})$.
\end{lemma}
\subsection{Determining a candidate primary problem, auxiliary problem and its solution.}
We start with a reformulation of the problem (\ref{teacher}-\ref{eq:student}) in order to obtain an acceptable primary problem in the framework of Theorem \ref{CGMT}.
Partitioning the Gaussian distribution, we can rewrite the matrices U and $\mathcal{V}$ in the following way, introducing the standard normal vector:
\begin{equation}
    \begin{bmatrix}
    \mathbf{a} \\
    \mathbf{b}
    \end{bmatrix} \in \mathbb{R}^{p+d} \sim \mathcal{N}(0, \mat{I}_{\tdim+\sdim})
\end{equation}
We can then rewrite the vectors $\mathbf{u},\mathbf{v}$ and matrices $U,\mathcal{V}$ as:
\begin{align}
    \mathbf{u} &= \Psi^{1/2}\mathbf{a}, \quad U = A\Psi^{1/2} \\
    \mathbf{v}&= \Phi^{\top}\Psi^{-1/2}\mathbf{a}+\left(\Omega-\Phi^{\top}\Psi^{-1}\Phi\right)^{1/2}\mathbf{b}, \quad \mathcal{V} = A\Psi^{-1/2}\Phi+B\left(\Omega-\Phi^{\top}\Psi^{-1}\Phi\right)^{1/2}
\end{align}
where the matrices $A$ and $B$ have independent standard normal entries and are independent of $\vec{\theta}_{0}$. The learning problem then becomes equivalent to :
\begin{align}
    &\mbox{Generate labels according to} : \quad \mathbf{y} = f_{0}\left(\frac{1}{\sqrt{\tdim}}A\Psi^{1/2}\vec{\theta}_{0}\right)  \\
    &\mbox{Learn according to} : \quad \argmin_{\mathbf{w}} \loss\left(\frac{1}{\sqrt{\sdim}}\left(A\Psi^{-1/2}\Phi+B\left(\Omega-\Phi^{\top}\Psi^{-1}\Phi\right)^{1/2}\right)\mathbf{w},\mathbf{y}\right)+\reg(\mathbf{w}) 
\end{align}
We are then interested in the optimal cost of the following problem
\begin{equation}
\label{norm-student}
    \min_{\mathbf{w}} \frac{1}{\sdim}\left[ \loss\left(\frac{1}{\sqrt{\sdim}}\left(A\Psi^{-1/2}\Phi+B\left(\Omega-\Phi^{\top}\Psi^{-1}\Phi\right)^{1/2}\right)\mathbf{w},\mathbf{y}\right)+\reg(\mathbf{w})\right]
\end{equation}
Introducing the auxiliary variable $\mathbf{z}$: 
\begin{align}
    &\min_{\mathbf{w}} \loss\left(\frac{1}{\sqrt{\sdim}}\left(A\Psi^{-1/2}\Phi+B\left(\Omega-\Phi^{\top}\Psi^{-1}\Phi\right)^{1/2}\right)\mathbf{w},\mathbf{y}\right)+\reg(\mathbf{w}) \\
 \iff &\min_{\mathbf{w},\mathbf{z}} \loss\left(\mathbf{z},\mathbf{y}\right)+\reg(\mathbf{w}) \notag\\ &\mbox{s.t.} \thickspace \mathbf{z} =\frac{1}{\sqrt{\sdim}} \left(A\Psi^{-1/2}\Phi+B\left(\Omega-\Phi^{\top}\Psi^{-1}\Phi\right)^{1/2}\right)\mathbf{w}
    \end{align}
Introducing the corresponding Lagrange multiplier $\boldsymbol{\lambda} \in \mathbb{R}^{\samples}$ and using strong duality, the problem is equivalent to :
    \begin{align}
    \label{inter1}
    \min_{\mathbf{w},\mathbf{z}}\max_{\boldsymbol{\lambda}} \boldsymbol{\lambda}^{\top}\frac{1}{\sqrt{\sdim}}\left(A\Psi^{-1/2}\Phi+B\left(\Omega-\Phi^{\top}\Psi^{-1}\Phi\right)^{1/2}\right)\mathbf{w}-\boldsymbol{\lambda}^{\top}\mathbf{z}+\loss(\mathbf{z},\mathbf{y})+\reg(\mathbf{w})
\end{align}
In the remainder of the proof, the preceding cost function will be denoted
\begin{equation}
    \mathbf{C}(\mathbf{w},\mathbf{z}) = \max_{\boldsymbol{\lambda}} \boldsymbol{\lambda}^{\top}\frac{1}{\sqrt{\sdim}}\left(A\Psi^{-1/2}\Phi+B\left(\Omega-\Phi^{\top}\Psi^{-1}\Phi\right)^{1/2}\right)\mathbf{w}-\boldsymbol{\lambda}^{\top}\mathbf{z}+\loss(\mathbf{z},\mathbf{y})+\reg(\mathbf{w})
\end{equation}
such that the problem reads $\min_{\mathbf{w},\mathbf{z}} \mathbf{C}(\mathbf{w},\mathbf{z})$. Theorem \ref{CGMT} requires working with compact feasibility sets. Adopting similar approaches to the ones from \cite{thrampoulidis2018precise,dhifallah2020precise}, the next lemma shows that the optimization problem (\ref{inter1}) can be equivalently recast as one over compact sets.
    
    \begin{lemma}(Compactness of feasibility set)
    \label{compacity-lemma}
    Let $\mathbf{w}^{*},\mathbf{z}^{*},\boldsymbol{\lambda}^{*}$ be optimal in (\ref{inter1}). Then there exists positive constants $C_{\mathbf{w}}, C_{\mathbf{z}}$ and $C_{\boldsymbol{\lambda}}$ such that
    \begin{align}
        \mathbb{P}\left(\norm{\mathbf{w}^{*}}_{2} \leqslant C_{\mathbf{w}}\sqrt{\sdim}\right) \xrightarrow[d\to \infty]{P}1, \thickspace
        \mathbb{P}\left(\norm{\mathbf{z}^{*}}_{2} \leqslant C_{\mathbf{z}}\sqrt{n}\right) \xrightarrow[n\to \infty]{P}1, \thickspace
        \mathbb{P}\left(\norm{\boldsymbol{\lambda}^{*}}_{2} \leqslant C_{\boldsymbol{\lambda}}\sqrt{n}\right) \xrightarrow[\samples\to \infty]{P}1 
    \end{align}
    \end{lemma}
    \emph{Proof of Lemma \ref{compacity-lemma}}:  consider the initial minimisation problem:
\begin{equation}
    \label{copy-init}
    \hat{\mathbf{w}} = \argmin_{\mathbf{w}\in \mathbb{R}^{\sdim}} g\left(\frac{1}{\sqrt{\sdim}}\mathcal{V}\mathbf{w},\mathbf{y}\right)+\reg(\mathbf{w})
\end{equation}
From assumption (A3), the cost function $\loss+\reg$ is coercive, proper and lower semi-continuous. Since it is proper, there exists
$\mathbf{w}_{0} \in \mathbb{R}^{\sdim}$ such that $g\left(\frac{1}{\sqrt{\sdim}}\mathcal{V}\mathbf{w},\mathbf{y}\right)+\reg(\mathbf{w}) \in \mathbb{R}$. The coercivity implies that
there exists $\eta \in ]0,+\infty[$ such that, for every $\mathbf{w} \in \mathbb{R}^{\sdim}$ satisfying $\norm{\mathbf{w}-\mathbf{w}_{0}} \geqslant \eta$, $g\left(\frac{1}{\sqrt{\sdim}}\mathcal{V}\mathbf{w},\mathbf{y}\right)+\reg(\mathbf{w}) \geqslant g\left(\frac{1}{\sqrt{\sdim}}\mathcal{V}\mathbf{w}_{0},\mathbf{y}\right)+\reg(\mathbf{w}_{0})$.
Let $S = \{\mathbf{w} \in \mathbb{R}^{\sdim} \vert \norm{\mathbf{w}-\mathbf{w}_{0}} \leqslant \eta \}$. Then $S \cap \mathbb{R}^{\sdim} \neq \emptyset$ and $S$ is compact. Then, there exists $\mathbf{w}^{*} \in S$ such that 
$g\left(\frac{1}{\sqrt{\sdim}}\mathcal{V}\mathbf{w}^{*},\mathbf{y}\right)+\reg(\mathbf{w}^{*}) = \inf_{\mathbf{w}\in S} g\left(\frac{1}{\sqrt{\sdim}}\mathcal{V}\mathbf{w},\mathbf{y}\right)+\reg(\mathbf{w}) \leqslant g\left(\frac{1}{\sqrt{\sdim}}\mathcal{V}\mathbf{w}_{0},\mathbf{y}\right)+f(\mathbf{w}_{0})$.
Thus $g\left(\frac{1}{\sqrt{\sdim}}\mathcal{V}\mathbf{w}^{*},\mathbf{y}\right)+\reg(\mathbf{w}^{*}) \in \inf_{\mathbf{w}\in \mathbb{R}^{\sdim}} g\left(\frac{1}{\sqrt{\sdim}}\mathcal{V}\mathbf{w},\mathbf{y}\right)+\reg(\mathbf{w})$ and the set of minimisers is bounded. Closure is immediately checked by considering a sequence of minimisers converging to $\mathbf{w}^{*}$. \\
We conclude that the set of minimisers of problem (\ref{copy-init}) is a non-empy compact set. Then there exists a constant $C_{\mathbf{w}}$ independent of the dimension $d$, such that:
\begin{equation}
    \norm{\mathbf{w}}_{2} \leqslant C_{\mathbf{w}}\sqrt{\sdim}
\end{equation}
Now consider the equivalent formulation of problem (\ref{copy-init}):
\begin{align}
    \min_{\mathbf{w},\mathbf{z}}\max_{\boldsymbol{\lambda}} \boldsymbol{\lambda}^{\top}\frac{1}{\sqrt{\sdim}}\mathcal{V}\mathbf{w}-\boldsymbol{\lambda}^{\top}\mathbf{z}+\loss(\mathbf{z},\mathbf{y})+\reg(\mathbf{w})
\end{align}
Its optimality condition reads :
\begin{align}
    \nabla_{\boldsymbol{\lambda}}: \frac{1}{\sqrt{\sdim}}\mathcal{V}\mathbf{w}=\mathbf{z}, &&
    \nabla_{\mathbf{z}}: \boldsymbol{\lambda} \in \partial \loss(\mathbf{z},\mathbf{y}), &&
    \nabla_{\mathbf{w}}: \frac{1}{\sqrt{\sdim}}\mathcal{V}^{\top}\boldsymbol{\lambda} \in \partial \reg(\mathbf{w})
\end{align}
The optimality condition in $\boldsymbol{\lambda}$ gives:
\begin{align}
    \norm{\mathbf{z}}_{2} &\leqslant \norm{\frac{1}{\sqrt{\sdim}}\mathcal{V}}_{op}\norm{\mathbf{w}}_{2} \notag\\
    &\leqslant \norm{\frac{1}{\sqrt{\sdim}}\left(A\Psi^{-1/2}\Phi+B\left(\Omega-\Phi^{\top}\Psi^{-1}\Phi\right)^{1/2}\right)}_{op}\norm{\mathbf{w}}_{2} \notag\\
    &\leqslant \left[\norm{\Psi^{-1/2}\Phi}_{op}\norm{\frac{1}{\sqrt{\sdim}}A}_{op}+\norm{\left(\Omega-\Phi^{\top}\Psi^{-1}\Phi\right)^{1/2}}_{op}\norm{\frac{1}{\sqrt{\sdim}}B}_{op}\right]\norm{\mathbf{w}}_{2}
\end{align}
According to assumption (A2), the operator norms of the matrices involving the covariance matrices are bounded with high probability and using known results on random matrices, see e.g. \cite{vershynin2010introduction}, the operator norms of $\frac{1}{\sqrt{\sdim}}A$ and $\frac{1}{\sqrt{\sdim}}B$ are bounded by finite constants with high probability when the dimensions go to infinity.
 Thus there exists a constant $C_{\mathbf{z}}$ also independent of d such that:
\begin{equation}
    \mathbb{P}\left(\norm{\mathbf{z}}_{2} \leqslant C_{\mathbf{Z}}\sqrt{n}\right) \xrightarrow[n \to \infty]{P} 1
\end{equation}
Finally, the scaling condition from assumption (A3) directly shows that there exists a constant $C_{\boldsymbol{\lambda}}$ such that 
\begin{equation}
    \mathbb{P}\left(\norm{\boldsymbol{\lambda}}_{2} \leqslant C_{\boldsymbol{\lambda}}\sqrt{n}\right) \xrightarrow[n \to \infty]{P} 1
\end{equation}
This concludes the proof of Lemma \ref{compacity-lemma}. \qed \newline 
\quad \newline
    Defining the sets $\mathcal{S}_{\mathbf{w}} = \{\mathbf{w}\in \mathbb{R}^{\sdim} \vert \norm{\mathbf{w}}_{2}\leqslant C_{\mathbf{w}}\sqrt{\sdim}\},\mathcal{S}_{\mathbf{z}} = \{\mathbf{z}\in \mathbb{R}^{\samples} \vert \norm{\mathbf{z}}_{2}\leqslant C_{\mathbf{z}}\sqrt{n}\}$ and $\mathcal{S}_{\boldsymbol{\lambda}} = \{\boldsymbol{\lambda}\in \mathbb{R}^{\samples} \vert \norm{\boldsymbol{\lambda}}_{2} \leqslant C_{\boldsymbol{\lambda}}\sqrt{n}\}$, the optimization problem can now be reduced to:
    \begin{align}
    \label{compact-pre-PO}
    \min_{\mathbf{w}\in\mathcal{S}_{\mathbf{w}},\mathbf{z}\in\mathcal{S}_{\mathbf{z}}}\max_{\boldsymbol{\lambda}\in\mathcal{S}_{\boldsymbol{\lambda}}} \boldsymbol{\lambda}^{\top}\frac{1}{\sqrt{\sdim}}\left(A\Psi^{-1/2}\Phi+B\left(\Omega-\Phi^{\top}\Psi^{-1}\Phi\right)^{1/2}\right)\mathbf{w}-\boldsymbol{\lambda}^{\top}\mathbf{z}+\loss(\mathbf{z},\mathbf{y})+\reg(\mathbf{w})
    \end{align}
    The rest of this section can then be summarized by the following lemma, the proof of which shows how to find an acceptable (PO) for problem (\ref{compact-pre-PO}), the corresponding (AO) and how to reduce the (AO) to a scalar optimization problem.
    At this point we will assume the teacher vector $\boldsymbol{\theta}_{0}$ is deterministic, and relax this assumption in paragraph \ref{relax-teacher}. For this reason we do not add it to the initial list of assumptions in section \ref{main-assumptions}.
    \begin{lemma}(Scalar equivalent problem)
    \label{lemma:scalar-eq}
        In the framework of Theorem \ref{CGMT}, acceptable (AO)s of problem (\ref{compact-pre-PO}) can be reduced to the following scalar optimization problems 
        \begin{align}
            &\mbox{For $\vec{\theta}_{0}\notin\mbox{Ker}(\Phi^{\top})$}:\max_{\kappa,\nu,\tau_{2}} \min_{m,\eta,\tau_{1}}\mathcal{E}_{n}(\tau_{1},\tau_{2},\kappa,\eta,\nu,m) \\
            &\mbox{For $\vec{\theta}_{0}\in\mbox{Ker}(\Phi^{\top})$}:\max_{\kappa,\tau_{2}} \min_{\eta,\tau_{1}}\mathcal{E}^{0}_{n}(\tau_{1},\tau_{2},\kappa,\eta)
        \end{align}
        where 
       \begin{align}
    &\mathcal{E}_{n}(\tau_{1},\tau_{2},\kappa,\eta,\nu,m) = \frac{\kappa\tau_{1}}{2}-\frac{\eta\tau_{2}}{2}+m\nu\sqrt{\gamma}-\frac{\tau_{2}}{2\eta}\frac{m^{2}}{\rho} \notag \\
    &-\frac{\eta}{2\tau_{2}d}(\nu\mathbf{v}+\kappa\Omega^{1/2}\mathbf{g})^{\top}\Omega^{-1}(\nu\mathbf{v}+\kappa\Omega^{1/2}\mathbf{g})-\kappa\mathbf{g}^{\top}\left(\Sigma^{1/2}-\Omega^{1/2}\right)\frac{m\sqrt{\gamma}}{\norm{\tilde{\mathbf{v}}}_{2}^{2}}{\mathbf{v}} \notag \\
    &+\frac{1}{\sdim}\mathcal{M}_{\frac{\tau_{1}}{\kappa}\loss(.,\mathbf{y})}\left(\frac{m}{\sqrt{\rho}}\mathbf{s}+\eta\mathbf{h}\right)+\frac{1}{\sdim}\mathcal{M}_{\frac{\eta}{\tau_{2}}\reg(\Omega^{-1/2}.)}\left(\frac{\eta}{\tau_{2}}\left(\nu\Omega^{-1/2}\tilde{\mathbf{v}}+\kappa\mathbf{g}\right)\right), \\
    &\mathcal{E}_{n}^{0}(\tau_{1},\tau_{2},\kappa,\nu)= -\frac{\eta\tau_{2}}{2}+\frac{\kappa\tau_{1}}{2}+\frac{1}{\sdim}\mathcal{M}_{\frac{\tau_{1}}{\kappa}\loss(.,\mathbf{y})}(\eta\mathbf{h})+\frac{1}{\sdim}\mathcal{M}_{\frac{\eta}{\tau_{2}}f(\Omega^{-1/2}.)}(\frac{\eta}{\tau_{2}}\kappa\mathbf{g})-\frac{\eta}{2\tau_{2}d}\kappa^{2}\mathbf{g}^{\top}\mathbf{g}
\end{align}
and 
\begin{equation}
    \Sigma = \Omega-\frac{\tilde{\mathbf{v}}\tilde{\mathbf{v}}^{T}}{\rho \tdim} \quad \tilde{\mathbf{v}} = \Phi^{T}\vec{\theta}_{0} \quad \rho = \frac{1}{\tdim}\vec{\theta}_{0}^{\top}\Psi\vec{\theta}_{0}
\end{equation}
    \end{lemma}
    \emph{Proof of Lemma \ref{lemma:scalar-eq}}:
    We need to find an i.i.d. Gaussian matrix independent from the rest of the problem in order to use Theorem \ref{CGMT}. We thus decompose the mixing matrix A by taking conditional expectations w.r.t. $\mathbf{y}$, which amounts to conditioning on a linear subset of the Gaussian space generated by A. Dropping the feasibility sets for confort of notation in the following lines:
    \begin{align}
    &\min_{\mathbf{w},\mathbf{z}}\max_{\boldsymbol{\lambda}} \boldsymbol{\lambda}^{\top}\frac{1}{\sqrt{\sdim}}\left(\left(\mathbb{E}\left[A\vert \mathbf{y}\right]+A-\mathbb{E}\left[A\vert \mathbf{y}\right]\right)\Psi^{-1/2}\Phi+B\left(\Omega-\Phi^{\top}\Psi^{-1}\Phi\right)^{1/2}\right)\mathbf{w}\notag\\
    &\hspace{2.5cm}-\boldsymbol{\lambda}^{\top}\mathbf{z}+\loss(\mathbf{z},\mathbf{y})+\reg(\mathbf{w}) \\
    & \iff \min_{\mathbf{w},\mathbf{z}}\max_{\boldsymbol{\lambda}} \boldsymbol{\lambda}^{\top}\frac{1}{\sqrt{\sdim}}\bigg(\left(\mathbb{E}\left[A\vert A\Psi^{1/2}\vec{\theta}_{0}\right]+A-\mathbb{E}\left[A\vert A\Psi^{1/2}\vec{\theta}_{0}\right]\right)\Psi^{-1/2}\Phi \notag\\
    &\hspace{2cm}+B\left(\Omega-\Phi^{\top}\Psi^{-1}\Phi\right)^{1/2}\bigg)\mathbf{w}-\boldsymbol{\lambda}^{\top}\mathbf{z}+\loss(\mathbf{z},\mathbf{y})+\reg(\mathbf{w})
    \end{align}
    Conditioning in Gaussian spaces amounts to doing orthogonal projections. Denoting $\tilde{\boldsymbol{\theta}}_{0} = \Psi^{1/2}\vec{\theta}_{0}$ and $\tilde{A}$ a copy of $A$ independent of $\mathbf{y}$, the minimisation problem then becomes:
     \begin{align}
     &\min_{\mathbf{w},\mathbf{z}}\max_{\boldsymbol{\lambda}} \boldsymbol{\lambda}^{\top}\frac{1}{\sqrt{\sdim}}\left(\left(A\mathbf{P}_{\tilde{\boldsymbol{\theta}}_{0}}+\tilde{A}\mathbf{P}^{\perp}_{\tilde{\boldsymbol{\theta}}_{0}}\right)\Psi^{-1/2}\Phi+B\left(\Omega-\Phi^{\top}\Psi^{-1}\Phi\right)^{1/2}\right)\mathbf{w}-\boldsymbol{\lambda}^{\top}\mathbf{z}+\loss(\mathbf{z},\mathbf{y})+\reg(\mathbf{w}) \\
     & \iff \min_{\mathbf{w},\mathbf{z}}\max_{\boldsymbol{\lambda}} \boldsymbol{\lambda}^{\top}\frac{1}{\sqrt{\sdim}}A\mathbf{P}_{\tilde{\boldsymbol{\theta}}_{0}}\Psi^{-1/2}\Phi\mathbf{w}+\boldsymbol{\lambda}^{\top}\frac{1}{\sqrt{\sdim}}\tilde{A}\mathbf{P}^{\perp}_{\tilde{\boldsymbol{\theta}}_{0}}\Psi^{-1/2}\Phi\mathbf{w}+\boldsymbol{\lambda}^{\top}\frac{1}{\sqrt{\sdim}}B\left(\Omega-\Phi^{\top}\Psi^{-1}\Phi\right)^{1/2}\mathbf{w}\notag\\
     &-\boldsymbol{\lambda}^{\top}\mathbf{z}+\loss(\mathbf{z},\mathbf{y})+\reg(\mathbf{w}) \\
     & \iff \min_{\mathbf{w},\mathbf{z}}\max_{\boldsymbol{\lambda}} \boldsymbol{\lambda}^{\top}\frac{1}{\sqrt{\sdim}}\mathbf{s}\frac{\tilde{\boldsymbol{\theta}}_{0}^{\top}}{\norm{\tilde{\vec{\theta}_{0}}}_{2}}\Psi^{-1/2}\Phi\mathbf{w}+\boldsymbol{\lambda}^{\top}\frac{1}{\sqrt{\sdim}}\tilde{A}\mathbf{P}^{\perp}_{\tilde{\boldsymbol{\theta}}_{0}}\Psi^{-1/2}\Phi\mathbf{w}+\boldsymbol{\lambda}^{\top}\frac{1}{\sqrt{\sdim}}B\left(\Omega-\Phi^{\top}\Psi^{-1}\Phi\right)^{1/2}\mathbf{w}\notag\\
     &-\boldsymbol{\lambda}^{\top}\mathbf{z}+\loss(\mathbf{z},\mathbf{y})+\reg(\mathbf{w})
\end{align}
where we used $\mathbf{P}_{\tilde{\boldsymbol{\theta}}_{0}} = \frac{\tilde{\boldsymbol{\theta}}_{0}\tilde{\boldsymbol{\theta}}_{0}^{\top}}{\norm{\tilde{\boldsymbol{\theta}}_{0}}_{2}^{2}}$ and $\mathbf{s} = A\frac{\tilde{\vec{\theta}_{0}}}{\norm{\tilde{\vec{\theta}_{0}}}_{2}}$. Knowing that $\tilde{A},B$ are independent standard Gaussian matrices, and independent from $\mathbf{A},\mathbf{y}, f_{0}$, we can rewrite the problem as :
\begin{align}
    \min_{\mathbf{w},\mathbf{z}}\max_{\boldsymbol{\lambda}} \boldsymbol{\lambda}^{\top}\frac{1}{\sqrt{\sdim}}\mathbf{s}\frac{\vec{\theta}_{0}^{\top}}{\norm{\Psi^{1/2}\vec{\theta}_{0}}}\Phi\mathbf{w}+\boldsymbol{\lambda}^{\top}\frac{1}{\sqrt{\sdim}}Z\Sigma^{1/2}\mathbf{w}-\boldsymbol{\lambda}^{\top}\mathbf{z}+\loss(\mathbf{z},\mathbf{y})+\reg(\mathbf{w})
\end{align}
where $\Sigma = \Phi^{\top}\Psi^{-1/2}\mathbf{P}^{\perp}_{\tilde{\boldsymbol{\theta}}_{0}}\Psi^{-1/2}\Phi+\Omega-\Phi^{\top}\Psi^{-1}\Phi = \Omega-\Phi^{\top}\Psi^{-1/2}\mathbf{P}_{\tilde{\theta}_{0}}\Psi^{-1/2}\Phi$, and $Z$ is a standard Gaussian matrix independent of $\mathbf{A},\mathbf{y}, f_{0}$. Recall $\rho = \frac{1}{p}\vec{\theta}_{0}^{\top}\Psi\vec{\theta}_{0}$ from the main text. Replacing with the expression of $\tilde{\boldsymbol{\theta}}_{0}$ and letting $\tilde{\mathbf{v}} = \Phi^{\top}\vec{\theta}_{0}$, we have
\begin{align}
    \Sigma &= \Omega-\phi^{\top}\Psi^{-1/2}\tilde{\boldsymbol{\theta}}_{0}\tilde{\vec{\theta}_{0}}^{\top}\Psi^{-1/2}\Phi\frac{1}{\norm{\tilde{\theta}_{0}}_{2}^{2}} = \Omega-\frac{\phi^{\top}\vec{\theta}_{0}\vec{\theta}_{0}^{\top}\Phi}{\vec{\theta}_{0}^{\top}\Psi\vec{\theta}_{0}}\\
    &= \Omega-\frac{\tilde{\mathbf{v}}\tilde{\mathbf{v}}^{\top}}{p\rho}
\end{align} 
The problem then becomes
\begin{align}
    \label{inter-PO}
    \min_{\mathbf{w},\mathbf{z}}\max_{\boldsymbol{\lambda}} \boldsymbol{\lambda}^{\top}\frac{1}{\sqrt{\sdim\tdim}}\mathbf{s}\frac{\tilde{\mathbf{v}}^{\top}}{\sqrt{\rho}}\mathbf{w}+\boldsymbol{\lambda}^{\top}\frac{1}{\sqrt{\sdim}}Z\Sigma^{1/2}\mathbf{w}-\boldsymbol{\lambda}^{\top}\mathbf{z}+\loss(\mathbf{z},\mathbf{y})+\reg(\mathbf{w})
\end{align}
Two cases must now be considered, $\vec{\theta}_{0} \notin \mbox{Ker}(\phi^{\top})$ and $\vec{\theta}_{0} \in \mbox{Ker}(\phi^{\top})$. Another possible case is $\Phi = 0_{p\times d}$, however it leads to the same steps as the case $\vec{\theta}_{0} \in \mbox{Ker}(\Phi^{\top})$.
\paragraph{Case 1: $\vec{\theta}_{0}\notin\mbox{Ker}(\Phi^{\top})$} \quad \newline
\quad \newline
It is tempting to invert the matrix $\Sigma^{1/2}$ to make the change of variable $\mathbf{w}_{\perp}=\Sigma^{1/2}\mathbf{w}$ and continue the calculation. However there is no guarantee that $\Sigma$ is invertible : it is only semi-positive definite. Taking identities everywhere gives for examples $\mathbf{P}^{\perp}_{\tilde{\theta}_{0}}$ which is non-invertible. We thus introduce an additional variable:
\begin{align}
     \min_{\mathbf{w},\mathbf{z},\mathbf{p}}\max_{\boldsymbol{\lambda},\boldsymbol{\mu}} \boldsymbol{\lambda}^{\top}\frac{1}{\sqrt{\sdim\tdim}}\mathbf{s}\frac{\tilde{\mathbf{v}}^{\top}}{\sqrt{\rho}}\mathbf{w}+\boldsymbol{\lambda}^{\top}\frac{1}{\sqrt{\sdim}}Z\mathbf{p}-\boldsymbol{\lambda}^{\top}\mathbf{z}+\loss(\mathbf{z},\mathbf{y})+\reg(\mathbf{w})+\boldsymbol{\mu}^{\top}\left(\Sigma^{1/2}\mathbf{w}-\mathbf{p}\right)
\end{align}
 Here the minimisation on $f$ and $g$ is linked by the bilinear form $\boldsymbol{\lambda}^{\top}\mathbf{s}\tilde{\mathbf{v}}^{\top}\mathbf{w}$. We wish to separate them in order for the Moreau envelopes to appear later on in simple fashion. To do so, we introduce the orthogonal decomposition of $\mathbf{w}$ on the direction of $\tilde{\mathbf{v}}$:
\begin{align}
    \mathbf{w} &= \left(\mathbf{P}_{\tilde{\mathbf{v}}}+\mathbf{P}_{\tilde{\mathbf{v}}}^{\perp}\right)\mathbf{w} = \frac{\tilde{\mathbf{v}}^{\top}\mathbf{w}}{\norm{\tilde{\mathbf{v}}}_{2}^{2}}\tilde{\mathbf{v}}+\mathbf{P}_{\tilde{\mathbf{v}}}^{\perp}\mathbf{w} \notag\\
    &= \frac{\tilde{\mathbf{v}}^{\top}\mathbf{w}}{\norm{\tilde{\mathbf{v}}}_{2}^{2}}\tilde{\mathbf{v}}+\mathbf{w}_{\perp} \thickspace \mbox{where} \thickspace \mathbf{w}_{\perp}\perp\tilde{\mathbf{v}} \notag\\
    &=\frac{m\sqrt{\sdim\tdim}}{\norm{\tilde{\mathbf{v}}}_{2}^{2}}{\tilde{\mathbf{v}}}+\mathbf{w}_{\perp} \quad \mbox{where $m = \frac{1}{\sqrt{\sdim\tdim}}\tilde{\mathbf{v}}^{\top}\mathbf{w}$}
\end{align}
\noindent where the parameter $m$ corresponds to the one defined in (\ref{nice-overlaps}). This gives the following, after introducing the scalar Lagrange multiplier $\nu \in \mathbb{R}$ to enforce the constraint $\mathbf{w}_{\perp} \perp \tilde{\mathbf{v}}$. Note that several methods can be used to express the orthogonality constraint, as in e.g. \cite{dhifallah2020precise}, but the one chosen here allows to complete the proof and match the replica prediction. Reintroducing the normalization, we then have the equivalent form for (\ref{norm-student}):
\begin{align}
\label{finalPO}
     \min_{m,\mathbf{w}_{\perp},\mathbf{z},\mathbf{p}}\max_{\boldsymbol{\lambda},\boldsymbol{\mu},\nu}\frac{1}{\sdim}&\bigg[ \boldsymbol{\lambda}^{\top}\frac{m}{\sqrt{\rho}}\mathbf{s}+\boldsymbol{\lambda}^{\top}\frac{1}{\sqrt{\sdim}}Z\mathbf{m}-\boldsymbol{\lambda}^{\top}\mathbf{z}+\loss(\mathbf{z},\mathbf{y})+\reg\left(\frac{m\sqrt{\sdim\tdim}}{\norm{\tilde{\mathbf{v}}}_{2}^{2}}{\tilde{\mathbf{v}}}+\mathbf{w}_{\perp}\right)\notag\\
     &\quad+\boldsymbol{\mu}^{\top}\left(\Sigma^{1/2}\left(\frac{m\sqrt{\sdim\tdim}}{\norm{\tilde{\mathbf{v}}}_{2}^{2}}{\tilde{\mathbf{v}}}+\mathbf{w}_{\perp}\right)-\mathbf{p}\right)-\nu\tilde{\mathbf{v}}^{\top}\mathbf{w}_{\perp}\bigg]
\end{align}\\
A follow-up of the previous equations shows that the feasibility set now reads : 
\begin{align}
    \mathcal{S}_{m,\mathbf{w}_{\perp},\mathbf{z},\mathbf{p},\boldsymbol{\lambda},\boldsymbol{\mu},\nu} = &\bigg\{m \in \mathbb{R},\mathbf{w}_{\perp}\in \mathbb{R}^{d-1},\mathbf{z}\in \mathbb{R}^{\samples},\mathbf{p}\in \mathbb{R}^{\sdim},\boldsymbol{\lambda}\in \mathbb{R}^{\samples},\boldsymbol{\mu}\in \mathbb{R}^{\sdim},\nu\in \mathbb{R}\hspace{0.1cm} \vert \notag \\ &\quad\sqrt{m^{2}+\frac{\norm{\mathbf{w}_{\perp}}_{2}^{2}}{d}} \leqslant C_{\mathbf{w}}, \norm{\mathbf{z}}_{2}\leqslant C_{\mathbf{z}}\sqrt{n},\norm{\mathbf{p}}_{2} \leqslant \sigma_{max}(\Sigma^{1/2})C_{\mathbf{w}}\sqrt{\sdim},\norm{\boldsymbol{\lambda}}_{2}\leqslant C_{\boldsymbol{\lambda}}\sqrt{n}\bigg\}
\end{align}
where the boundedness of $\norm{\mathbf{p}}_{2}$ follows immediately from the assumptions on the covariance matrices and Lemma \ref{compacity-lemma}. We denote $\mathcal{S}_{\mathbf{p}} = \{\mathbf{p}\in \mathbb{R}^{\sdim} \vert \norm{\mathbf{p}}_{2}\leqslant C_{\mathbf{p}}\}$ for some constant $C_{\mathbf{p}} \geqslant \sigma_{max}(\Sigma^{1/2})C_{\mathbf{w}}$. \\
\quad\\
The set $\mathcal{S}_{\mathbf{p}}\times \mathcal{S}_{\boldsymbol{\lambda}}$ is compact and the matrix $Z$ is independent of all other random quantities of the problem, thus problem (\ref{finalPO}) is an acceptable (PO). We can now write the auxiliary optimization problem (AO) corresponding to the primary one (\ref{finalPO}), dropping the feasibility sets again for convenience:
\begin{align}
    \label{AO_init}
     \min_{m,\mathbf{w}_{\perp},\mathbf{z},\mathbf{p}}\max_{\boldsymbol{\lambda},\boldsymbol{\mu},\nu}\frac{1}{\sdim}&\bigg[ \boldsymbol{\lambda}^{\top}\frac{m}{\sqrt{\rho}}\mathbf{s}+\frac{1}{\sqrt{\sdim}}\norm{\boldsymbol{\lambda}}_{2}\mathbf{g}^{\top}\mathbf{p}+\frac{1}{\sqrt{\sdim}}\norm{\mathbf{p}}_{2}\mathbf{h}^{\top}\boldsymbol{\lambda}-\boldsymbol{\lambda}^{\top}\mathbf{z}+\loss(\mathbf{z},\mathbf{y})+\reg\left(\frac{m\sqrt{\sdim\tdim}}{\norm{\tilde{\mathbf{v}}}_{2}^{2}}{\tilde{\mathbf{v}}}+\mathbf{w}_{\perp}\right)\notag\\
     &\quad+\boldsymbol{\mu}^{\top}\left(\Sigma^{1/2}\left(\frac{m\sqrt{\sdim\tdim}}{\norm{\tilde{\mathbf{v}}}_{2}^{2}}{\tilde{\mathbf{v}}}+\mathbf{w}_{\perp}\right)-\mathbf{p}\right)-\nu\tilde{\mathbf{v}}^{\top}\mathbf{w}_{\perp}\bigg]
\end{align}
We now turn to the simplification of this problem. \\
\quad \\
The variable $\vec{\lambda}$ only appears in linear terms, we can thus directly optimize over its direction, introducing the positive scalar variable $\kappa = \norm{\boldsymbol{\lambda}}_{2}/\sqrt{\sdim}$:
\begin{align}
\label{invert1}
     \min_{m,\mathbf{w}_{\perp},\mathbf{z},\mathbf{p}}\max_{\kappa,\boldsymbol{\mu},\nu} \frac{1}{\sdim}&\bigg[ \kappa\mathbf{g}^{\top}\mathbf{p}+\kappa\norm{\frac{m}{\sqrt{\rho}}\sqrt{\sdim}\mathbf{s}+\norm{\mathbf{p}}_{2}\mathbf{h}-\sqrt{\sdim}\mathbf{z}}_{2}+\loss(\mathbf{z},\mathbf{y})+\reg\left(\frac{m\sqrt{\sdim\tdim}}{\norm{\tilde{\mathbf{v}}}_{2}^{2}}{\tilde{\mathbf{v}}}+\mathbf{w}_{\perp}\right)\notag\\
     &\quad+\boldsymbol{\mu}^{\top}\left(\Sigma^{1/2}\left(\frac{m\sqrt{\sdim\tdim}}{\norm{\tilde{\mathbf{v}}}_{2}^{2}}{\tilde{\mathbf{v}}}+\mathbf{w}_{\perp}\right)-\mathbf{p}\right)-\nu\tilde{\mathbf{v}}^{\top}\mathbf{w}_{\perp}\bigg]
\end{align}
\quad \\
The previous expression may not be convex-concave because of the term $\norm{\mathbf{p}}_{2}\mathbf{h}$. However, it was shown in \cite{thrampoulidis2018precise} that the order of the min and max can still be inverted in this case, because of the convexity of the original problem. As the proof would be very similar, we do not reproduce it. Inverting the max-min order and performing the linear optimization on $\mathbf{p}$ with $\eta = \norm{\mathbf{p}}_{2}/\sqrt{\sdim}$:
\begin{align}
     \max_{\kappa,\boldsymbol{\mu},\nu} \min_{m,\mathbf{w}_{\perp},\mathbf{z},\eta} & \bigg\{-\frac{\eta}{\sqrt{\sdim}}\norm{\boldsymbol{\mu}+\kappa\mathbf{g}}_{2}+\frac{\kappa}{\sqrt{\sdim}}\norm{\frac{m}{\sqrt{\rho}}\mathbf{s}+\eta\mathbf{h}-\mathbf{z}}_{2}+\notag\\
     &+\frac{1}{\sdim}\bigg[\loss(\mathbf{z},\mathbf{y})
     +\reg\left(\frac{m\sqrt{\sdim\tdim}}{\norm{\tilde{\mathbf{v}}}_{2}^{2}}{\tilde{\mathbf{v}}}+\mathbf{w}_{\perp}\right)+\boldsymbol{\mu}^{\top}\Sigma^{1/2}\left(\frac{m\sqrt{\sdim\tdim}}{\norm{\tilde{\mathbf{v}}}_{2}^{2}}{\tilde{\mathbf{v}}}+\mathbf{w}_{\perp}\right)-\nu\tilde{\mathbf{v}}^{\top}\mathbf{w}_{\perp}\bigg]\bigg\}
\end{align}
using the following representation of the norm, as in \cite{thrampoulidis2018precise}, for any vector $t$, $\norm{t}_{2} = \min_{\tau>0} \frac{\tau}{2}+\frac{\norm{t}_{2}^{2}}{2\tau}$:
\begin{align}
    \label{simple-convex}
     \max_{\kappa,\boldsymbol{\mu},\nu,\tau_{2}} \min_{m,\mathbf{w}_{\perp},\mathbf{z},\eta,\tau_{1}}&\bigg\{ \frac{\kappa\tau_{1}}{2}-\frac{\eta\tau_{2}}{2}-\frac{\eta}{2\tau_{2}d}\norm{\boldsymbol{\mu}+\kappa\mathbf{g}}^{2}_{2}+\frac{\kappa}{2\tau_{1}d}\norm{\frac{m}{\sqrt{\rho}}\mathbf{s}+\eta\mathbf{h}-\mathbf{z}}^{2}_{2}\notag\\
     &+\frac{1}{\sdim}\bigg[\loss(\mathbf{z},\mathbf{y})+\reg\left(\frac{m\sqrt{\sdim\tdim}}{\norm{\tilde{\mathbf{v}}}_{2}^{2}}{\tilde{\mathbf{v}}}+\mathbf{w}_{\perp}\right)+\boldsymbol{\mu}^{\top}\Sigma^{1/2}\left(\frac{m\sqrt{\sdim\tdim}}{\norm{\tilde{\mathbf{v}}}_{2}^{2}}{\tilde{\mathbf{v}}}+\mathbf{w}_{\perp}\right)-\nu\tilde{\mathbf{v}}^{\top}\mathbf{w}_{\perp}\bigg]\bigg\}
\end{align}
performing the minimisation over $\mathbf{z}$ and recognizing the Moreau envelope of $\loss(.,\mathbf{y})$:
\begin{align}
     \max_{\kappa,\boldsymbol{\mu},\nu,\tau_{2}} \min_{m,\mathbf{w}_{\perp},\eta,\tau_{1}}&\bigg\{ \frac{\kappa\tau_{1}}{2}-\frac{\eta\tau_{2}}{2}+\frac{1}{\sdim}\mathcal{M}_{\frac{\tau_{1}}{\kappa}\loss(.,\mathbf{y})}\left(\frac{m}{\sqrt{\rho}}\mathbf{s}+\beta\mathbf{h}\right)-\frac{\eta}{2\tau_{2}d}\norm{\boldsymbol{\mu}+\kappa\mathbf{g}}^{2}_{2} \notag \\
     &+\frac{1}{\sdim}\bigg[\reg\left(\frac{m\sqrt{\sdim\tdim}}{\norm{\tilde{\mathbf{v}}}_{2}^{2}}{\tilde{\mathbf{v}}}+\mathbf{w}_{\perp}\right)+\boldsymbol{\mu}^{\top}\Sigma^{1/2}\left(\frac{m\sqrt{\sdim\tdim}}{\norm{\tilde{\mathbf{v}}}_{2}^{2}}{\tilde{\mathbf{v}}}+\mathbf{w}_{\perp}\right)-\nu\tilde{\mathbf{v}}^{\top}\mathbf{w}_{\perp}\bigg]\bigg\}
\end{align}
At this point we have a convex-concave problem. Inverting the min-max order, $\boldsymbol{\mu}$ appears in a well defined strictly convex least-square problem.
\begin{align}
     &\max_{\kappa,\nu,\tau_{2}} \min_{m,\mathbf{w}_{\perp},\eta,\tau_{1}} \frac{\kappa\tau_{1}}{2}-\frac{\eta\tau_{2}}{2}+\frac{1}{\sdim}\mathcal{M}_{\frac{\tau_{1}}{\kappa}\loss(.,\mathbf{y})}\left(\frac{m}{\sqrt{\rho}}\mathbf{s}+\eta\mathbf{h}\right)-\frac{\nu}{d}\tilde{\mathbf{v}}^{\top}\mathbf{w}_{\perp}+\frac{1}{\sdim}\reg\left(\frac{m\sqrt{\sdim\tdim}}{\norm{\tilde{\mathbf{v}}}_{2}^{2}}{\tilde{\mathbf{v}}}+\mathbf{w}_{\perp}\right)\notag\\
     &+\frac{1}{\sdim}\max_{\boldsymbol{\mu}}\left\{-\frac{\eta}{2\tau_{2}}\norm{\boldsymbol{\mu}+\kappa\mathbf{g}}^{2}_{2}+\boldsymbol{\mu}^{\top}\Sigma^{1/2}\left(\frac{m\sqrt{\sdim\tdim}}{\norm{\tilde{\mathbf{v}}}_{2}^{2}}{\tilde{\mathbf{v}}}+\mathbf{w}_{\perp}\right)\right\}
\end{align}
Solving it:
\begin{align}
    &\max_{\boldsymbol{\mu}}\left\{-\frac{\eta}{2\tau_{2}}\norm{\boldsymbol{\mu}+\kappa\mathbf{g}}^{2}_{2}+\boldsymbol{\mu}^{\top}\Sigma^{1/2}\left(\frac{m\sqrt{\sdim\tdim}}{\norm{\tilde{\mathbf{v}}}_{2}^{2}}{\tilde{\mathbf{v}}}+\mathbf{w}_{\perp}\right)\right\} \notag\\
    &\boldsymbol{\mu}^{*} = \frac{\tau_{2}}{\eta }\Sigma^{1/2}\left(\frac{m\sqrt{\sdim\tdim}}{\norm{\tilde{\mathbf{v}}}_{2}^{2}}{\tilde{\mathbf{v}}}+\mathbf{w}_{\perp}\right)-\kappa\mathbf{g} \notag\\
    &\mbox{with optimal cost} \thickspace \frac{\tau_{2}}{2\eta }\norm{\Sigma^{1/2}\left(\frac{m\sqrt{\sdim\tdim}}{\norm{\tilde{\mathbf{v}}}_{2}^{2}}{\tilde{\mathbf{v}}}+\mathbf{w}_{\perp}\right)}_{2}^{2}-\kappa\mathbf{g}^{\top}\Sigma^{1/2}\left(\frac{m\sqrt{\sdim\tdim}}{\norm{\tilde{\mathbf{v}}}_{2}^{2}}{\tilde{\mathbf{v}}}+\mathbf{w}_{\perp}\right)
\end{align}
remembering that $\Sigma = \Omega-\tilde{\mathbf{v}}\tilde{\mathbf{v}}^{\top}/(p\rho)$ and $\mathbf{w}_{\perp}\perp\tilde{\mathbf{v}}$, the optimal cost of this least-square problem simplifies to:
\begin{align}
    c^{*} = \frac{\tau_{2}}{2\eta }\left(\norm{\Omega^{1/2}\left(\frac{m\sqrt{\sdim\tdim}}{\norm{\tilde{\mathbf{v}}}_{2}^{2}}{\tilde{\mathbf{v}}}+\mathbf{w}_{\perp}\right)}_{2}^{2}-\frac{m^{2}}{\rho}d\right)-\kappa\mathbf{g}^{\top}\Sigma^{1/2}\left(\frac{m\sqrt{\sdim\tdim}}{\norm{\tilde{\mathbf{v}}}_{2}^{2}}{\tilde{\mathbf{v}}}+\mathbf{w}_{\perp}\right)
\end{align}
The (AO) then reads :
\begin{align}
\label{inter-ridge}
     \max_{\kappa,\nu,\tau_{2}} \min_{m,\mathbf{w}_{\perp},\eta,\tau_{1}}&\bigg\{ \frac{\kappa\tau_{1}}{2}-\frac{\eta\tau_{2}}{2}+\frac{1}{\sdim}\mathcal{M}_{\frac{\tau_{1}}{\kappa}\loss(.,\mathbf{y})}\left(\frac{m}{\sqrt{\rho}}\mathbf{s}+\eta\mathbf{h}\right)-\frac{\tau_{2}}{2\eta}\frac{m^{2}}{\rho}-\frac{\nu}{d}\tilde{\mathbf{v}}^{\top}\mathbf{w}_{\perp}\notag\\
     &\hspace{-3cm}\quad+\frac{1}{\sdim}\reg\left(\frac{m\sqrt{\sdim\tdim}}{\norm{\tilde{\mathbf{v}}}_{2}^{2}}{\tilde{\mathbf{v}}}+\mathbf{w}_{\perp}\right)+\frac{\tau_{2}}{2\eta d}\norm{\Omega^{1/2}\left(\frac{m\sqrt{\sdim\tdim}}{\norm{\tilde{\mathbf{v}}}_{2}^{2}}{\tilde{\mathbf{v}}}+\mathbf{w}_{\perp}\right)}_{2}^{2}-\frac{\kappa}{d}\mathbf{g}^{\top}\Sigma^{1/2}\left(\frac{m\sqrt{\sdim\tdim}}{\norm{\tilde{\mathbf{v}}}_{2}^{2}}{\tilde{\mathbf{v}}}+\mathbf{w}_{\perp}\right)\bigg\}
\end{align}
We now need to solve in $\mathbf{w}_{\perp}$. To do so, we can replace $\reg$ with its convex conjugate and solve the least-square problem in $\mathbf{w}_{\perp}$. This will lead to a Moreau envelope of $\reg^{*}$ in the introduced dual variable, which can be linked to the Moreau envelope of $\reg$ by Moreau decomposition. Intuitively, it is natural to think that the corresponding primal variable will be $\frac{m\sqrt{\sdim\tdim}}{\norm{\tilde{\mathbf{v}}}_{2}^{2}}{\tilde{\mathbf{v}}}+\mathbf{w}_{\perp} = \mathbf{w}$ for any feasible $m,\mathbf{w}_{\perp}$. However, we would like to have an explicit follow-up of the variables we optimize on, as we had for the Moreau envelpe of $g$ which is defined with $\mathbf{z}$, so we prefer to introduce a slack variable $\mathbf{w}' = \frac{m\sqrt{\sdim\tdim}}{\norm{\tilde{\mathbf{v}}}_{2}^{2}}{\tilde{\mathbf{v}}}+\mathbf{w}_{\perp}$ with corresponding dual parameter $\boldsymbol{\eta}$ to show that the (AO) can be reformulated in terms of the original variable $\mathbf{w}$. Note that the feasibility set on $\mathbf{w}'$ is almost surely compact.
\begin{align}
     &\max_{\kappa,\nu,\tau_{2},\boldsymbol{\eta}} \min_{m,\mathbf{w}_{\perp},\mathbf{w}',\eta,\tau_{1}} \frac{\kappa\tau_{1}}{2}-\frac{\eta\tau_{2}}{2}+\frac{1}{\sdim}\mathcal{M}_{\frac{\tau_{1}}{\kappa}\loss(.,\mathbf{y})}\left(\frac{m}{\sqrt{\rho}}\mathbf{s}+\eta\mathbf{h}\right)+\frac{1}{\sdim}\reg(\mathbf{w}')-\frac{1}{\sdim}\boldsymbol{\eta}^{T}\mathbf{w}'-\frac{\tau_{2}}{2\eta}\frac{m^{2}}{\rho}\notag\\
     &-\frac{\nu}{d}\tilde{\mathbf{v}}^{\top}\mathbf{w}_{\perp}+\frac{\tau_{2}}{2\eta d}\norm{\Omega^{1/2}\left(\frac{m\sqrt{\sdim\tdim}}{\norm{\tilde{\mathbf{v}}}_{2}^{2}}{\tilde{\mathbf{v}}}+\mathbf{w}_{\perp}\right)}_{2}^{2}-\frac{\kappa}{d}\mathbf{g}^{\top}\Sigma^{1/2}\left(\frac{m\sqrt{\sdim\tdim}}{\norm{\tilde{\mathbf{v}}}_{2}^{2}}{\tilde{\mathbf{v}}}+\mathbf{w}_{\perp}\right)+\frac{1}{\sdim}\boldsymbol{\eta}^{\top}\left(\frac{m\sqrt{\sdim\tdim}}{\norm{\tilde{\mathbf{v}}}_{2}^{2}}{\tilde{\mathbf{v}}}+\mathbf{w}_{\perp}\right)
\end{align}
Isolating the terms depending on $\mathbf{w}_{\perp}$, we get a strictly convex least-square problem, remembering that $\Omega \in \mathcal{S}_{d}^{++}$:
\begin{align}
     &\max_{\kappa,\nu,\tau_{2},\boldsymbol{\eta}} \min_{m,\mathbf{w}_{\perp},\mathbf{w}',\eta,\tau_{1}} \frac{\kappa\tau_{1}}{2}-\frac{\eta\tau_{2}}{2}+\frac{1}{\sdim}\mathcal{M}_{\frac{\tau_{1}}{\kappa}\loss(.,\mathbf{y})}\left(\frac{m}{\sqrt{\rho}}\mathbf{s}+\eta\mathbf{h}\right)+\frac{1}{\sdim}\reg(\mathbf{w}')-\frac{1}{\sdim}\boldsymbol{\eta}^{T}\mathbf{w}'-\frac{\tau_{2}}{2\eta}\frac{m^{2}}{\rho}+\boldsymbol{\eta}^{\top}\frac{m\sqrt{\kappa_{2}}}{\norm{\tilde{\mathbf{v}}}_{2}^{2}}{\tilde{\mathbf{v}}}\notag\\
     &-\kappa\mathbf{g}^{\top}\Sigma^{1/2}\frac{m\sqrt{\gamma}}{\norm{\tilde{\mathbf{v}}}_{2}^{2}}{\tilde{\mathbf{v}}}-\frac{\nu}{d}\tilde{\mathbf{v}}^{\top}\mathbf{w}_{\perp}+\frac{\tau_{2}}{2\eta d}\norm{\Omega^{1/2}\left(\frac{m\sqrt{\sdim\tdim}}{\norm{\tilde{\mathbf{v}}}_{2}^{2}}{\tilde{\mathbf{v}}}+\mathbf{w}_{\perp}\right)}_{2}^{2}-\frac{\kappa}{d}\mathbf{g}^{\top}\Sigma^{1/2}\mathbf{w}_{\perp}+\frac{1}{\sdim}\boldsymbol{\eta}^{\top}\mathbf{w}_{\perp} \\
     &\max_{\kappa,\nu,\tau_{2},\boldsymbol{\eta}} \min_{m,\mathbf{w}',\eta,\tau_{1}} \frac{\kappa\tau_{1}}{2}-\frac{\eta\tau_{2}}{2}+\frac{1}{\sdim}\mathcal{M}_{\frac{\tau_{1}}{\kappa}\loss(.,\mathbf{y})}\left(\frac{m}{\sqrt{\rho}}\mathbf{s}+\eta\mathbf{h}\right)+\frac{1}{\sdim}\reg(\mathbf{w}')-\frac{1}{\sdim}\boldsymbol{\eta}^{T}\mathbf{w}'-\frac{\tau_{2}}{2\eta}\frac{m^{2}}{\rho}+\boldsymbol{\eta}^{\top}\frac{m\sqrt{\kappa_{2}}}{\norm{\tilde{\mathbf{v}}}_{2}^{2}}{\tilde{\mathbf{v}}}\notag \\
     &-\kappa\mathbf{g}^{\top}\Sigma^{1/2}\frac{m\sqrt{\gamma}}{\norm{\tilde{\mathbf{v}}}_{2}^{2}}{\tilde{\mathbf{v}}}+\frac{1}{\sdim}\bigg[\min_{\mathbf{w}_{\perp}}\frac{\tau_{2}}{2\eta}\norm{\Omega^{1/2}\left(\frac{m\sqrt{\sdim\tdim}}{\norm{\tilde{\mathbf{v}}}_{2}^{2}}{\tilde{\mathbf{v}}}+\mathbf{w}_{\perp}\right)}_{2}^{2}-\mathbf{w}_{\perp}^{\top}\left(\kappa\Sigma^{1/2}\mathbf{g}-\boldsymbol{\eta}+\nu\tilde{\mathbf{v}}\right)\bigg]
\end{align}

The quantity $\mathbf{g}^{\top}\Sigma^{1/2}\mathbf{w}_{\perp}$ is a Gaussian random variable with variance $\norm{\Sigma^{1/2}\mathbf{w}_{\perp}}^{2}_{2} = \mathbf{w}_{\perp}^{\top}(\Omega-\tilde{\mathbf{v}}\tilde{\mathbf{v}}^{\top}/(p\rho))\mathbf{w}_{\perp} = \mathbf{w}_{\perp}\Omega\mathbf{w}_{\perp}=\norm{\Omega^{1/2}\mathbf{w}_{\perp}}_{2}^{2}$ using the expression of $\Sigma$ and the orthogonality of $\mathbf{w}_{\perp}$ with respect to $\tilde{\mathbf{v}}$. We can thus change $\Sigma^{1/2}$ for $\Omega^{1/2}$ in front of $\mathbf{w}_{\perp}$ combined with $\mathbf{g}$. 
The least-square problem, its solution and optimal cost then read:
\begin{align}
   & \min_{\mathbf{w}_{\perp}}\frac{\tau_{2}}{2\eta}\norm{\Omega^{1/2}\left(\frac{m\sqrt{\sdim\tdim}}{\norm{\tilde{\mathbf{v}}}_{2}^{2}}{\tilde{\mathbf{v}}}+\mathbf{w}_{\perp}\right)}_{2}^{2}-\mathbf{w}_{\perp}^{\top}\left(\kappa\Omega^{1/2}\mathbf{g}-\boldsymbol{\eta}+\nu\tilde{\mathbf{v}}\right) \\
    &\mathbf{w}_{\perp}^{*} = \frac{\eta}{\tau_{2}}\Omega^{-1}\left(\kappa\Omega^{1/2}\mathbf{g}-\boldsymbol{\eta}+\nu\mathbf{v}\right)-\frac{m\sqrt{\sdim\tdim}}{\norm{\tilde{\mathbf{v}}}_{2}^{2}}\tilde{\mathbf{v}} \\
     &\mbox{with optimal cost} \thickspace -\frac{\eta}{2\tau_{2}}\left(\kappa\Omega^{1/2}\mathbf{g}-\boldsymbol{\eta}+\nu\tilde{\mathbf{v}}\right)^{\top}\Omega^{-1}\left(\kappa\Omega^{1/2}\mathbf{g}-\boldsymbol{\eta}+\nu\tilde{\mathbf{v}}\right)+\frac{m\sqrt{\sdim\tdim}}{\norm{\tilde{\mathbf{v}}}_{2}^{2}}\tilde{\mathbf{v}}^{\top}\left(\kappa\Omega^{1/2}\mathbf{g}-\boldsymbol{\eta}+\nu\tilde{\mathbf{v}}\right)
\end{align}
replacing in the (AO) and simplifying :
\begin{align}
&\iff \max_{\kappa,\nu,\tau_{2},\boldsymbol{\eta}} \min_{m,\mathbf{w}',\eta,\tau_{1}} \frac{\kappa\tau_{1}}{2}-\frac{\eta\tau_{2}}{2}+\frac{1}{\sdim}\mathcal{M}_{\frac{\tau_{1}}{\kappa}\loss(.,\mathbf{y})}\left(\frac{m}{\sqrt{\rho}}\mathbf{s}+\eta\mathbf{h}\right)+\frac{1}{\sdim}\reg(\mathbf{w}')-\frac{1}{\sdim}\boldsymbol{\eta}^{T}\mathbf{w}'-\frac{\tau_{2}}{2\eta}\frac{m^{2}}{\rho} \notag\\
     &-\kappa\mathbf{g}^{\top}\left(\Sigma^{1/2}-\Omega^{1/2}\right)\frac{m\sqrt{\gamma}}{\norm{\tilde{\mathbf{v}}}_{2}^{2}}{\tilde{\mathbf{v}}}-\frac{\eta}{2\tau_{2}d}\left(\kappa\Omega^{1/2}\mathbf{g}-\boldsymbol{\eta}+\nu\tilde{\mathbf{v}}\right)^{\top}\Omega^{-1}\left(\kappa\Omega^{1/2}\mathbf{g}-\boldsymbol{\eta}+\nu\tilde{\mathbf{v}}\right)+m\nu\sqrt{\gamma}
\end{align}
Another strictly convex least-square problem appears on $\boldsymbol{\eta}$, the solution and optimal value of which read
\begin{align}
    &\boldsymbol{\eta}^{*} = -\frac{\tau_{2}}{\eta}\Omega \mathbf{w}'+(\kappa\Omega^{1/2}\mathbf{g}+\nu\tilde{\mathbf{v}}) \\
    &\mbox{with optimal cost} \thickspace \frac{\tau_{2}}{2\eta d}\mathbf{w}'^{\top}\Omega\mathbf{w}'-\mathbf{w}'^{\top}(\kappa\Omega^{1/2}\mathbf{g}+\nu\tilde{\mathbf{v}})
\end{align}
At this point we have expressed feasible solutions of $\boldsymbol{\eta},\mathbf{w}_{\perp}$ as functions of the remaining variables. For any feasible solution in those variables, $\mathbf{w}$ and $\mathbf{w}'$ are the same. Replacing in the (AO) and a completion of squares leads to 
\begin{align}
     &\max_{\kappa,\nu,\tau_{2}} \min_{m,\eta,\tau_{1}} \frac{\kappa\tau_{1}}{2}-\frac{\eta\tau_{2}}{2}+m\nu\sqrt{\gamma}-\frac{\tau_{2}}{2\eta}\frac{m^{2}}{\rho}-\frac{\eta}{2\tau_{2}d}(\nu\tilde{\mathbf{v}}+\kappa\Omega^{1/2}\mathbf{g})^{\top}\Omega^{-1}(\nu\tilde{\mathbf{v}}+\kappa\Omega^{1/2}\mathbf{g})\notag\\
     &-\kappa\mathbf{g}^{\top}\left(\Sigma^{1/2}-\Omega^{1/2}\right)\frac{m\sqrt{\gamma}}{\norm{\tilde{\mathbf{v}}}_{2}^{2}}{\tilde{\mathbf{v}}}+\min_{\mathbf{w}'}\left\{\reg(\mathbf{w}')+\frac{\tau_{2}}{2\eta}\norm{\Omega^{1/2}\mathbf{w}'-\frac{\eta}{\tau_{2}}(\nu\Omega^{-1/2}\tilde{\mathbf{v}}+\kappa\mathbf{g}))}_{2}^{2}\right\}
\end{align}
Recognizing the Moreau envelope of $f$ and introducing the variable $\tilde{\mathbf{w}}=\Omega^{1/2}\mathbf{w}'=\Omega^{1/2}\mathbf{w}$, it follows:
\begin{align}
\label{final_finite_AO}
     &\max_{\kappa,\nu,\tau_{2}} \min_{m,\eta,\tau_{1}} \frac{\kappa\tau_{1}}{2}-\frac{\eta\tau_{2}}{2}+m\nu\sqrt{\gamma}-\frac{\tau_{2}}{2\eta}\frac{m^{2}}{\rho}-\frac{\eta}{2\tau_{2}d}(\nu\tilde{\mathbf{v}}+\kappa\Omega^{1/2}\mathbf{g})^{\top}\Omega^{-1}(\nu\tilde{\mathbf{v}}+\kappa\Omega^{1/2}\mathbf{g})\notag\\
     &-\kappa\mathbf{g}^{\top}\left(\Sigma^{1/2}-\Omega^{1/2}\right)\frac{m\sqrt{\gamma}}{\norm{\tilde{\mathbf{v}}}_{2}^{2}}{\tilde{\mathbf{v}}}+\frac{1}{\sdim}\mathcal{M}_{\frac{\tau_{1}}{\kappa}\loss(.,\mathbf{y})}\left(\frac{m}{\sqrt{\rho}}\mathbf{s}+\eta\mathbf{h}\right)+\frac{1}{\sdim}\mathcal{M}_{\frac{\eta}{\tau_{2}}\reg(\Omega^{-1/2}.)}\left(\frac{\eta}{\tau_{2}}\left(\nu\Omega^{-1/2}\tilde{\mathbf{v}}+\kappa\mathbf{g}\right)\right)
\end{align}
where the Moreau envelopes of $f$ and $g$ are respectively defined w.r.t. the variables $\mathbf{w''}$ and $\mathbf{z}$. At this point we have reduced the initial high-dimensional minimisation problem (\ref{AO_init}) to a scalar
problem over six parameters. Another follow-up of the feasibility set shows that there exist positive constants $C_{m},C_{\kappa},C_{\eta}$ independent of $n,p,d$ such that $0\leqslant \kappa\leqslant C_{\kappa}$, $0\leqslant \eta \leqslant C_{\eta}$ and $0 \leqslant m \leqslant C_{m}$. 
\paragraph{Case 2: $\vec{\theta}_{0}\in\mbox{Ker}(\Phi^{\top})$}
In this case, the min-max problem (\ref{inter-PO}) becomes:
\begin{align}
    \min_{\mathbf{w},\mathbf{z}}\max_{\boldsymbol{\lambda}} \boldsymbol{\lambda}^{\top}\frac{1}{\sqrt{\sdim}}Z\Omega^{1/2}\mathbf{w}-\boldsymbol{\lambda}^{\top}\mathbf{z}+\loss(\mathbf{z},\mathbf{y})+f(\mathbf{w})
\end{align}
Since $\Omega$ is positive definite, we can define $\tilde{\mathbf{w}} = \Omega^{1/2}\mathbf{w}$ and write the equivalent problem:
\begin{align}
    \min_{\tilde{\mathbf{w}},\mathbf{z}}\max_{\boldsymbol{\lambda}} \boldsymbol{\lambda}^{\top}\frac{1}{\sqrt{\sdim}}Z\tilde{\mathbf{w}}-\boldsymbol{\lambda}^{\top}\mathbf{z}+\loss(\mathbf{z},\mathbf{y})+f(\Omega^{-1/2}\tilde{\mathbf{w}})
\end{align}
where the compactness of the feasibility set is preserved almost surely from the almost sure boundedness of the eigenvalues of $\Omega$. We can thus write the corresponding auxiliary optimization problem,
reintroducing the normalization by d:
\begin{align}
    \min_{\tilde{\mathbf{w}},\mathbf{z}}\max_{\boldsymbol{\lambda}} \frac{1}{\sdim}\left[\norm{\boldsymbol{\lambda}}_{2}\frac{1}{\sqrt{\sdim}}\mathbf{g}^{\top}\tilde{\mathbf{w}}+\norm{\tilde{\mathbf{w}}}_{2}\frac{1}{\sqrt{\sdim}}\mathbf{h}^{\top}\boldsymbol{\lambda}-\boldsymbol{\lambda}^{\top}\mathbf{z}+\loss(\mathbf{z},\mathbf{y})+f(\Omega^{-1/2}\tilde{\mathbf{w}})\right]
\end{align}
introducing the convex conjugate of $f$ with dual parameter $\boldsymbol{\eta}$:
\begin{align}
    \min_{\tilde{\mathbf{w}},\mathbf{z}}\max_{\boldsymbol{\lambda},\boldsymbol{\eta}} \frac{1}{\sdim}\left[\norm{\boldsymbol{\lambda}}_{2}\frac{1}{\sqrt{\sdim}}\mathbf{g}^{\top}\tilde{\mathbf{w}}+\norm{\mathbf{w}_{\perp}}_{2}\frac{1}{\sqrt{\sdim}}\mathbf{h}^{\top}\boldsymbol{\lambda}-\boldsymbol{\lambda}^{\top}\mathbf{z}+\loss(\mathbf{z},\mathbf{y})+\boldsymbol{\eta}^{\top}\Omega^{-1/2}\tilde{\mathbf{w}}-f^{*}(\boldsymbol{\eta})\right]
\end{align}
We then define the scalar quantities $\kappa = \frac{\norm{\boldsymbol{\lambda}}_{2}}{\sqrt{\sdim}}$ and $\eta = \frac{\norm{\tilde{\mathbf{w}}}_{2}}{\sqrt{\sdim}}$ and perform the linear optimization on $\boldsymbol{\lambda},\tilde{\mathbf{w}}$,
giving the equivalent:
\begin{align}
    \min_{\mathbf{z},\eta\geqslant0}\max_{\boldsymbol{\eta},\kappa\geqslant0} -\frac{\eta}{\sqrt{\sdim}}\norm{\kappa\mathbf{g}-\Omega^{-1/2}\boldsymbol{\eta}}_{2}+\frac{\kappa}{\sqrt{\sdim}}\norm{\eta\mathbf{h}-\mathbf{z}}_{2}+\frac{1}{\sdim}\loss(\mathbf{z},\mathbf{y})-\frac{1}{\sdim}f^{*}(\boldsymbol{\eta})
\end{align}
Using the square root trick with parameters $\tau_{1},\tau_{2}$:
\begin{align}
    \min_{\tau_{1}>0,\mathbf{z},\eta\geqslant0}\max_{\tau_{2}>0,\boldsymbol{\eta},\kappa\geqslant0} -\frac{\eta\tau_{2}}{2}-\frac{\eta}{2\tau_{2}d}\norm{\kappa\mathbf{g}-\Omega^{-1/2}\boldsymbol{\eta}}^{2}_{2}+\frac{\kappa\tau_{1}}{2}+\frac{\kappa}{2\tau_{1}d}\norm{\eta\mathbf{h}-\mathbf{z}}^{2}_{2}&+\frac{1}{\sdim}\loss(\mathbf{z},\mathbf{y})-\frac{1}{\sdim}f^{*}(\boldsymbol{\eta})
\end{align}
performing the optimizations on $\mathbf{z},\boldsymbol{\eta}$ and recognizing the Moreau envelopes, the problem becomes:
\begin{align}
    &\min_{\tau_{1}>0,\eta\geqslant0}\max_{\tau_{2}>0,\kappa\geqslant0} -\frac{\eta\tau_{2}}{2}+\frac{\kappa\tau_{1}}{2}+\frac{1}{\sdim}\mathcal{M}_{\frac{\tau_{1}}{\kappa}\loss(.,\mathbf{y})}(\eta\mathbf{h})-\frac{1}{\sdim}\mathcal{M}_{\frac{\tau_{2}}{\eta}f^{*}(\Omega^{1/2}.)}(\kappa\mathbf{g}) \\
    \iff &\min_{\tau_{1}>0,\eta\geqslant0}\max_{\tau_{2}>0,\kappa\geqslant0} -\frac{\eta\tau_{2}}{2}+\frac{\kappa\tau_{1}}{2}+\frac{1}{\sdim}\mathcal{M}_{\frac{\tau_{1}}{\kappa}\loss(.,\mathbf{y})}(\eta\mathbf{h})+\frac{1}{\sdim}\mathcal{M}_{\frac{\eta}{\tau_{2}}f(\Omega^{-1/2}.)}(\frac{\eta}{\tau_{2}}\kappa\mathbf{g})-\frac{\eta}{2\tau_{2}d}\kappa^{2}\mathbf{g}^{\top}\mathbf{g}
\end{align}
This concludes the proof of Lemma \ref{lemma:scalar-eq}. \qed
\subsection{Study of the scalar equivalent problem : geometry and asymptotics.}
Here we study the geometry, solutions and asymptotics of the scalar optimization problem (\ref{final_finite_AO}). We will focus on the case $\vec{\theta}_{0} \notin \mbox{Ker}(\Phi^{\top})$ as the other case simply shows that no learning is performed (see the remark at the end of this section). The following lemma 
characterizes the continuity and geometry of the cost function $\mathcal{E}_{n}$.
\begin{lemma}(Geometry of $\mathcal{E}_{n}$)
    \label{scalar-eq-geom}
Recall the function:
\begin{align}
    &\mathcal{E}_{n}(\tau_{1},\tau_{2},\kappa,\eta,\nu,m) = \frac{\kappa\tau_{1}}{2}-\frac{\eta\tau_{2}}{2}+m\nu\sqrt{\gamma}-\frac{\tau_{2}}{2\eta}\frac{m^{2}}{\rho}-\frac{\eta}{2\tau_{2}d}(\nu\tilde{\mathbf{v}}+\kappa\Omega^{1/2}\mathbf{g})^{\top}\Omega^{-1}(\nu\tilde{\mathbf{v}}+\kappa\Omega^{1/2}\mathbf{g})\notag\\
    &-\kappa\mathbf{g}^{\top}\left(\Sigma^{1/2}-\Omega^{1/2}\right)\frac{m\sqrt{\gamma}}{\norm{\tilde{\mathbf{v}}}_{2}^{2}}{\tilde{\mathbf{v}}}+\frac{1}{\sdim}\mathcal{M}_{\frac{\tau_{1}}{\kappa}\loss(.,\mathbf{y})}\left(\frac{m}{\sqrt{\rho}}\mathbf{s}+\eta\mathbf{h}\right)+\frac{1}{\sdim}\mathcal{M}_{\frac{\eta}{\tau_{2}}\reg(\Omega^{-1/2}.)}\left(\frac{\eta}{\tau_{2}}\left(\nu\Omega^{-1/2}\tilde{\mathbf{v}}+\kappa\mathbf{g}\right)\right)
\end{align}
Then $\mathcal{E}_{n}(\tau_{1},\tau_{2},\kappa,\eta,\nu,m)$ is continuous on its domain, jointly convex in $(m,\eta,\tau_{1})$ and jointly concave in $(\kappa,\nu,\tau_{2})$.
\end{lemma}
\emph{Proof of Lemma \ref{scalar-eq-geom}} :
\label{proof-scalar-eq-geom}
$\mathcal{E}_{n}(\tau_{1},\tau_{2},\kappa,\eta,\nu,m)$ is a linear combination of linear and quadratic terms with Moreau envelopes, which are all continuous on their domain.
Remembering the formulation
\begin{align}
    \mathcal{E}_{n}(\tau_{1},\tau_{2},\kappa,\eta,\nu,m)&=\frac{\kappa\tau_{1}}{2}-\frac{\eta\tau_{2}}{2}+m\nu\sqrt{\gamma}-\frac{\tau_{2}}{2\eta}\frac{m^{2}}{\rho}-\kappa\mathbf{g}^{\top}\left(\Sigma^{1/2}-\Omega^{1/2}\right)\frac{m\sqrt{\gamma}}{\norm{\tilde{\mathbf{v}}}_{2}^{2}}{\tilde{\mathbf{v}}}\notag\\
     &+\frac{1}{\sdim}\mathcal{M}_{\frac{\tau_{1}}{\kappa}\loss(.,\mathbf{y})}\left(\frac{m}{\sqrt{\rho}}\mathbf{s}+\eta\mathbf{h}\right)-\frac{1}{\sdim}\mathcal{M}_{\frac{\tau_{2}}{\eta}f^{*}(\Omega^{1/2}.)}\left(\Omega^{-1/2}\left(\nu\tilde{\mathbf{v}}+\kappa\Omega^{1/2}\mathbf{g}\right)\right)
\end{align}
and using the properties of Moreau envelopes, $\mathcal{M}_{\frac{\tau_{1}}{\kappa}\loss(.,\mathbf{y})}\left(\frac{m}{\sqrt{\rho}}\mathbf{s}+\eta\mathbf{h}\right)$ is jointly convex in $(\kappa,\tau_{1},m,\eta)$ as a composition of convex functions of those arguments.
The same applies for $\mathcal{M}_{\frac{\tau_{2}}{\eta}f^{*}(\Omega^{1/2}.)}\left(\Omega^{-1/2}\left(\nu\tilde{\mathbf{v}}+\kappa\Omega^{1/2}\mathbf{g}\right)\right)$, jointly convex in $(\tau_{2},\eta,\nu,\kappa)$, and its opposite is jointly concave in those parameters.
The remaining terms being linear in $\tau_{1},\tau_{2},\nu$, we conclude that $\mathcal{E}_{n}(\tau_{1},\tau_{2},\kappa,\eta,\nu,m)$ is jointly concave in $(\nu,\tau_{2})$ and convex in $\tau_{1}$ whatever the values of $(\kappa,\eta,m)$.
Going back to equation $(\ref{simple-convex})$, we can write
\begin{align}
    \mathcal{E}_{n}(\tau_{1},\tau_{2},\kappa,\eta,\nu,m) &= \max_{\boldsymbol{\mu}}\min_{\mathbf{z},\mathbf{w}_{\perp}} \frac{\kappa\tau_{1}}{2}-\frac{\eta\tau_{2}}{2}-\frac{\eta}{2\tau_{2}d}\norm{\boldsymbol{\mu}+\kappa\mathbf{g}}^{2}_{2}+\frac{\kappa}{2\tau_{1}d}\norm{\frac{m}{\sqrt{\rho}}\mathbf{s}+\eta\mathbf{h}-\mathbf{z}}^{2}_{2}\notag\\
    &+\frac{1}{\sdim}\bigg[\loss(\mathbf{z},\mathbf{y})+f\left(\frac{m\sqrt{\sdim\tdim}}{\norm{\tilde{\mathbf{v}}}_{2}^{2}}{\tilde{\mathbf{v}}}+\mathbf{w}_{\perp}\right)+\boldsymbol{\mu}^{\top}\Sigma^{1/2}\left(\frac{m\sqrt{\sdim\tdim}}{\norm{\tilde{\mathbf{v}}}_{2}^{2}}{\tilde{\mathbf{v}}}+\mathbf{w}_{\perp}\right)-\nu\tilde{\mathbf{v}}^{\top}\mathbf{w}_{\perp}\bigg]
\end{align}
The squared term in $m,\eta,\mathbf{z}$ can be written as
\begin{equation}
    \frac{\kappa}{2\tau_{1}d}\norm{\frac{m}{\sqrt{\rho}}\mathbf{s}+\eta\mathbf{h}-\mathbf{z}}^{2}_{2} = \tau_{1}\frac{\kappa}{2d}\norm{\frac{m}{\tau_{1}\sqrt{\rho}}\mathbf{s}+\frac{\eta}{\tau_{1}}\mathbf{h}-\frac{\mathbf{z}}{\tau_{1}}}^{2}_{2}
\end{equation}
which is the perspective function with parameter $\tau_{1}$ of a function jointly convex in $(\mathbf{z},m,\eta)$. Thus it is jointly convex in $(\tau_{1},\mathbf{z},m,\eta)$. Furthermore, the term $f\left(\frac{m\sqrt{\sdim\tdim}}{\norm{\tilde{\mathbf{v}}}_{2}^{2}}{\tilde{\mathbf{v}}}+\mathbf{w}_{\perp}\right)$ is a composition of a convex function with a linear one, thus it is jointly convex in $(m,\mathbf{w}_{\perp})$. The remaining terms in $\tau_{1},\eta,m$ are linear. Since minimisation on convex sets preserves convexity, minimizing with respect to $\mathbf{z},\mathbf{w}_{\perp}$ will lead to a jointly convex function in $(\tau_{1},\eta,m)$. Similarly, the term $-\frac{\eta}{2\tau_{2}d}\norm{\boldsymbol{\mu}+\kappa\mathbf{g}}_{2}^{2}$ is jointly concave in $\tau_{2},\kappa,\boldsymbol{\mu}$, and maximizing over $\boldsymbol{\mu}$ will result in a jointly concave function in $(\tau_{2},\nu,\kappa)$.
We conclude that $\mathcal{E}_{n}(\tau_{1},\tau_{2},\kappa,\eta,\nu,m)$ is jointly convex in $(\tau_{1},m,\eta)$ and jointly concave in $(\kappa,\nu,\tau_{2})$. \qed \\
\quad \\
The next lemma then characterizes the infinite dimensional limit of the scalar optimization problem (\ref{final_finite_AO}), along with the consistency of its
optimal value.
\begin{lemma}(Asymptotics of $\mathcal{E}_{n}$) 
    \label{lemma:asy-scalar-problem}
Recall the following quantities:
    \begin{align}
        \mathcal{L}_{g}(\tau_{1},\kappa,m,\eta) &= \frac{1}{n}\mathbb{E}\left[\mathcal{M}_{\frac{\tau_{1}}{\kappa}\loss(.,\mathbf{y})}\left(\frac{m}{\sqrt{\rho}}\mathbf{s}+\eta\mathbf{h}\right)\right] \thickspace \mbox{where} \thickspace \mathbf{y} = f_{0}(\sqrt{\rho_{p}}\mathbf{s}), \thickspace \mathbf{s}\sim\mathcal{N}(0,\mathbb{I}_{n}) \\
        \mathcal{L}_{\reg}(\tau_{2},\eta,\nu,\kappa) &= \frac{1}{\sdim}\mathbb{E}\left[\mathcal{M}_{\frac{\eta}{\tau_{2}}\reg(\Omega^{-1/2}.)}\left(\frac{\eta}{\tau_{2}}\left(\nu\Omega^{-1/2}\tilde{\mathbf{v}}+\kappa\mathbf{g}\right)\right)\right] \thickspace \mbox{where} \thickspace \tilde{\mathbf{v}} = \Phi^{\top}\vec{\theta}_{0} \\
        \chi &=  \frac{1}{\sdim}\vec{\theta}_{0}^{\top}\Phi\Omega^{-1}\Phi^{\top}\vec{\theta}_{0} \\
        \rho &= \frac{1}{p} \vec{\theta}_{0}^{\top}\Psi\vec{\theta}_{0}
    \end{align} 
and the potential:
\begin{align}
    \label{free-energy-proof}
    \mathcal{E}(\tau_{1},\tau_{2},\kappa,\eta,\nu,m) =  \frac{\kappa\tau_{1}}{2}-\frac{\eta\tau_{2}}{2}&+m\nu\sqrt{\gamma}-\frac{\tau_{2}}{2\eta}\frac{m^{2}}{\rho}-\frac{\eta}{2\tau_{2}}(\nu^{2}\chi+\kappa^{2})+\alpha\mathcal{L}_{g}(\tau_{1},\kappa,m,\eta)+\mathcal{L}_{\reg}(\tau_{2},\eta,\nu,\kappa)
\end{align}
Then:
\begin{equation}
     \max_{\kappa,\nu,\tau_{2}} \min_{m,\eta,\tau_{1}}\mathcal{E}_{n}(\tau_{1},\tau_{2},\kappa,\eta,\nu,m)  \xrightarrow[n,p,d \to \infty]{P} \max_{\kappa,\nu,\tau_{2}} \min_{m,\eta,\tau_{1}} \mathcal{E}(\tau_{1},\tau_{2},\kappa,\eta,\nu,m)
\end{equation}
and $\mathcal{E}(\tau_{1},\tau_{2},\kappa,\eta,\nu,m)$ is continuously differentiable on its domain, jointly convex in $(m,\eta,\tau_{1})$ and jointly concave in $(\kappa,\nu,\tau_{2})$.
\end{lemma}
\emph{Proof of Lemma \ref{lemma:asy-scalar-problem}}: 
\label{proof-l3}
The strong law of large numbers, see e.g. \cite{durrett2019probability} gives $\frac{1}{\sdim}\mathbf{g}^{\top}\mathbf{g} \xrightarrow[d \to \infty]{a.s.} 1$.
Additionally, using assumption (A2) on the summability of $\vec{\theta}_{0}$ and (A3) on the boundedness of the spectrum 
of the covariance matrices, the quantity $\chi = \lim_{d \to \infty}  \frac{1}{\sdim}\vec{\theta}_{0}^{\top}\Phi\Omega^{-1}\Phi^{\top}\vec{\theta}_{0}$ exists and is finite.
Since $\vec{\theta}_{0} \notin \mbox{Ker}(\Phi^{\top})$ and using the non-vanishing signal hypothesis, the quantity $\rho_{\tilde{\mathbf{v}}} =\lim_{d \to \infty} \frac{1}{\sdim}\tilde{\mathbf{v}}^{\top}\tilde{\mathbf{v}}$ exists, is finite and strictly positive.
Then $\kappa\mathbf{g}^{\top}\left(\Sigma^{1/2}-\Omega^{1/2}\right)\frac{m\sqrt{\gamma}}{\norm{\tilde{\mathbf{v}}}_{2}^{2}}\mathbf{v}$ is a centered Gaussian random variable with variance verifying:
\begin{align}
    \mbox{Var}\left[\kappa\mathbf{g}^{\top}\left(\Sigma^{1/2}-\Omega^{1/2}\right)\frac{m\sqrt{\gamma}}{\norm{\tilde{\mathbf{v}}}_{2}^{2}}\tilde{\mathbf{v}}\right] &\leqslant \kappa^{2}\sigma_{max}^{2}\left(\Sigma^{1/2}-\Omega^{1/2}\right)\frac{m^{2}\gamma}{\norm{\tilde{\mathbf{v}}}_{2}^{2}} \notag \\
    &= \kappa^{2}\sigma_{max}^{2}\left(\Sigma^{1/2}-\Omega^{1/2}\right)\frac{m^{2}\gamma}{d\rho_{\tilde{\mathbf{v}}}}
\end{align}
Using lemma \ref{compacity-lemma}, $\kappa$ and $m$ are finitely bounded independently of the dimension $d$. $\gamma,\sigma_{max}\left(\Sigma^{1/2}-\Omega^{1/2}\right)$ are finite.
Thus there exists a finite constant $C$ such that the standard deviation of $\kappa\mathbf{g}^{\top}\left(\Sigma^{1/2}-\Omega^{1/2}\right)\frac{m\sqrt{\gamma}}{\norm{\tilde{\mathbf{v}}}_{2}^{2}}\tilde{\mathbf{v}}$ is smaller than $\sqrt{C}/\sqrt{\sdim}$.
Then, for any $\epsilon>0$:
\begin{align}
    \mathbb{P}\left(\abs{\kappa\mathbf{g}^{\top}\left(\Sigma^{1/2}-\Omega^{1/2}\right)\frac{m\sqrt{\gamma}}{\norm{\tilde{\mathbf{v}}}_{2}^{2}}\tilde{\mathbf{v}}} \geqslant \epsilon\right) &\leqslant \mathbb{P}\left(\abs{\mathcal{N}(0,1)}\geqslant\epsilon\sqrt{\sdim}/\sqrt{C}\right) \notag \\
    &\leqslant \frac{\sqrt{C}}{\epsilon\sqrt{\sdim}}\frac{1}{\sqrt{2\pi}}\exp(-\frac{1}{2}\frac{\epsilon^{2}d}{C})
\end{align}
using the Gaussian tail. The Borel-Cantelli lemma and summability of this tail gives 
\begin{equation}
    \kappa\mathbf{g}^{\top}\left(\Sigma^{1/2}-\Omega^{1/2}\right)\frac{m\sqrt{\gamma}}{\norm{\tilde{\mathbf{v}}}_{2}^{2}}\tilde{\mathbf{v}} \xrightarrow[d \to \infty]{a.s.} 0
\end{equation}
Concentration of the Moreau envelopes of both $f$ and $g$ follows directly from lemma \ref{conc-Mor}. \\
We thus have the pointwise convergence:
\begin{equation}
    \mathcal{E}_{n}(\tau_{1},\tau_{2},\kappa,\eta,\nu,m) \xrightarrow[n,p,d \to \infty]{P} \mathcal{E}(\tau_{1},\tau_{2},\kappa,\eta,\nu,m)
\end{equation}
Since pointwise convergence preserves convexity, $\mathcal{E}(\tau_{1},\tau_{2},\kappa,\eta,\nu,m)$ is jointly convex in $(m,\eta,\tau_{1})$ and jointly concave in $(\kappa,\nu,\tau_{2})$. \\
Now recall the expression of $\mathcal{E}$
\begin{align}
    \mathcal{E}(\tau_{1},\tau_{2},\kappa,\eta,\nu,m) =  \frac{\kappa\tau_{1}}{2}-\frac{\eta\tau_{2}}{2}&+m\nu\sqrt{\gamma}-\frac{\tau_{2}}{2\eta}\frac{m^{2}}{\rho}-\frac{\eta}{2\tau_{2}}(\nu^{2}\chi+\kappa^{2})+\alpha\mathcal{L}_{g}(\tau_{1},\kappa,m,\eta)+\mathcal{L}_{f}(\tau_{2},\eta,\nu,\kappa)
\end{align}
The feasibility sets of $\kappa,\eta,m$ are compact from Lemma \ref{compacity-lemma} and the subsequent follow-up of the feasibility sets. Then, using Proposition 12.32 from \cite{bauschke2011convex}, for fixed $(\tau_{2},\kappa,\eta,\nu,m)$, we have:
\begin{equation}
    \lim_{\tau_{1}\to +\infty}\frac{1}{\sdim}\mathcal{M}_{\frac{\tau_{1}}{\kappa}\loss(.,\mathbf{y})}\left(\frac{m}{\sqrt{\rho}}\mathbf{s}+\eta\mathbf{h}\right) = \frac{1}{\sdim}\inf_{\mathbf{z} \in \mathbb{R}^{\samples}} \loss(\mathbf{z},\mathbf{y})
\end{equation}
which is a finite quantity since $\loss(.,\mathbf{y})$ is a proper, convex function verifying the scaling assumptions \ref{main-assumptions}. Then, since $\kappa>0$, we have:
\begin{equation}
    \lim_{\tau_{1}\to +\infty} \mathcal{E}_{n}(\tau_{1},\tau_{2},\kappa,\eta,\nu,m) = +\infty
\end{equation}
Similarly, for fixed $(\tau_{1},\kappa,\eta,\nu,m)$ and noting that composing $f$ with the positive definite matrix $\Omega^{-1/2}$ does not change its convexity, or it being proper and lower semi-continuous, we get:
\begin{equation}
    \lim_{\tau_{2} \to +\infty}\frac{1}{\sdim}\mathcal{M}_{\frac{\eta}{\tau_{2}}f(\Omega^{-1/2}.)}\left(\frac{\eta}{\tau_{2}}\left(\nu\Omega^{-1/2}\tilde{\mathbf{v}}+\kappa\mathbf{g}\right)\right) = \frac{1}{\sdim}f(0_{d})
\end{equation}
which is also a bounded quantity from the scaling assumptions made on $f$. Since $\beta >0$, we then have:
\begin{equation}
    \lim_{\tau_{2}\to +\infty} \mathcal{E}_{n}(\tau_{1},\tau_{2},\kappa,\eta,\nu,m) = -\infty
\end{equation}
Finally, the limit $\lim_{\nu\to +\infty} \mathcal{E}_{n}(\tau_{1},\tau_{2},\kappa,\eta,\nu,m)$ needs to be checked for both $+\infty$ and $-\infty$ since there is no restriction on the sign of $\nu$. From  the definition of the Moreau envelope, we can write:
\begin{equation}
    \frac{1}{\sdim}\mathcal{M}_{\frac{\eta}{\tau_{2}}f(\Omega^{-1/2}.)}\left(\frac{\eta}{\tau_{2}}\left(\nu\Omega^{-1/2}\tilde{\mathbf{v}}+\kappa\mathbf{g}\right)\right) \leqslant \frac{1}{\sdim}f(0_{d})+\frac{\tau_{2}}{2\eta}\norm{\frac{\eta}{d\tau_{2}}\left(\nu\Omega^{-1/2}\tilde{\mathbf{v}}+\kappa\mathbf{g}\right)}_{2}^{2}
\end{equation}
Thus, for any fixed $(\tau_{1},\tau_{2},m,\kappa,\eta)$:
\begin{align}
    \mathcal{E}_{n}(\tau_{1},\tau_{2},\kappa,\eta,\nu,m) &\leqslant \frac{\kappa\tau_{1}}{2}-\frac{\eta\tau_{2}}{2}+m\nu\sqrt{\gamma}-\kappa\mathbf{g}^{\top}\left(\Sigma^{1/2}-\Omega^{1/2}\right)\frac{m\sqrt{\gamma}}{\norm{\tilde{\mathbf{v}}}_{2}^{2}}{\mathbf{v}}+\frac{1}{\sdim}\mathcal{M}_{\frac{\tau_{1}}{\kappa}\loss(.,\mathbf{y})}\left(\frac{m}{\sqrt{\rho}}\mathbf{s}+\eta\mathbf{h}\right) \notag\\
    &\hspace{10cm}+\frac{1}{\sdim}f(0_{d})
\end{align}
which immediately gives $\lim_{\nu \to -\infty} \mathcal{E}_{n} = -\infty$. Turning to the other limit, remembering that $\mathcal{E}_{n}$ is continuously
differentiable on its domain, we have:
\begin{equation}
    \frac{\partial\mathcal{E}_{n}}{\partial \nu}(\tau_{1},\tau_{2},\kappa,\eta,\nu,m) = m\sqrt{\gamma}-\frac{1}{\sdim}\tilde{\mathbf{v}}^{\top}\Omega^{-1/2}\mbox{prox}_{\frac{\eta}{\tau_{2}}f(\Omega^{-1/2}.)}\left(\frac{\eta}{\tau_{2}}\left(\nu\Omega^{-1/2}\tilde{\mathbf{v}}+\kappa\mathbf{g}\right)\right)
\end{equation}
Thus $\lim_{\nu \to +\infty} \frac{\partial\mathcal{E}_{n}(\tau_{1},\tau_{2},\kappa,\eta,\nu,m)}{\partial \nu} \to -\infty$. Since $\mathcal{E}_{n}$ is continuously differentiable in $\nu$ on $[0,+\infty[$, and from the short argument led above, we have shown
\begin{equation}
    \lim_{\abs{\nu} \to +\infty} \mathcal{E}_{n}(\tau_{1},\tau_{2},\kappa,\eta,\nu,m)=-\infty
\end{equation}
Using similar arguments as in the proof of Lemma \ref{compacity-lemma}, we can now reduce the feasibility set of $\tau_{1},\tau_{2},\nu$ to a compact one.
Then, using the fact that convergence of convex functions on compact sets implies uniform convergence \cite{andersen1982cox}, we obtain 
\begin{equation}
    \max_{\kappa,\nu,\tau_{2}} \min_{m,\eta,\tau_{1}}\mathcal{E}_{n}(\tau_{1},\tau_{2},\kappa,\eta,\nu,m)  \xrightarrow[n,p,d \to +\infty]{P} \max_{\kappa,\nu,\tau_{2}} \min_{m,\eta,\tau_{1}} \mathcal{E}(\tau_{1},\tau_{2},\kappa,\eta,\nu,m)
\end{equation}
which is the desired result. \qed \\
\quad \\
At this point, it is necessary to characterize the set of solutions of the asymptotic minimisation problem (\ref{optminmax}). 
We start with the explicit form of the optimality condition associated to any solution.

\begin{lemma}(Fixed point equations)
\label{lemma:fixed-point-eq}
    The zero-gradient condition of the optimization problem (\ref{optminmax}) prescribes the following set of fixed point equations for any feasible solution:
    \begin{align}
    &\partial_{\kappa} : \tau_{1}=\frac{1}{\sdim}\mathbb{E}\left[\mathbf{g}^{\top}\mbox{prox}_{\frac{\eta}{\tau_{2}}f(\Omega^{-1/2}.)}\left(\frac{\eta}{\tau_{2}}\left(\nu\Omega^{-1/2}\tilde{\mathbf{v}}+\kappa\mathbf{g}\right)\right)\right] \\
    &\partial_{\nu}:m\sqrt{\gamma}=\frac{1}{\sdim}\mathbb{E}\left[\tilde{\mathbf{v}}^{\top}\Omega^{-1/2}\mbox{prox}_{\frac{\eta}{\tau_{2}}f(\Omega^{-1/2}.)}\left(\frac{\eta}{\tau_{2}}\left(\nu\Omega^{-1/2}\tilde{\mathbf{v}}+\kappa\mathbf{g}\right)\right)\right] \\
    &\partial_{\eta}: \tau_{2}=\alpha\frac{\kappa}{\tau_{1}}\eta-\frac{\kappa\alpha}{\tau_{1}n}\mathbb{E}\left[\mathbf{h}^{\top}\mbox{prox}_{\frac{\tau_{1}}{\kappa}\loss(.,\mathbf{y})}\left(\frac{m}{\sqrt{\rho}}\mathbf{s}+\eta\mathbf{h}\right)\right] \\
    &\partial_{\tau_{2}}:\frac{1}{2d}\frac{\tau_{2}}{\eta}\mathbb{E}\left[\norm{\frac{\eta}{\tau_{2}}(\nu\Omega^{-1/2}\tilde{\mathbf{v}}+\kappa\mathbf{g})-\mbox{prox}_{\frac{\eta}{\tau_{2}}f(\Omega^{-1/2}.)}\left(\frac{\eta}{\tau_{2}}\left(\nu\Omega^{-1/2}\tilde{\mathbf{v}}+\kappa\mathbf{g}\right)\right)}_{2}^{2}\right]=\notag\\
    &\frac{\eta}{2\tau_{2}}(\nu^{2}\chi+\kappa^{2})-m\nu\sqrt{\gamma}-\kappa\tau_{1}+\frac{\eta\tau_{2}}{2}+\frac{\tau_{2}}{2\eta}\frac{m^{2}}{\rho}\\
    &\partial_{m}:\nu\sqrt{\gamma}=\alpha\frac{\kappa}{n\tau_{1}}\mathbb{E}\left[(\frac{m}{\eta\rho}\mathbf{h}-\frac{\mathbf{s}}{\sqrt{\rho}})^{\top}\mbox{prox}_{\frac{\tau_{1}}{\kappa}\loss(.,\mathbf{y})}\left(\frac{m}{\sqrt{\rho}}\mathbf{s}+\eta\mathbf{h}\right)\right] \\
    &\partial_{\tau_{1}}:\frac{\tau_{1}^{2}}{2}=\frac{1}{2}\alpha\frac{1}{n}\mathbb{E}\left[\norm{\frac{m}{\sqrt{\rho}}\mathbf{s}+\eta \mathbf{h}-\mbox{prox}_{\frac{\tau_{1}}{\kappa}g(.,y)}\left(\frac{m}{\sqrt{\rho}}\mathbf{s}+\eta\mathbf{h}\right)}_{2}^{2}\right]
\end{align}
This set of equations can be converted to the replica notations using the table (\ref{match-rep-gordon}).
\end{lemma}
\emph{Proof of Lemma \ref{lemma:fixed-point-eq}}:
Using arguments similar to the ones in the proof of Lemma \ref{conc-Mor}, Moreau envelopes and their derivatives verify the necessary conditions of the dominated convergence theorem. Additionally, uniform convergence of the sequence of derivatives can be verified in a straightforward manner as all involved functions are firmly non-expansive and integrated w.r.t. Gaussian measures. We can therefore invert the limits and derivatives, and invert expectations and derivatives.
We can now write explicitly the optimality condition for the scalar problem (\ref{free-energy-proof}), using the expressions for derivatives of Moreau envelopes from Appendix \ref{toolbox-app}. Some algebra and replacing with prescriptions obtained from each partial derivative leads to the set of equations above. \qed \\
\quad \\
\textbf{Remark} : Here we see that the potential function (\ref{free-energy-proof}) can be further 
studied using the fixed point equations (\ref{lemma:fixed-point-eq}) and the relation (\ref{link-prox-mor}). For any optimal $(\tau_{1},\tau_{2},\kappa,\eta,\nu,m)$, it holds that
\begin{align}
    &\mathcal{E}(\tau_{1},\tau_{2},\kappa,\eta,\nu,m) \notag \\
    & = \alpha\frac{1}{n}\mathbb{E}\left[g\left(\mbox{prox}_{\frac{\tau_{1}}{\kappa}g(.,y)}\left(\frac{m}{\sqrt{\rho}}\mathbf{s}+\eta\mathbf{h}\right),\mathbf{y}\right)\right]+\frac{1}{\sdim}\mathbb{E}\left[f\left(\Omega^{-1/2}\mbox{prox}_{\frac{\eta}{\tau_{2}}f(\Omega^{-1/2}.)}\left(\frac{\eta}{\tau_{2}}\left(\nu\Omega^{-1/2}\tilde{\mathbf{v}}+\kappa\mathbf{g}\right)\right)\right)\right]
\end{align}
Finally, we give a strict-convexity and strict-concavity property of the asymptotic potential $\mathcal{E}$ which will be helpful to prove Lemma \ref{uniqueness}.
\begin{lemma}(Strict convexity and strict concavity near minimisers)
\label{lemma:strct_conv}
 Consider the asymptotic potential function $\mathcal{E}(\tau_{1},\tau_{2},\kappa,\eta,\nu,m)$. Then for any fixed $(\eta,m,\tau_{1})$ in
 their feasibility sets, the function \begin{equation}
 \tau_{2},\kappa,\nu \to \mathcal{E}(\tau_{1},\tau_{2},\kappa,\eta,\nu,m)
 \end{equation}
 is jointly strictly concave in $(\tau_{2},\kappa,\nu)$. \newline
 Additionally, consider the set $\mathcal{S}_{\partial_{\nu,\tau_{2}}}$ defined by:
 \begin{align}
     \mathcal{S}_{\partial_{\nu,\tau_{2}}} = \bigg\{\tau_{1},\tau_{2},\kappa,\eta,\nu,m \thickspace \vert \thickspace m\sqrt{\gamma}=\frac{1}{\sdim}\mathbb{E}\left[\tilde{\mathbf{v}}^{T}\Omega^{-1/2}\mbox{prox}_{\frac{\eta}{\tau_{2}}f(\Omega^{-1/2}.)}\left(\frac{\eta}{\tau_{2}}\left(\nu\Omega^{-1/2}\tilde{\mathbf{v}}+\kappa\mathbf{g}\right)\right)\right],\notag \\\frac{1}{2d}\frac{1}{\eta}\mathbb{E}\left[\norm{\mbox{prox}_{\frac{\eta}{\tau_{2}}f(\Omega^{-1/2}.)}\left(\frac{\eta}{\tau_{2}}\left(\nu\Omega^{-1/2}\tilde{\mathbf{v}}+\kappa\mathbf{g}\right)\right)}_{2}^{2}\right]=\frac{\eta}{2}+\frac{1}{2\eta}\frac{m^{2}}{\rho}\bigg\}
 \end{align}
 then for any fixed $\tau_{2},\kappa,\nu$ in $\mathcal{S}_{\partial_{\nu,\tau_{2}}}$, the function $(\eta,m,\tau_{1}) \to \mathcal{E}(\tau_{1},\tau_{2},\kappa,\eta,\nu,m)$ is jointly strictly convex in $(\eta,m,\tau_{1})$ on $\mathcal{S}_{\partial_{\nu,\tau_{2}}}$
\end{lemma}
\emph{Proof of Lemma \ref{lemma:strct_conv}}:\label{proof-uniqueness}
We will use the following first order characterization of strictly convex functions: $
    f \thickspace \mbox{is strictly convex} \iff \langle\mathbf{x}-\mathbf{y}\vert \nabla f(\mathbf{x})-\nabla f(\mathbf{y})\rangle >0 \thickspace \forall \mathbf{x} \neq \mathbf{y} \in \mbox{dom}(f)$.
To simplify notations, we will write, for any fixed $(m,\eta,\tau_{1})$
\begin{align}
    \left(\nabla_{\kappa,\nu,\tau_{2}}\mathcal{E}\right) =\left( \left(\partial_{\kappa}\mathcal{E},\partial_{\nu}\mathcal{E},\partial_{\tau_{2}}\mathcal{E}\right)(\tau_{1},\tau_{2},\kappa,\eta,\nu,m)\right)_{i}
\end{align}
as the i-th component of the gradient of $\mathcal{E}(\tau_{1},\tau_{2},\kappa,\eta,\nu,m)$ with respect to $(\kappa,\nu,\tau_{2})$ for any fixed $(m,\eta,\tau_{1})$ in the feasibility set. 
Then for any distinct triplets $(\kappa,\nu,\tau_{2}),(\tilde{\kappa},\tilde{\nu},\tilde{\tau}_{2})$ and fixed $(\eta,m,\tau_{1})$ in the feasibility set, determining the partial derivatives of $\mathcal{E}$ in similar fashion as is implied in the proof of Lemma \ref{lemma:fixed-point-eq}, we have:
\begin{align}
&\left((\kappa,\nu,\tau_{2})-(\tilde{\kappa},\tilde{\nu},\tilde{\tau}_{2})\right)^{\top}\left(\nabla\mathcal{E}_{\kappa,\nu,\tau_{2}}-\nabla\mathcal{E}_{\tilde{\kappa},\tilde{\nu},\tilde{\tau}_{2}}\right) \notag \\
&=(\kappa-\tilde{\kappa})\alpha\frac{1}{2\tau_{1}}\frac{1}{n}\left(\mathbb{E}\left[\norm{\mathbf{r}_{1}-\mbox{prox}_{\frac{\tau_{1}}{\kappa}g(.,\mathbf{y})}\left(\mathbf{r}_{1}\right)}_{2}^{2}-\norm{\mathbf{r}_{1}-\mbox{prox}_{\frac{\tau_{1}}{\tilde{\kappa}}g(.,\mathbf{y})}\left(\mathbf{r}_{1}\right)}_{2}^{2}\right]\right) \notag \\
&+\left(\mbox{prox}_{\frac{\eta}{\tau_{2}}f(\Omega^{-1/2}.)}\left(\frac{\eta}{\tau_{2}}\mathbf{r}_{2}\right)-\mbox{prox}_{\frac{\eta}{\tilde{\tau}_{2}}f(\Omega^{-1/2}.)}\left(\frac{\eta}{\tilde{\tau}_{2}}\tilde{\mathbf{r}}_{2}\right)\right)^{\top}\bigg(\tilde{\mathbf{r}}_{2}-\mathbf{r}_{2}\notag \\
&\hspace{3cm}+\frac{\tau_{2}-\tilde{\tau}_{2}}{2\eta d}\left(\mbox{prox}_{\frac{\eta}{\tau_{2}}f(\Omega^{-1/2}.)}\left(\frac{\eta}{\tau_{2}}\mathbf{r}_{2}\right)+\mbox{prox}_{\frac{\eta}{\tilde{\tau}_{2}}f(\Omega^{-1/2}.)}\left(\frac{\eta}{\tilde{\tau}_{2}}\tilde{\mathbf{r}}_{2}\right)\right)\bigg) \notag \\
&\leqslant(\kappa-\tilde{\kappa})\alpha\frac{1}{2\tau_{1}}\frac{1}{n}\left(\mathbb{E}\left[\norm{\mathbf{r}_{1}-\mbox{prox}_{\frac{\tau_{1}}{\kappa}g(.,y)}\left(\mathbf{r}_{1}\right)}_{2}^{2}-\norm{\mathbf{r}_{1}-\mbox{prox}_{\frac{\tau_{1}}{\tilde{\kappa}}g(.,y)}\left(\mathbf{r}_{1}\right)}_{2}^{2}\right]\right) \notag\\
&\hspace{3cm}+\left(\frac{(\tau_{2}+\tilde{\tau}_{2})}{2\eta d}\mathbb{E}\left[-\norm{\mbox{prox}_{\frac{\eta}{\tau_{2}}f(\Omega^{-1/2}.)}\left(\frac{\eta}{\tau_{2}}\mathbf{r}_{2}\right)-\mbox{prox}_{\frac{\eta}{\tilde{\tau}_{2}}f(\Omega^{-1/2}.)}\left(\frac{\eta}{\tilde{\tau}_{2}}\tilde{\mathbf{r}}_{2}\right)}_{2}^{2}\right]\right) 
\end{align} 
\noindent where the last line follows from the inequality in Lemma \ref{useful-ineq1}, and we defined the shorthands, $\mathbf{r}_{1} = \frac{m}{\sqrt{\rho}}\mathbf{s}+\eta\mathbf{h}$, $\mathbf{r}_{2} = \nu\Omega^{-1/2}\tilde{\mathbf{v}}+\kappa\mathbf{g},  \tilde{\mathbf{r}}_{2} = \tilde{\nu}\Omega^{-1/2}\mathbf{v}+\tilde{\kappa}\mathbf{g}$. Using Lemma \ref{useful-functions}, the first term of the r.h.s of the last inequality is also negative as an increment of a nonincreasing function. Thus, both expectations are taken on negative functions. If those functions are not zero almost everywhere with respect to the Lebesgue measure, then the result will be strictly negative. Moreover, the functional taking each operator $T$ to its resolvent $(\rm{Id}+T)^{-1}$ is a bijection on the set of non-trivial, maximally monotone operators, see e.g. \cite{bauschke2011convex} Proposition 23.21 and the subsequent discussion. The subdifferential of a proper, closed, convex function being maximally 
monotone, for two different parameters the corresponding proximal operator cannot be equal almost everywhere. The previously studied increment $\left((\kappa,\nu,\tau_{2})-(\tilde{\kappa},\tilde{\nu},\tilde{\tau}_{2})\right)^{\top}\left(\nabla\mathcal{E}_{\kappa,\nu,\tau_{2}}-\nabla\mathcal{E}_{\tilde{\kappa},\tilde{\nu},\tilde{\tau}_{2}}\right)$ is therefore strictly negative, giving the desired strict concavity in $(\kappa,\nu,\tau_{2})$. Restricting ourselves to the set $\mathcal{S}_{\partial \nu,\tau_{2}}$, the increment in $(m,\eta,\tau_{1})$ can be written similarly. Note that $Id-\mbox{prox}$ will appear in the expressions instead of $\mbox{prox}$. The appropriate terms can then be brought to the form of the inequality from Lemma $\ref{useful-ineq1}$ using Moreau's decomposition. Using the definitions of the set $\mathcal{S}_{\partial \nu,\tau_{2}}$ and the increments from Lemma \ref{useful-functions}, a similar argument as the previous one can be carried out.
The lemma is proved.
\qed\\
\quad \\
What is now left to do is link the properties of the scalar optimization problem (\ref{optminmax}) to the original learning  problem (\ref{eq:student}) using the tight inequalities from Theorem $\ref{CGMT}$. \newline
\quad \\
\textbf{Remark}: in the case $\vec{\theta}_{0} \in \mbox{Ker}(\Phi^{T})$, the cost function $\mathcal{E}^{0}_{n}$ will uniformly converge to the following potential:
\begin{align}
-\frac{\eta\tau_{2}}{2}+\frac{\kappa\tau_{1}}{2}-\frac{\eta}{2\tau_{2}}\kappa^{2}+\frac{\alpha}{n}\mathbb{E}\left[\mathcal{M}_{\frac{\tau_{1}}{\kappa}\loss(.,\mathbf{y})}(\eta\mathbf{h})\right]+\frac{1}{\sdim}\mathbb{E}\left[\mathcal{M}_{\frac{\eta}{\tau_{2}}f(\Omega^{-1/2}.)}(\frac{\eta}{\tau_{2}}\kappa\mathbf{g})\right]
\end{align}
As we will see in the next section, this will lead to estimators solely based on noise.
\subsection{Back to the original problem : proof of Theorem \ref{Train_gen} and \ref{main-th}}
\label{proof-connect}
We begin this part by considering that the "necessary assumptions for exponential rates" from the set of assumptions \ref{main-assumptions} are verified. In the end we will discuss how relaxing these assumptions modifies the convergence speed. 
We closely follow the analysis introduced in \cite{miolane2018distribution} and further developed in \cite{celentano2020lasso}. The main difference resides in checking the concentration properties of generic Moreau envelopes depending on the regularity of the target function instead of specific instances such as the LASSO. Since the dimensions $n,p,d$ are linked by multiplicative constants, we can express the rates with any of the three. Recall the original reformulation of the problem defining the student.
\begin{align}
 &\max_{\boldsymbol{\lambda}}\min_{\mathbf{w},\mathbf{z}} \loss(\mathbf{z},\mathbf{y})+f(\mathbf{w})+\boldsymbol{\lambda}^{\top}\left(\frac{1}{\sqrt{\sdim}} \left(A\Psi^{-1/2}\Phi+B\left(\Omega-\Phi^{\top}\Psi^{-1}\Phi\right)^{1/2}\right)\mathbf{w}-\mathbf{z}\right)
\end{align}
     Introducing the variable $\tilde{\mathbf{w}} = \Omega^{1/2}\mathbf{w}$ it can be equivalently written, since $\Omega$ is almost surely invertible and the problem is convex concave with a closed convex feasibility set on $\tilde{\mathbf{w}},\mathbf{z}$.
   \begin{align}
   \label{eq:rem_student}
 &\min_{\tilde{\mathbf{w}},\mathbf{z}}\max_{\boldsymbol{\lambda}} \loss(\mathbf{z},\mathbf{y})+f(\Omega^{-1/2}\tilde{\mathbf{w}})+\boldsymbol{\lambda}^{\top}\left(\frac{1}{\sqrt{\sdim}} \left(A\Psi^{-1/2}\Phi+B\left(\Omega-\Phi^{\top}\Psi^{-1}\Phi\right)^{1/2}\right)\Omega^{-1/2}\tilde{\mathbf{w}}-\mathbf{z}\right)
\end{align}
Recall the equivalent scalar auxiliary problem at finite dimension $\mathcal{E}_{n}$
and its asymptotic counterpart $\mathcal{E}$ both defined on the same variables as the original problem $\tilde{\mathbf{w}},\mathbf{z}$ through the Moreau envelopes of $g$ and $r$:
\begin{align}
\label{eq:rem_alt_student}
    &\mathcal{E}(\tau_{1},\tau_{2},\kappa,\eta,\nu,m)=\frac{\kappa\tau_{1}}{2}-\frac{\eta\tau_{2}}{2}+m\nu\sqrt{\gamma}-\frac{\tau_{2}}{2\eta}\frac{m^{2}}{\rho}-\frac{\eta}{2\tau_{2}}(\nu^{2}\chi+\kappa^{2})+\alpha\mathcal{L}_{g}(\tau_{1},\kappa,m,\eta)+\mathcal{L}_{f}(\tau_{2},\eta,\nu,\kappa) \\
    &\mathcal{E}_{n}(\tau_{1},\tau_{2},\kappa,\eta,\nu,m) = \frac{\kappa\tau_{1}}{2}-\frac{\eta\tau_{2}}{2}+m\nu\sqrt{\gamma}-\frac{\tau_{2}}{2\eta}\frac{m^{2}}{\rho}-\frac{\eta}{2\tau_{2}d}(\nu\tilde{\mathbf{v}}+\kappa\Omega^{1/2}\mathbf{g})^{\top}\Omega^{-1}(\nu\mathbf{v}+\kappa\Omega^{1/2}\mathbf{g})\notag\\
    &-\kappa\mathbf{g}^{\top}\left(\Sigma^{1/2}-\Omega^{1/2}\right)\frac{m\sqrt{\gamma}}{\norm{\tilde{\mathbf{v}}}_{2}^{2}}{\mathbf{v}}+\frac{1}{\sdim}\mathcal{M}_{\frac{\tau_{1}}{\kappa}\loss(.,\mathbf{y})}\left(\frac{m}{\sqrt{\rho}}\mathbf{s}+\eta\mathbf{h}\right)+\frac{1}{\sdim}\mathcal{M}_{\frac{\eta}{\tau_{2}}\reg(\Omega^{-1/2}.)}\left(\frac{\eta}{\tau_{2}}\left(\nu\Omega^{-1/2}\tilde{\mathbf{v}}+\kappa\mathbf{g}\right)\right)
\end{align}
Recall the variables:
\begin{align}
\label{eq:strong_sol}
    \tilde{\mathbf{w}}^{*} = \mbox{\rm prox}_{\frac{\eta^{*}}{\tau_{2}^{*}}f(\Omega^{-1/2}.)}(\frac{\eta^{*}}{\tau_{2}^{*}}(\nu^{*}\mathbf{t}+\kappa^{*}\mathbf{g})), && \mathbf{z}^{*} = \mbox{\rm prox}_{\frac{\tau_{1}^{*}}{\kappa^{*}}\loss(.,\mathbf{y})}\left(\frac{m^{*}}{\sqrt{\rho}}\mathbf{s}+\eta^{*}\mathbf{h}\right)
\end{align}
Denote $(\tau_{1}^{*},\tau_{2}^{*},\kappa^{*},\eta^{*},\nu^{*},m^{*})$ the unique solution to the optimization problem $(\ref{optminmax})$ and $\mathcal{E}^{*}$ the corresponding optimal cost. $\mathcal{E}^{*}$ defines a strongly convex optimization problem (due to the Moreau envelopes) on $\tilde{\mathbf{w}}, \mathbf{z}$ whose solution is given by Eq.\eqref{eq:strong_sol}.  Similarly, denote $(\tau_{1,n}^{*},\tau_{2,n}^{*},\kappa_{n}^{*},\eta_{n}^{*},\nu_{n}^{*},m_{n}^{*})$ any solution to the optimization problem on $\mathcal{E}_{n}$ and $\mathcal{E}^{*}_{n}$ the corresponding optimal value. Finally, we write $E_{n}(\tilde{\mathbf{w}},\mathbf{z})$ the cost function of the optimization problem on $\tilde{\mathbf{w}},\mathbf{z}$ defined by $\mathcal{E}^{*}_{n}$ for any optimal solution $(\tau_{1,n}^{*},\tau_{2,n}^{*},\kappa_{n}^{*},\eta_{n}^{*},\nu_{n}^{*},m_{n}^{*})$, such that:
\begin{align}
    \mathcal{E}_{n}^{*} = \min_{\tilde{\mathbf{w}},\mathbf{z}} E_{n}(\tilde{\mathbf{w}},\mathbf{z})
\end{align}
By the definition of Moreau envelopes, we have that $E_{n}(\tilde{\mathbf{w}},\mathbf{z})$ is $\frac{\kappa_{n}^{*}}{2d\tau_{1,n}^{*}}$ strongly convex in $\mathbf{z}$ and $\frac{\tau_{2,n}^{*}}{2d\eta_{n}^{*}}$ strongly convex in $\tilde{\mathbf{w}}$. The following lemma ensures that these strong convexity constants are non-zero for any finite $n$.
\begin{lemma}
\label{str-conv-lemma}
Consider the finite size scalar optimization problem 
\begin{align}
    \max_{\kappa,\nu,\tau_{2}} \min_{m,\eta,\tau_{1}}\mathcal{E}_{n}(\tau_{1},\tau_{2},\kappa,\eta,\nu,m)
\end{align}
where the feasibility set of $(\tau_{1},\tau_{2},\kappa,\eta,\nu,m)$ is compact and $\tau_{1}>0,\tau_{2}>0$. Then any optimal values $\kappa^{*},\tau_{2}^{*}$ verify:
\begin{equation}
    \kappa^{*} \neq 0 \quad \tau^{*}_{2} \nrightarrow 0
\end{equation}
\end{lemma}
\emph{Proof of Lemma \ref{str-conv-lemma}}: from the analysis carried out in the proof of Lemma \ref{lemma:asy-scalar-problem}, the feasibility set of the optimization problem is compact. Suppose $\kappa^{*}=0$. Then the value of $m$ minimizing the cost function is $-\infty$, which contradicts the compactness of the feasibility set. A similar argument holds for $\tau_{2}$. \qed \\
\quad \\
The next lemma characterizes the speed of convergence of the optimal value of the finite dimensional scalar optimization problem to its asymptotic counterpart, which has a unique solution in $\tau_{1},\tau_{2},\kappa,\eta,\nu,m$. The intuition is that, using the strong convexity of the auxiliary problems, we can show that the solution in $\tilde{\mathbf{w}}, \tilde{\mathbf{z}}$ to the finite size problem $\mathcal{E}^{*}_{n}$ converges to the solution $\tilde{\mathbf{w}}^{*}, \tilde{\mathbf{z}}^{*}$ of the asymptotic problem $\mathcal{E}^{*}$, with convergence rates governed by those of the finite size cost towards its asymptotic counterpart.
\begin{lemma}
\label{lemma:conc-1}
    For any $\epsilon>0$, there exist constants $C,c,\gamma$ such that:
    \begin{align}
        \mathbb{P}\left(\abs{\mathcal{E}^{*}_{n}-\mathcal{E}^{*}}\geqslant \gamma \epsilon\right)\leqslant \frac{C}{\epsilon}\exp^{-cn\epsilon^{2}}
    \end{align}
    which is equivalent to 
     \begin{align}
        \mathbb{P}\left(\abs{\min_{\tilde{\mathbf{w}},\mathbf{z}}E_{n}(\tilde{\mathbf{w}},\mathbf{z})-\mathcal{E}^{*}}\geqslant \gamma \epsilon\right)\leqslant \frac{C}{\epsilon}\exp^{-cn\epsilon^{2}}
    \end{align}
\end{lemma}
\emph{Proof of Lemma \ref{lemma:conc-1}}: for any fixed $(\tau_{1},\tau_{2},\kappa,\nu,\eta,m)$, we can determine the rates of convergence of all the random quantities in $\mathcal{E}_{n}$. The linear terms involving $\frac{1}{d}\mathbf{g}^{T}\mathbf{v}$ are sub-Gaussian with sub-Gaussian norm bounded by $C/d$ for some constant $C>0$. Thus we can find constants, $C,c>0$ such that, for any $\epsilon>0$ :
\begin{equation}
    \mathbb{P}\left(\abs{\frac{1}{d}\mathbf{g}^{T}\tilde{\mathbf{v}}} \geqslant \epsilon\right) \leqslant Ce^{-cn\epsilon^{2}}
\end{equation}
The term involving $\mathbf{v}^{T}\Omega\mathbf{v}$ is deterministic in this setting. We will see in section \ref{relax-teacher} how a random $\boldsymbol{\theta}_{0}$ affects the convergence rates. The term involving $\frac{1}{d}\mathbf{g}^{T}\mathbf{g}$ is a weighted sum of sub-exponential random variables, the tail of which can be determined using Bernstein's inequality, see e.g. \cite{vershynin2018high} Corollary 2.8.3, which gives a sub-Gaussian tail for small deviations and a sub-exponential tail for large deviations. Parametrizing the deviation $\epsilon$ with a scalar variable $c'$, we thus get the following bound : for any $\epsilon>0$, there exists constants $C,c,c'>0$ such that:
\begin{equation}
    \mathbb{P}\left(\abs{\frac{1}{d}\mathbf{g}^{T}\mathbf{g}-1}\geqslant c'\epsilon\right) \leqslant Ce^{-cn\epsilon^{2}}
\end{equation}
Since, in this case, we assume that the eigenvalues of the covariance matrices are bounded with probability one, multiplications by these matrices do not change these two previous rates. The remaining convergence rates that need to be determined are those of the Moreau envelopes. By assumption, the function $g$ is separable, and pseudo-Lipschitz of order two. Moreover, the argument $\frac{m}{\sqrt{\rho}}\mathbf{s}+\eta\mathbf{h}$ is an i.i.d. Gaussian random vector with finite variance. The Moreau envelope $\frac{1}{\sdim}\mathcal{M}_{\frac{\eta}{\tau_{2}}\reg(\Omega^{-1/2}.)}\left(\frac{\eta}{\tau_{2}}\left(\nu\Omega^{-1/2}\tilde{\mathbf{v}}+\kappa\mathbf{g}\right)\right)$ is therefore a sum of pseudo-Lipschitz functions of order 2 of scalar Gaussian random variables. Using the concentration Lemma \ref{conc-pseudo-lip-2}, we can find constants $C,c,\gamma>0$ such that, for any $\epsilon>0$, the following holds:
\begin{equation}
    \mathbb{P}\left(\abs{\alpha\frac{1}{n}\mathcal{M}_{\frac{\tau_{1}}{\kappa}\loss(.,\mathbf{y})}\left(\frac{m}{\sqrt{\rho}}\mathbf{s}+\eta\mathbf{h}\right)-\mathbb{E}\left[\alpha\frac{1}{n}\mathcal{M}_{\frac{\tau_{1}}{\kappa}\loss(.,\mathbf{y})}\left(\frac{m}{\sqrt{\rho}}\mathbf{s}+\eta\mathbf{h}\right)\right]}\geqslant \gamma \epsilon\right)\leqslant Ce^{-cn\epsilon^{2}}
\end{equation}
For the second Moreau envelope, the argument $\frac{\eta}{\tau_{2}}\left(\nu\Omega^{-1/2}\tilde{\mathbf{v}}+\kappa\mathbf{g}\right)$ is not separable. If the regularization is a square, it is the concentration will reduce to that of the terms $\frac{1}{d}\mathbf{g}^{T}\mathbf{v}$ and $\frac{1}{d}\mathbf{g}^{T}\mathbf{g}$. If the regularization is a Lipschitz function, then the Moreau envelope is also Lipschitz from Lemma \ref{Mor-pseudo-Lip}. Furthermore, since the eigenvalues of the covariance matrix $\Omega$ are bounded with probability one, the composition with the deterministic term $\nu\Omega^{1/2}\mathbf{v}$ does not change the Lipschitz property. Gaussian concentration of Lipschitz functions then gives an exponential decay indepedent of the magnitude of the deviation. Taking the loosest bound, which is the one obtained with the square penalty, we obtain that, for any $\epsilon>0$, there exist constants $C,c,\gamma>0$ such that the event
\begin{equation}
    \left\{\abs{\frac{1}{\sdim}\mathcal{M}_{\frac{\eta}{\tau_{2}}\reg(\Omega^{-1/2}.)}\left(\frac{\eta}{\tau_{2}}\left(\nu\Omega^{-1/2}\tilde{\mathbf{v}}+\kappa\mathbf{g}\right)\right)-\mathbb{E}\left[\frac{1}{\sdim}\mathcal{M}_{\frac{\eta}{\tau_{2}}\reg(\Omega^{-1/2}.)}\left(\frac{\eta}{\tau_{2}}\left(\nu\Omega^{-1/2}\tilde{\mathbf{v}}+\kappa\mathbf{g}\right)\right)\right]}\geqslant \gamma \epsilon\right\}
\end{equation}
has probability at most $Ce^{-cn\epsilon^{2}}$.
Combining these bounds gives the exponential rate for the convergence of $\mathcal{E}_{n}$ to $\mathcal{E}$ for any fixed $(\tau_{1},\tau_{2},\kappa,\nu,\eta,m)$. An $\varepsilon$-net argument can then be used to obtain the bound on the minmax values. \qed\\
\quad \\
The next lemma shows that the function $E_{n}$ evaluated at $\tilde{\mathbf{w}}^{*},\mathbf{z}^{*}$ is close to the optimal value $\mathcal{E}^{*}$.
\begin{lemma}
\label{lemma:E-close}
For any $\epsilon>0$, there exist constants $C,c,\gamma$ such that:
\begin{equation}
    \mathbb{P}\left(\abs{E_{n}(\tilde{\mathbf{w}}^{*},\mathbf{z}^{*})-\mathcal{E}^{*}}\geqslant \gamma\epsilon\right)\leqslant Ce^{-cn\epsilon^{2}}
\end{equation}
\end{lemma}
\emph{Proof of Lemma \ref{lemma:E-close}}: this Lemma can be proved in similar fashion to \cite{miolane2018distribution} Theorem B.1. using the strong convexity in $\tilde{\mathbf{w}}$ and $\mathbf{z}$ of $E_{n}(\tilde{\mathbf{w}},\mathbf{z})$ along with Gordon's Lemma. We leave the detail of this part to a longer version of this paper.
\begin{lemma}
For any $\epsilon>0$, there exists constants $\gamma,c,C>0$ such that the event
\begin{align}
\label{lemma:close-minim}
    \exists (\tilde{\mathbf{w}},\mathbf{z}) \in \mathbb{R}^{n+d}, \thickspace \frac{1}{d}\min(\frac{\kappa^{*}_{n}}{2\tau_{1,n}^{*}},\frac{\tau^{*}_{2,n}}{2\eta_{n}^{*}})\norm{(\tilde{\mathbf{w}},\mathbf{z})-(\tilde{\mathbf{w}}^{*},\mathbf{z}^{*})}_{2}^{2}>\epsilon \thickspace \mbox{and} \thickspace  \min_{\tilde{\mathbf{w},\mathbf{z}}}E_{n}(\tilde{\mathbf{w}},\mathbf{z})\leqslant E_{n}(\tilde{\mathbf{w}}^{*},\mathbf{z}^{*})+\gamma\epsilon
\end{align}
has probability at most $\frac{C}{\epsilon}e^{-cn\epsilon^{2}}$.
\end{lemma}
This lemma can be proven using the same arguments as in \cite{miolane2018distribution} Appendix B, Theorem B.1. Intuitively, if two values of a strongly convex function are arbitrarily close, then the corresponding points are arbitrarily close. Note that we are normalizing the norm of a vector of size $(n+d)$ with $d$, which are proportional.
This shows that any solution outside the ball centered around $\tilde{\mathbf{w}}^{*},\mathbf{z}^{*}$ is sub-optimal.
Now define the set:
\begin{equation}
    D_{\tilde{\mathbf{w}},\mathbf{z},\epsilon} = \left\{\tilde{\mathbf{w}}\in \mathbb{R}^{\sdim},\mathbf{z}\in \mathbb{R}^{\samples}: \abs{\phi_{1}(\frac{\tilde{\mathbf{w}}}{\sqrt{\sdim}})-\mathbb{E}\left[\phi_{1}\left(\frac{\tilde{\mathbf{w}}^{*}}{\sqrt{\sdim}}\right)\right]}>\epsilon,\thickspace \abs{\phi_{2}(\frac{\mathbf{z}}{\sqrt{n}})-\mathbb{E}\left[\phi_{2}\left(\frac{\mathbf{z}^{*}}{\sqrt{n}}\right)\right]}>\epsilon\right\}
\end{equation}
where $\phi_{1}$ is either a square or a Lipschitz function, and $\phi_{2}$ is a separable, pseudo-Lipschitz function of order 2. Using the same arguments as in the proof of Lemma \ref{lemma:E-close} and the assumptions on $\phi_{1}, \phi_{2}$, Gaussian concentration will give sub-exponential rates for the event $(\tilde{\mathbf{w}}^{*},\mathbf{z}^{*}) \in D_{\tilde{\mathbf{w}},\mathbf{z},\epsilon}$.
A similar argument to the proof of Lemma B.3 from \cite{celentano2020lasso} then shows that a distance of $\epsilon$ in $D_{\tilde{\mathbf{w}},\mathbf{z},\epsilon}$ results in a distance of $\epsilon^{2}$ in the event \eqref{lemma:close-minim}, leading to the following result:
\begin{lemma}
For any $\epsilon>0$, there exists constants $\gamma,c,C>0$ such that the event
\begin{align}
\label{lemma:close-minim}
    \exists (\tilde{\mathbf{w}},\mathbf{z}) \in \mathbb{R}^{n+d},(\tilde{\mathbf{w}}^{*},\mathbf{z}^{*}) \in D_{\tilde{\mathbf{w}},\mathbf{z},\epsilon} \thickspace \mbox{and} \thickspace  \min_{\tilde{\mathbf{w},\mathbf{z}}}E_{n}(\tilde{\mathbf{w}},\mathbf{z})\leqslant E_{n}(\tilde{\mathbf{w}}^{*},\mathbf{z}^{*})+\gamma\epsilon^{2}
\end{align}
has probability at most $\frac{C}{\epsilon^{2}}e^{-cn\epsilon^{4}}$.
\end{lemma}
which proves Theorem \ref{main-th} using the fact that $\hat{\mathbf{w}},\hat{\mathbf{z}}$ are minimizers of the initial cost function.
Theorem \ref{Train_gen} is a consequence of Theorem \ref{main-th}. \quad \\
\quad \\
If the restriction on $f,g,\phi_{1},\phi_{2}$ are relaxed to any pseudo-Lipschitz functions of finite orders, the exponential rates involving them are lost and become linear following Lemma \ref{conc-Mor}. 
\subsection{Relaxing the deterministic teacher assumption}
\label{relax-teacher}
The entirety of the previous proof has been done with a deterministic vector $\vec{\theta}_{0}$. Now, if $\vec{\theta}_{0}$ is assumed to be a random vector independent of all other quantities, as prescribed in the set of assumptions \ref{main-assumptions}, we can "freeze" the variable $\vec{\theta}_{0}$ by conditioning on it. The whole proof can then be understood as studying the value of the cost conditioned on the value of $\vec{\theta}_{0}$. Note that, in the Gaussian case, correlations between the teacher and student are expressed through the covariance matrices, thus leaving the possibility to parametrise the teacher with a vector $\vec{\theta}_{0}$ indeed independent of all the rest. To lift the conditioning in the end, one only needs to average out on the distribution of $\vec{\theta}_{0}$, the summability conditions of which are prescribed in the set of assumptions \ref{main-assumptions}. Thus, random teacher vectors can be treated simply by taking an additional expectation in the expressions of Theorem \ref{main-th}, provided $\vec{\theta}_{0}$ is independent of the matrices $A,B$ and the randomness in $f_{0}$. \\
\quad \\
As mentioned at the end of the previous section, the finite size rates will be determined by the assumptions made on the teacher vector and decay of the eigenvalues of the covariance matrices. We do not investigate in detail the limiting assumptions under which exponential rates still hold regarding the randomness of the teacher or tails of the eigenvalue distributions of covariance matrices.
\subsection{The 'vanilla' teacher-student scenario}
In this section, we give the explicit forms of the fixed points equations and optimal asymptotic estimators in the case where the teacher and the student are sampled from the same distribution, i.e. $\Omega=\Phi=\Psi=\Sigma$ where $\Sigma$ is a positive definite matrix with sub-Gaussian eigenvalue decay. This setup was rigorously studied in \cite{celentano2020lasso} for the LASSO and heuristically in \cite{huang2020large} for the ridge regularized logistic regression. In this case, the fixed point equations become
\begin{align}
   \tau_{1} & =\frac{1}{\sdim}\mathbb{E}\left[\mathbf{g}^{\top}\mbox{prox}_{\frac{\eta}{\tau_{2}}f(\Sigma^{-1/2}.)}\left(\frac{\eta}{\tau_{2}}\left(\nu\Sigma^{1/2}\vec{\theta}_{0}+\kappa\mathbf{g}\right)\right)\right] \\
    m\sqrt{\gamma}&=\frac{1}{\sdim}\mathbb{E}\left[\mathbf{v}^{\top}\Sigma^{-1/2}\mbox{prox}_{\frac{\eta}{\tau_{2}}f(\Sigma^{-1/2}.)}\left(\frac{\eta}{\tau_{2}}\left(\nu\Sigma^{1/2}\vec{\theta}_{0}+\kappa\mathbf{g}\right)\right)\right] \\
     \tau_{2}&=\alpha\frac{\kappa}{\tau_{1}}\eta-\frac{\kappa\alpha}{\tau_{1}n}\mathbb{E}\left[\mathbf{h}^{\top}\mbox{prox}_{\frac{\tau_{1}}{\kappa}\loss(.,\mathbf{y})}\left(\frac{m}{\sqrt{\rho}}\mathbf{s}+\eta\mathbf{h}\right)\right] \\
    \eta^{2}+\frac{m^{2}}{\rho}&=\frac{1}{\sdim}\mathbb{E}\left[\norm{\mbox{prox}_{\frac{\eta}{\tau_{2}}f(\Sigma^{-1/2}.)}\left(\frac{\eta}{\tau_{2}}\left(\nu\Sigma^{1/2}\vec{\theta}_{0}+\kappa\mathbf{g}\right)\right)}_{2}^{2}\right]\\
    \nu\sqrt{\gamma}&=\alpha\frac{\kappa}{n\tau_{1}}\mathbb{E}\left[(\frac{m}{\eta\rho}\mathbf{h}-\frac{\mathbf{s}}{\sqrt{\rho}})^{\top}\mbox{prox}_{\frac{\tau_{1}}{\kappa}\loss(.,\mathbf{y})}\left(\frac{m}{\sqrt{\rho}}\mathbf{s}+\eta\mathbf{h}\right)\right] \\
    \tau_{1}^{2}&=\frac{\alpha}{n}\mathbb{E}\left[\norm{\frac{m}{\sqrt{\rho}}\mathbf{s}+\eta \mathbf{h}-\mbox{prox}_{\frac{\tau_{1}}{\kappa}g(.,y)}\left(\frac{m}{\sqrt{\rho}}\mathbf{s}+\eta\mathbf{h}\right)}_{2}^{2}\right]
\end{align}
\noindent and the asymptotic optimal estimators read:
\begin{align}
    \mathbf{w}^{*} = \Sigma^{-1/2}\mbox{\rm prox}_{\frac{\eta^{*}}{\tau_{2}^{*}}f(\Sigma^{-1/2}.)}(\frac{\eta^{*}}{\tau_{2}^{*}}(\nu^{*}\Sigma^{1/2}\vec{\theta}_{0}+\kappa^{*}\mathbf{g})), &&\mathbf{z}^{*} = \mbox{\rm prox}_{\frac{\tau_{1}^{*}}{\kappa^{*}}\loss(.,\mathbf{y})}\left(\frac{m^{*}}{\sqrt{\rho}}\mathbf{s}+\eta^{*}\mathbf{h}\right)
\end{align}

%% file: sections/appendix/l2_matching.tex
In this Appendix, we show that the rigorous result of Theorem \ref{main-th} can be used to prove the replica prediction in the case of a separable loss, a ridge penalty. For simplicity, we restrict ourselves to the case of random teacher weights with $\vec{\theta}_{0}\sim\mathcal{N}(0,\mat{I}_{\tdim})$. We provide an exact analytical matching between the replica prediction and the one obtained with Gordon's theorem. We start by an explicit derivation of the form presented in Corollary \ref{thm:corollary:factorised} from the main result (\ref{free-energy}).
\subsection{Solution for separable loss and ridge regularization}
Replacing $\reg$ with a ridge penalty, we can go back to step (\ref{inter-ridge}) of the main proof and finish the calculation without inverting the matrix $\Omega$. The assumption on the invertibility of $\Omega$ can thus be dropped in the case of $\ell_{2}$ regularization. Letting $G = \left(\frac{\tau_{2}}{\eta}\Omega+\lambda_{2}\mathbf{I}_{\sdim}\right)^{-1}$, we get 
\begin{align}
     \mathcal{E}(\tau_{1},\tau_{2},\kappa,\eta,\nu,m)&=\frac{\kappa\tau_{1}}{2}-\frac{\eta\tau_{2}}{2}+m\nu\sqrt{\gamma}-\frac{\tau_{2}}{2\eta}\frac{m^{2}}{\rho}+\alpha\frac{1}{n}\mathbb{E}\left[\mathcal{M}_{\frac{\tau_{1}}{\kappa}g(.,\mathbf{y})}\left(\frac{m}{\sqrt{\rho}}\mathbf{s}+\eta\mathbf{h}\right)\right] \notag \\
     &-\frac{1}{2d}\nu^{2}\boldsymbol{\theta}_{0}^{\top}\Phi G\Phi^{\top}\boldsymbol{\theta}_{0}
     -\frac{1}{2d}\kappa^{2}\rm{Tr}\left(\Omega^{1/2}G\Omega^{1/2}\right)
\end{align}
using Lemma \ref{conc-Mor} with a separable function, the expectation over the Moreau envelope converges to:
\begin{equation}
    \frac{1}{n}\mathbb{E}\left[\mathcal{M}_{\frac{\tau_{1}}{\kappa}g(.,\mathbf{y})}\left(\frac{m}{\sqrt{\rho}}\mathbf{s}+\eta\mathbf{h}\right)\right] = \mathbb{E}\left[\mathcal{M}_{\frac{\tau_{1}}{\kappa}g(.,y)}\left(\frac{m}{\sqrt{\rho}}s+\eta h\right)\right]
\end{equation}
where $s$ and $h$ are standard normal random variables and $y=f_{0}(\sqrt{\rho}s)$.
The corresponding optimality conditions then reads:
\begin{align}
   &\frac{\partial}{\partial\kappa}: \frac{\tau_{1}}{2}+\frac{1}{2\tau_{1}}\alpha\mathbb{E}\left[\left(\frac{m}{\sqrt{\rho}}s+\eta h-\mbox{prox}_{\frac{\tau_{1}}{\kappa}g(.,y)}\left(\frac{m}{\sqrt{\rho}}s+\eta h\right)\right)^{2}\right]-\kappa \frac{1}{d}\rm{Tr}\left(\Omega^{1/2}G\Omega^{1/2}\right) = 0 \\
   & \frac{\partial}{\partial \nu}: m\sqrt{\gamma}-\frac{1}{d}\nu\boldsymbol{\theta}_{0}\Phi^{\top}G\Phi^{\top}\boldsymbol{\theta}_{0} = 0 \\
   & \frac{\partial}{\partial \tau_{2}}: -\frac{\eta}{2}-\frac{m^{2}}{2\rho\eta}+\frac{1}{2}\frac{\nu^{2}}{\eta}\left(\Omega^{1/2}\Phi^{\top}\boldsymbol{\theta}_{0}\right)^{\top}G^{2}\Omega^{1/2}\Phi^{\top}\boldsymbol{\theta}_{0}+\frac{\kappa^{2}}{2\eta}Tr\left(G^{2}\Omega^{2}\right) = 0 \\
   & \frac{\partial}{\partial m}: \nu\sqrt{\gamma}-\frac{\tau_{2}}{\rho\eta}m+\alpha\mathbb{E}\left[\frac{\kappa}{\tau_{1}}\frac{s}{\sqrt{\rho}}(\frac{m}{\sqrt{\rho}}s+\eta h-\mbox{prox}_{\frac{\tau_{1}}{\kappa}g(.,y)}\left(\frac{m}{\sqrt{\rho}}s+\eta h\right))\right] = 0 \\
   & \frac{\partial}{\partial \eta} : -\frac{\tau_{2}}{2}+\frac{\tau_{2}m^{2}}{2\rho\eta^{2}}+\alpha\mathbb{E}\left[\frac{\kappa}{\tau_{1}}h\left(\frac{m}{\sqrt{\rho}}s+\eta h-\mbox{prox}_{\frac{\tau_{1}}{\kappa}g(.,y)}\left(\frac{m}{\sqrt{\rho}}s+\eta h\right)\right)\right]\notag\\
   &-\frac{1}{2}\frac{\tau_{2}\nu^{2}}{\eta^{2}}\left(\Omega^{1/2}\Phi^{\top}\boldsymbol{\theta}_{0}\right)^{\top}G^{2}\Omega^{1/2}\Phi^{\top}\boldsymbol{\theta}_{0}-\frac{\tau_{2}\kappa^{2}}{2\eta^{2}}\rm{Tr}(G^{2}\Omega^{2}) = 0 \\
  & \frac{\partial}{\partial \tau_{1}}: \frac{\kappa}{2}-\frac{\kappa}{2\tau_{1}^{2}}\alpha\mathbb{E}\left[\left(\frac{m}{\sqrt{\rho}}s+\eta h-\mbox{prox}_{\frac{\tau_{1}}{\kappa}g(.,y)}\left(\frac{m}{\sqrt{\rho}}s+\eta h\right)\right)^{2}\right] = 0
\end{align}
simplifying these equations using Stein's lemma, we get:
\begin{align}
   &\frac{\partial}{\partial\kappa}: \frac{\tau_{1}}{\kappa}= \frac{1}{d}\rm{Tr}\left(\Omega^{1/2}\left(\frac{\tau_{2}}{\eta}\Omega+\lambda_{2}\mathbf{I}_{\sdim}\right)^{-1}\Omega^{1/2}\right) \\
   & \frac{\partial}{\partial \nu}: m\sqrt{\gamma}=\frac{1}{d}\nu\boldsymbol{\theta}_{0}\Phi\left(\frac{\tau_{2}}{\eta}\Omega+\lambda_{2}\mathbf{I}_{\sdim}\right)^{-1}\Phi^{\top}\boldsymbol{\theta}_{0}\\
   & \frac{\partial}{\partial \tau_{2}}: \eta^{2}+\frac{m^{2}}{\rho} = \frac{1}{d}\nu^{2}\left(\Omega^{1/2}\Phi^{\top}\boldsymbol{\theta}_{0}\right)^{\top}\left(\frac{\tau_{2}}{\eta}\Omega+\lambda_{2}\mathbf{I}_{\sdim}\right)^{-2}\left(\Omega^{1/2}\Phi^{\top}\boldsymbol{\theta}_{0}\right)+\frac{1}{d}\kappa^{2}\rm{Tr}(\left(\frac{\tau_{2}}{\eta}\Omega+\lambda_{2}\mathbf{I}_{\sdim}\right)^{-2}\Omega^{2}) \\
   & \frac{\partial}{\partial m}: \nu\sqrt{\gamma}=\alpha\frac{\kappa}{\sqrt{\rho}\tau_{1}}\left(\mathbb{E}\left[s\mbox{prox}_{\frac{\tau_{1}}{\kappa}g(.,f_{0}(\sqrt{\rho}s))}\left(\frac{m}{\sqrt{\rho}}+\eta h\right)\right]-\frac{m}{\sqrt{\rho}}\mathbb{E}\left[\mbox{prox}^{'}_{\frac{\kappa}{\tau_{1}}g(.,f_{0}(\sqrt{\rho}s))}\left(\frac{m}{\sqrt{\rho}}s+\eta h\right)\right]\right)\\
   & \frac{\partial}{\partial \eta} : \frac{\tau_{2}}{\eta} = \alpha\frac{\kappa}{\tau_{1}}\left(1-\mathbb{E}\left[\mbox{prox}^{'}_{\frac{\tau_{1}}{\kappa}g(.,f_{0}(\sqrt{\rho}s))}\left(\frac{m}{\sqrt{\rho}}s+\eta h\right)\right]\right) \\
  & \frac{\partial}{\partial \tau_{1}}: \kappa^{2} = \left(\frac{\kappa}{\tau_{1}}\right)^{2}\alpha\mathbb{E}\left[\left(\frac{m}{\sqrt{\rho}}s+\eta h-\mbox{prox}_{\frac{\tau_{1}}{\kappa}g(.,f_{0}(\sqrt{\rho}s))}\left(\frac{m}{\sqrt{\rho}}s+\eta h\right)\right)^{2}\right]
\end{align}
\subsection{Matching with Replica equations}
In this section, we show that the fixed point equations obtained from the asymptotic optimality condition of the scalar minimization problem \ref{thm:corollary:factorised} match the ones obtained using the replica method. In what follows we will use the same notations as in \cite{gerace2020generalisation}, and an explicit, clear match with the notations from the proof of the main theorem will be shown. The replica computation, similar to the one from \cite{gerace2020generalisation}, leads to the following fixed point equations, in the replica notations:
\begin{align}
        V&=\frac{1}{p}\mbox{Tr}\left(\lambda \hat{V} I_{p}+\Omega\right)^{-1}\Omega \\
        q&=\frac{1}{p}\mbox{Tr}\left[(\hat{q}\Omega+\hat{m}^{2}\Phi^{\top}\Phi)\Omega\left(\lambda \hat{V} I_{p}+\Omega\right)^{-2}\right] \\
        m&=\frac{1}{\sqrt{\gamma}}\frac{\hat{m}}{p}\mbox{Tr}\left[\Phi^{\top}\Phi\left(\lambda \hat{V} I_{p}+\Omega\right)^{-1}\right] \\
		\hat{V} &= \alpha\mathbb{E}_{\xi}\left[\int_{\mathbb{R}}\dd y~\mathcal{Z}^{0}_{y}\left(y, \frac{m}{\sqrt{q}}, \rho-\frac{m^2}{q}\right) \partial_{\omega}f_{\loss}(y,\sqrt{q}\xi, V)\right]\\
		\hat{q} &=  \alpha\mathbb{E}_{\xi}\left[\int_{\mathbb{R}}\dd y~\mathcal{Z}^{0}_{y}\left(y, \frac{m}{\sqrt{q}}, \rho-\frac{m^2}{q}\right) f_{\loss}(y,\sqrt{q}\xi, V)^2 \right]\\
		\hat{m} &= \frac{\alpha}{\sqrt{\gamma}}\mathbb{E}_{\xi}\left[\int_{\mathbb{R}}\dd y~\partial_{\omega}\mathcal{Z}^{0}_{y}\left(y, \frac{m}{\sqrt{q}}, \rho-\frac{m^2}{q}\right)f_{\loss}(y,\sqrt{q}\xi, V) \right]
\end{align}
\noindent where $f_{g}(y,\omega,V) = -\partial_{\omega}\mathcal{M}_{V\loss(y,\cdot)}(\omega)$ and $\mathcal{Z}_{0}$ is given by:
\begin{align}
    \mathcal{Z}_{0}\left(y,\omega,V\right) = \int\frac{\dd x}{\sqrt{2\pi V}}e^{-\frac{1}{2V}(x-\omega)^2}\delta(y-f^{0}(x)).
\end{align}
In particular we have:
\begin{align}
    \partial_{\omega}\mathcal{Z}_{0}\left(y,\omega,V\right) = \int\frac{\dd x}{\sqrt{2\pi V}}e^{-\frac{1}{2V}(x-\omega)^2}\left(\frac{x-\omega}{V}\right)\delta(y-f^{0}(x))
\end{align}
To be explicit with the notation, let's open the equations up. Take for instance the one for $\hat{m}$. Opening all the integrals:
\begin{align}
    \hat{m} &= \int\frac{\dd\xi}{\sqrt{2 \pi}}e^{-\frac{1}{2}\xi^2}\int\dd y \int\frac{\dd x}{\sqrt{2\pi \left(\rho-m^2/q\right)}}e^{-\frac{1}{2}\frac{\left(x-\frac{m}{\sqrt{q}}\xi\right)^2}{\rho-m^2/q}}\left(\frac{x-\frac{m}{\sqrt{q}}\xi}{\rho-m^2/q}\right)f_{\loss}(y,\sqrt{q}\xi, V)\notag\\
    &\overset{(a)}{=} \int\frac{\dd\xi}{\sqrt{2 \pi}}e^{-\frac{1}{2}\xi^2} \int\frac{\dd x}{\sqrt{2\pi \left(\rho-m^2/q\right)}}e^{-\frac{1}{2}\frac{\left(x-\frac{m}{\sqrt{q}}\xi\right)^2}{\rho-m^2/q}}\left(\frac{x-\frac{m}{\sqrt{q}}\xi}{\rho-m^2/q}\right)f_{\loss}(f_{0}(x),\sqrt{q}\xi, V)\notag\\
\end{align}
\noindent where in $(a)$ we integrated over $y$ explicitly. A direct comparison between the two sets of equations suggests the following mapping to navigate between the replica derivation and the proof using Gaussian comparison theorems. We denote replica quantities with \emph{Rep} indices:
\begin{align}
\label{match-rep-gordon}
    V_{Rep} \iff \frac{\tau_{1}}{\kappa}, &&
    \hat{V}_{Rep} \iff \frac{\tau_{2}}{\eta}, &&
    q_{Rep} \iff \eta^{2}+\frac{m^{2}}{\rho} \notag\\
    \hat{q}_{Rep} \iff \kappa^{2}, &&
    m_{Rep} \iff m, &&
    \hat{m}_{Rep} \iff \nu
\end{align}
with these notations, we get :
\begin{align}
   &\frac{\partial}{\partial\kappa}: V = \frac{1}{d}\rm{Tr}((\hat{V}\Omega+\lambda_{2}\mathbf{I}_{\sdim})^{-1}\Omega) \\
   & \frac{\partial}{\partial \nu}: m=\frac{1}{\sqrt{\gamma}}\frac{\hat{m}}{d}\rm{Tr}((\hat{V}\Omega+\lambda_{2}\mathbf{I}_{\sdim})^{-1}\Phi^{\top}\Phi)\\
   & \frac{\partial}{\partial \tau_{2}}: q = \frac{1}{d}\rm{Tr}((\hat{q}\Omega+\hat{m}^{2}\Phi^{\top}\Phi)\Omega(\hat{V}\Omega+\lambda_{2}\mathbf{I}_{\sdim})^{-2})\\
   & \frac{\partial}{\partial m}: \hat{m}=\frac{\alpha}{\sqrt{\gamma}}\frac{1}{V}\left(\mathbb{E}\left[\frac{s}{\sqrt{\rho}}\mbox{prox}_{Vg(.,f_{0}(\sqrt{\rho }s))}\left(\frac{m}{\sqrt{\rho}}s+\sqrt{q-\frac{m^{2}}{\rho}} h\right)\right]\right.\notag\\
   &\hspace{4cm}\left.-\frac{m}{\rho}\mathbb{E}\left[\mbox{prox}^{'}_{Vg(.,f_{0}(\sqrt{\rho}s))}\left(\frac{m}{\sqrt{\rho}}s+\sqrt{q-\frac{m^{2}}{\rho}}~ h\right)\right]\right)\\
   & \frac{\partial}{\partial \eta} : \hat{V} = \frac{\alpha}{V}\left(1-\mathbb{E}\left[\mbox{prox}^{'}_{Vg(.,f_{0}(\sqrt{\rho}s))}\left(\frac{m}{\sqrt{\rho}}s+\sqrt{q-\frac{m^{2}}{\rho}} h\right)\right]\right) \\
  & \frac{\partial}{\partial \tau_{1}}: \hat{q} = \left(\frac{\alpha}{V^{2}}\right)\mathbb{E}\left[\left(\frac{m}{\sqrt{\rho}}s+\sqrt{q-\frac{m^{2}}{\rho}}~ h-\mbox{prox}_{Vg(.,f_{0}(\sqrt{\rho}s))}\left(\frac{m}{\sqrt{\rho}}s+\sqrt{q-\frac{m^{2}}{\rho}} h\right)\right)^{2}\right]
\end{align}

The first three equations match the replica prediction, the last three can be exactly matched using the following change of variable and Gaussian integration:
\begin{equation}
    \tilde{x} = \frac{x}{\sqrt{\rho}} \quad \tilde{\xi} = \left(\frac{\rho}{\rho-\frac{m^{2}}{q}}\right)^{1/2}\left(\frac{m}{\sqrt{q\rho}}\tilde{x}-\xi\right)
\end{equation}

%% file: sections/appendix/details.tex
In this Appendix we give full details on the numerics used to generate the plots in the main manuscript. An implementation of all the pipelines described below is available at~\url{https://github.com/IdePHICS/GCMProject}.

%%%%%%%%%%%%%%%%%%%%%%%%%%%%%%%%%%%%%%%%%%%
\subsection{Ridge regression on real data}
\label{sec:app:realdata}
%%%%%%%%%%%%%%%%%%%%%%%%%%%%%%%%%%%%%%%%%%%
Consider a real data set $\{\vec{x}^{\mu}, y{^\mu}\}_{\mu=1}^{n_{\tot}}$, where $n_{\tot}$ denote the total number of samples available. In Figs.~\ref{fig:realdata} and \ref{fig:realdata} we work with the MNIST and fashion MNIST data sets for which $n_{\tot} = 6\times 10^{4}$ and $\ddim = 28\times 28 = 764$. In both cases, we center the data and normalise by dividing it by the global standard deviation. We work with binary labels $y^{\mu} \in\{-1, 1\}$, with $y^{\mu}=1$ for even digits (MNIST) or clothes above the waist (fashion MNIST) and $y^{\mu}=-1$ for odd digitis (MNIST) or clothes below the waist (fashion MNIST). In a ridge regression task, we assume $y^{\mu} = \vec{\theta}_{0}^{\top}\vec{u}^{\mu}$ for a teacher feature map $\vec{u}^{\mu}= \vec{\varphi}_{t}(\vec{x}^{\mu})$ and we are interested in studying the performance of the estimator $\hat{y} = \vec{v}^{\top}\hat{\vec{w}}$ where $\vec{v}=\vec{\varphi}_{s}(\vec{x})$ obtained by solving the empirical risk minimisation problem in eq.~\eqref{eq:app:argmin} with the squared loss $\loss(x,y) = \frac{1}{2}(y-x)^2$ and $\ell_2$ regularisation $\lambda>0$.

\paragraph{Simulations:} First, we discuss in detail how we conducted the numerical simulations in Figs.~\ref{fig:realdata} and \ref{fig:realdata} in the main manuscript.

In Fig.~\ref{fig:realdata}, the student feature maps $\vec{\varphi}_{s}$ is taken to be different transforms used in the literature. For the scattering transform, we have used the out-of-the-box python package \pyth{Kymatio} \cite{andreux2020kymatio} with hyperparameters $J=3$ and $L=8$, which defines a feature map $\vec{\varphi}_{s}:\mathbb{R}^{28\times 28}\to\mathbb{R}^{217\times 3\times 3}$, and thus $d=1953$. For the random features, a random matrix $\mat{F}\in\mathbb{R}^{\sdim \times 784}$ with i.i.d. $\mathcal{N}(0,1/784)$ entries is generated and fixed. Note that the number of features $d=1953$ is chosen to match the ones for the scattering transform. The random feature map is then applied to the flattened MNIST image as $\vec{\varphi}_{s}(\vec{x}) = \rm{erf}\left(\mat{F}\vec{x}\right)$. Finally, we have chosen a kernel corresponding to the limit of this random feature map \cite{williams96}:
\begin{align}
    K(\vec{x}_{1},\vec{x}_{2}) = \frac{2}{\pi}\sin^{-1}\left(\frac{2\vec{x}_{1}^{\top}\vec{x}_{2}}{\sqrt{\left(1/d+2||\vec{x}_{1}||^2_{2}\right)\left(1/d+2||\vec{x}_{2}||^2_{2}\right)}}\right).
    \label{eq:app:erfkernel}
\end{align}
In Fig.~\ref{fig:realdata}, the feature $\vec{\varphi}_{s}^{t}$ is taken from a learned neural network at different epochs $t\in\{0,5,50,200\}$ of training. For this experiment, we chose the following architecture implemented in \pyth{Pytorch}:
%\begin{minted}[
%framesep=2mm,
%baselinestretch=1.2,
%fontsize=\footnotesize,
%]{python}
%Sequential(
%  (0): Linear(in_features=784, out_features=2352, bias=False)
%  (1): ReLU()
%  (2): Linear(in_features=2352, out_features=2352, bias=False)
%  (3): ReLU()
%  (4): Linear(in_features=2352, out_features=1, bias=False)
%)
%\end{minted}
\begin{python}
Sequential(
  (0): Linear(in_features=784, out_features=2352, bias=False)
  (1): ReLU()
  (2): Linear(in_features=2352, out_features=2352, bias=False)
  (3): ReLU()
  (4): Linear(in_features=2352, out_features=1, bias=False)
)
\end{python}

The first two layers of the network therefore defines a feature map $\vec{\varphi}_{s}:\mathbb{R}^{784}\to\mathbb{R}^{2352}$ acting on flattened fashion MNIST images. The network was initialized using the \pyth{pyTorch}'s default Kaiming initialisation \cite{kaiminginit} and was trained on the full data set ($n_{\tot}$ samples) with Adam \cite{Adam} optimiser (learning rate $10^{-3}$) on the MSE loss for a total of $500$ epochs. Snapshots were taken at epochs $t\in\{0,5,50,200\}$, defining the feature maps $\vec{\varphi}_{s}^t(\cdot)$ at each of these epochs. 

In both experiments, we ran ridge regression at fixed regularisation $\lambda>0$ by sub-sampling $\samples$ samples from the data set $\mathcal{D} = \{\vec{v}^{\mu}, y^{\mu}\}_{\mu=1}^{n_{\tot}}$, $\vec{v}^{\mu} = \vec{\varphi}_{s}\left(\vec{x}^{\mu}\right)$, with the estimator given by the closed-form expression:
\begin{align}
    \hat{\vec{w}} = 
    \begin{cases}
    \left(\lambda\mat{I}_{\sdim}+\mat{V}^{\top}\mat{V}\right)^{-1}\mat{V}^{\top}\vec{y}, & \text{ if } \samples \geq \sdim \\
    \mat{V}^{\top}\left(\lambda\mat{I}_{\samples}+\mat{V}\mat{V}^{\top}\right)^{-1}\vec{y}, & \text{ if } \samples < \sdim
    \end{cases}
\end{align}
\noindent where $\mat{V}\in\mathbb{R}^{\samples \times \sdim}$ is the normalised matrix obtained by concatenating $\{\vec{v}^{\mu}/\sqrt{\sdim}\}_{\mu=1}^{\samples}$. A similar closed-form expression in terms of the Gram matrix was used in the kernel case. The averaged training and test errors were computed over $10$ independent draws sub-samples of $\mathcal{D}$. To reduce the effect spurious correlations due to the sampling of a finite universe $\mathcal{D}$, we have always evaluated the test error on the whole universe $\mathcal{D}$. The code for these two experiments is available in~\url{https://github.com/IdePHICS/GCMProject}.

\begin{figure}[t]
  \begin{center}
  \centerline{\includegraphics[width=0.5\columnwidth]{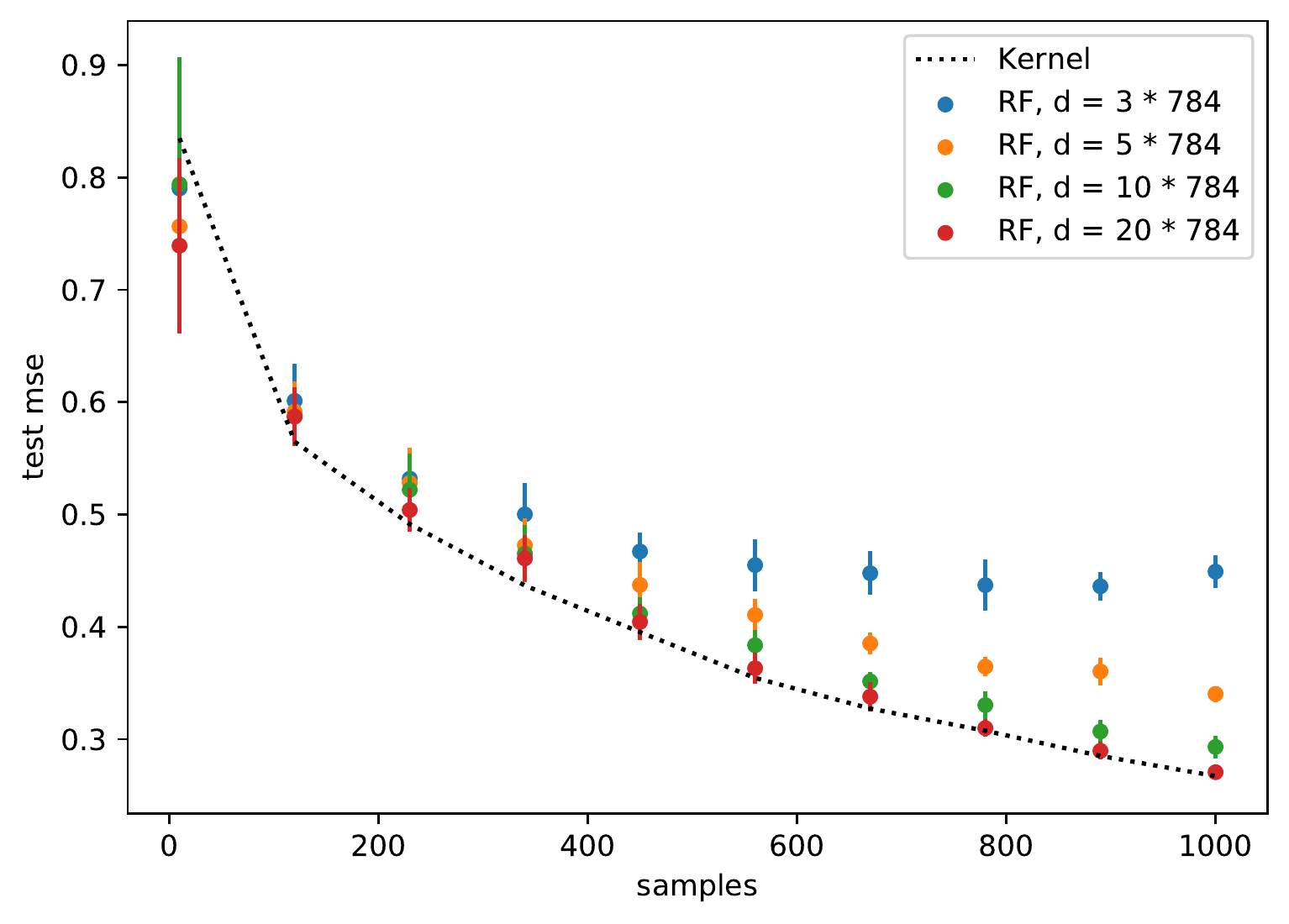}}
  \caption{Test error as a function of the number of samples for kernel ridge regression task on MNIST odd vs. even data, with $\lambda=10^{-1}$. The different curves compare the performance of a random features approximation $\vec{\varphi}_{s}(\vec{x}) = \rm{erf}\left(\mat{F}\vec{x}\right)$ with the performance of the limiting kernel eq.~\eqref{eq:app:erfkernel}. Different curves correspond to different aspect ratios of the Gaussian projection matrix $\mat{F}\in\mathbb{R}^{\sdim\times 784}$.
  }
\label{fig:app:kernelvsrf}
\end{center}
\end{figure}

\paragraph{Self-consistent equations: } For the theoretical curves, we need to provide the population covariances $(\Omega, \Phi, \Psi)$ and the teacher weights $\vec{\theta}_{0}\in\mathbb{R}^{\tdim}$ corresponding to the task of interest. Since when dealing with real data we have a limited number of samples $n_{\tot}$ at our disposal, we estimate the population covariances by the empirical covariances on the whole universe:
\begin{align}
    \Psi = \frac{1}{n_{\tot}}\sum\limits_{\mu=1}^{n_{\tot}}\vec{u}^{\mu}{\vec{u}^{\mu}}^{\top}, && \Phi = \frac{1}{n_{\tot}}\sum\limits_{\mu=1}^{n_{\tot}}\vec{u}^{\mu}{\vec{v}^{\mu}}^{\top}, && \Omega = \frac{1}{n_{\tot}}\sum\limits_{\mu=1}^{n_{\tot}}\vec{v}^{\mu}{\vec{v}^{\mu}}^{\top}.
    \label{eq:app:cov:real}
\end{align}
In principle, the teacher weights need to be estimated by inverting $\vec{y} = \mat{U}\vec{\theta}_{0}$. However, as explained in $n_{\tot}$ in Sec.~\ref{sec:regression:real}, one can avoid doing so by noting the teacher weights only appear in the self-consistent equations \ref{eq:main:sp} through $\rho = \frac{1}{k}\vec{\theta}_{0}^{\top}\Psi\vec{\theta}_{0}$ and $\Phi^{\top}\vec{\theta}_{0}$. Therefore, \emph{all teacher vector $\vec{\theta}_{0}$ and feature map $\varphi_{s}$ that linearly interpolate the data set $\{\vec{x}^{\mu}, y^{\mu}\}_{\mu=1}^{n_{\tot}}$ are equivalent}, since we can write:
\begin{align}
    \rho = \frac{1}{n_{\tot}}\sum\limits_{\mu=1}^{n_{\tot}}\left(y^{\mu}\right)^2, && \Phi^{\top}\vec{\theta}_{0} = \frac{1}{n_{\tot}} \sum\limits_{\mu=1}^{n_{\tot}}\vec{v}^{\mu}y^{\mu}.
    \label{eq:app:teacher:real}
\end{align}
\noindent which is independent from $\left(\vec{\varphi}_{t}, \vec{\theta}_{0}\right)$. In particular, note that for our binary labels $y^{\mu}\in\{+1, -1\}$, we have $\rho = 1$. In both Fig.~\ref{fig:realdata} and \ref{fig:realdata} of the main, we estimated the covariance $\Omega$ as in eq.~\eqref{eq:app:cov:real} by applying the feature maps $\vec{\varphi}_{s}$ described above to the whole data set, took $\rho = 1$ (since in both we have binary labels) and used eq.~\eqref{eq:app:teacher:real} to estimate $\Phi^{\top}\vec{\theta}_{0}$. This was then fed to our iterator package (\url{https://github.com/IdePHICS/GCMProject}) to compute the curves. For the kernel curve, we used the random features approximation of eq.~\eqref{eq:app:erfkernel} with a $\sdim = 20\times 1953$ dimensional feature space to estimate the covariance $\Omega$. We have checked that this indeed provide a good approximation of $K$ for the sample range considered, see Fig.\ref{fig:app:kernelvsrf}.

\begin{figure}[t]
  \begin{center}
    \begin{subfigure}[t]{0.45\textwidth}                
    \includegraphics[width=\textwidth]{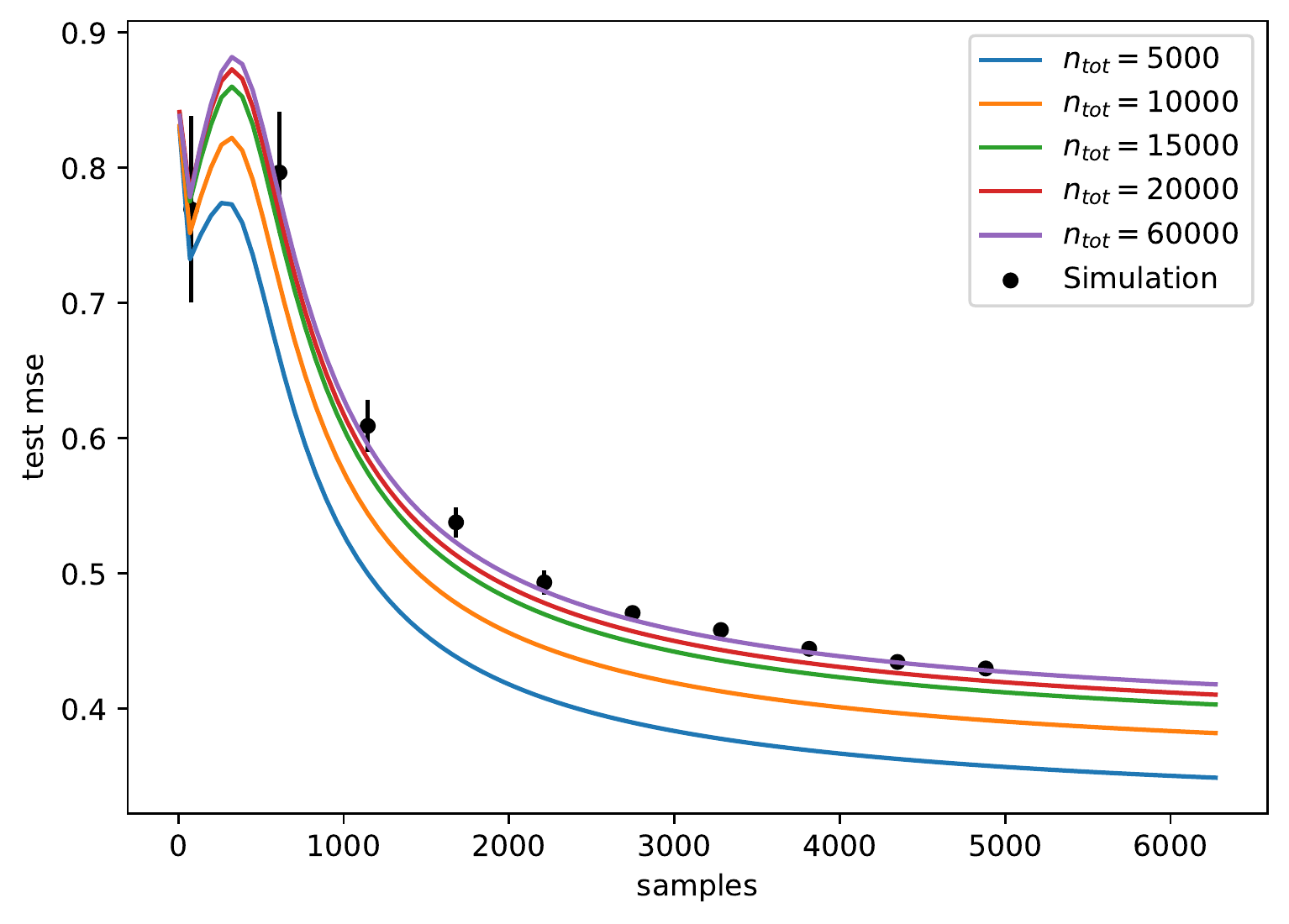}
  \end{subfigure}\quad
\begin{subfigure}[t]{0.45\textwidth}                
\includegraphics[width=\textwidth]{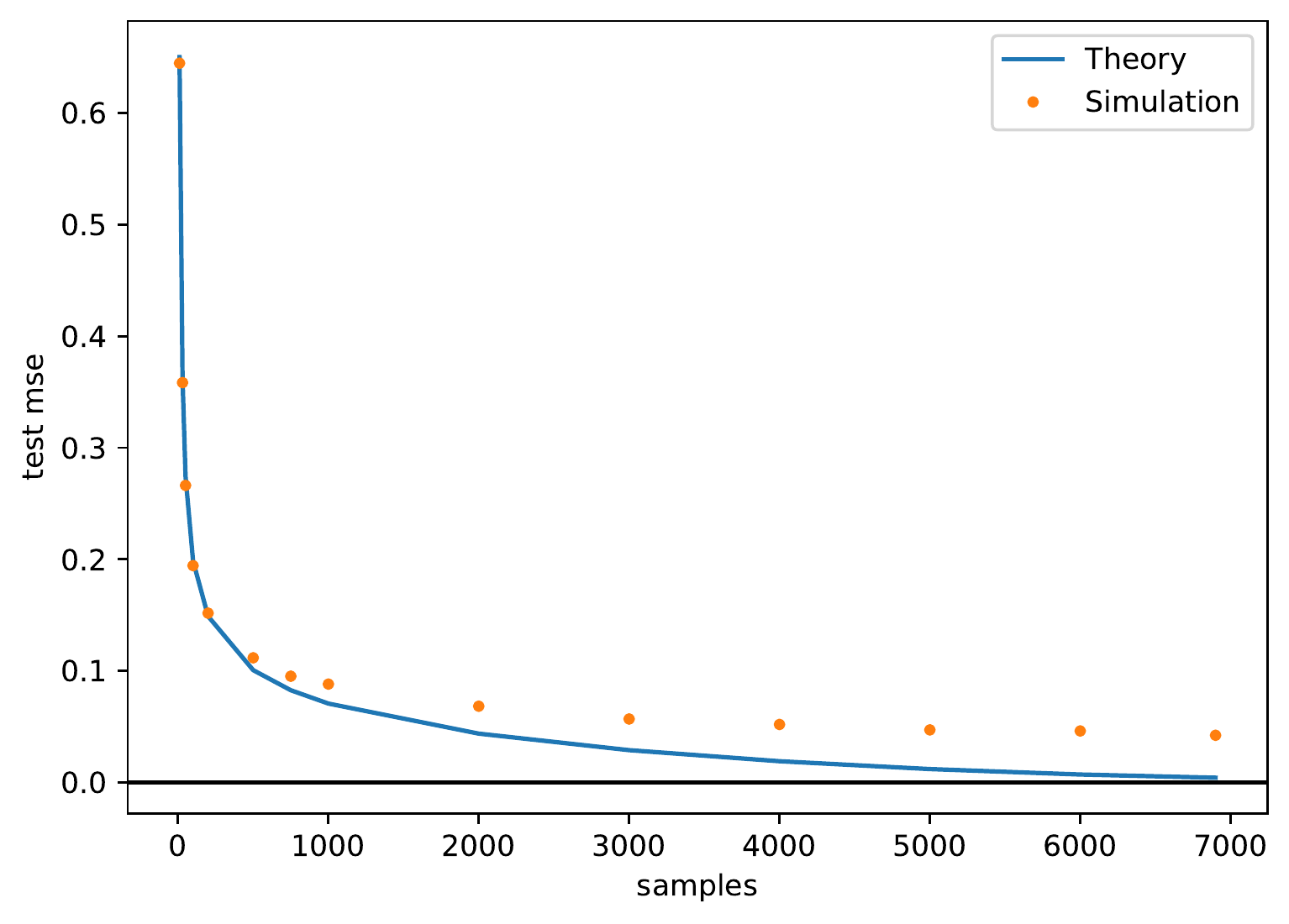}
  \end{subfigure}
  \caption{(Left) Test mse for ridge regression on MNIST odd vs. even task and $\lambda=0.01$. Different curves show the theoretical prediction when the population covariances are estimated using a smaller number of samples $n_{\tot}$ in the universe. (Right) Test mse for NTK kernel regression on MNIST 8 vs. 9 task with $\lambda=0.01$. Note that for this task we have $n_{\tot} = 7000$, and while the theoretical result predicts perfect generalisation as the number of samples approach $n_{\tot}$, the true test error goes to a constant.}
\label{fig:app:ntot}
\end{center}
\end{figure}

\paragraph{Limitations: } As we have discussed above, a key ingredient of our theoretical analysis is the estimation of the population covariances. For real data, this relies on the empirical covariance of the whole data set with $n_{\tot}$ samples. We expect this approximation to be good only for $n\ll n_{\tot}$ samples, as it is the case for the ranges plotted in Figs.~\ref{fig:realdata} and \ref{fig:realdata}. Indeed, as $n\approx n_{\tot}$ we start observing deviations between the theoretical prediction and the simulations. In Fig.~\ref{fig:app:ntot} (right) we show an example of a NTK kernel regression task on 8 vs 9 MNIST digit classification, for which $n_{\tot} = 7000$. Note that while the theoretical prediction reach perfect generalisation at $n\approx n_{\tot}$, the simulated error approaches a plateau. Alternatively, instead of varying the sample range, in Fig.~\ref{fig:app:ntot} (left) we show how the matching betweem theory and simulation degrades by varying $n_{\tot}$ on a fixed sample range for a MNIST odd vs. even task.

\begin{figure}[h]
  \begin{center}
  \includegraphics[width=0.5\textwidth]{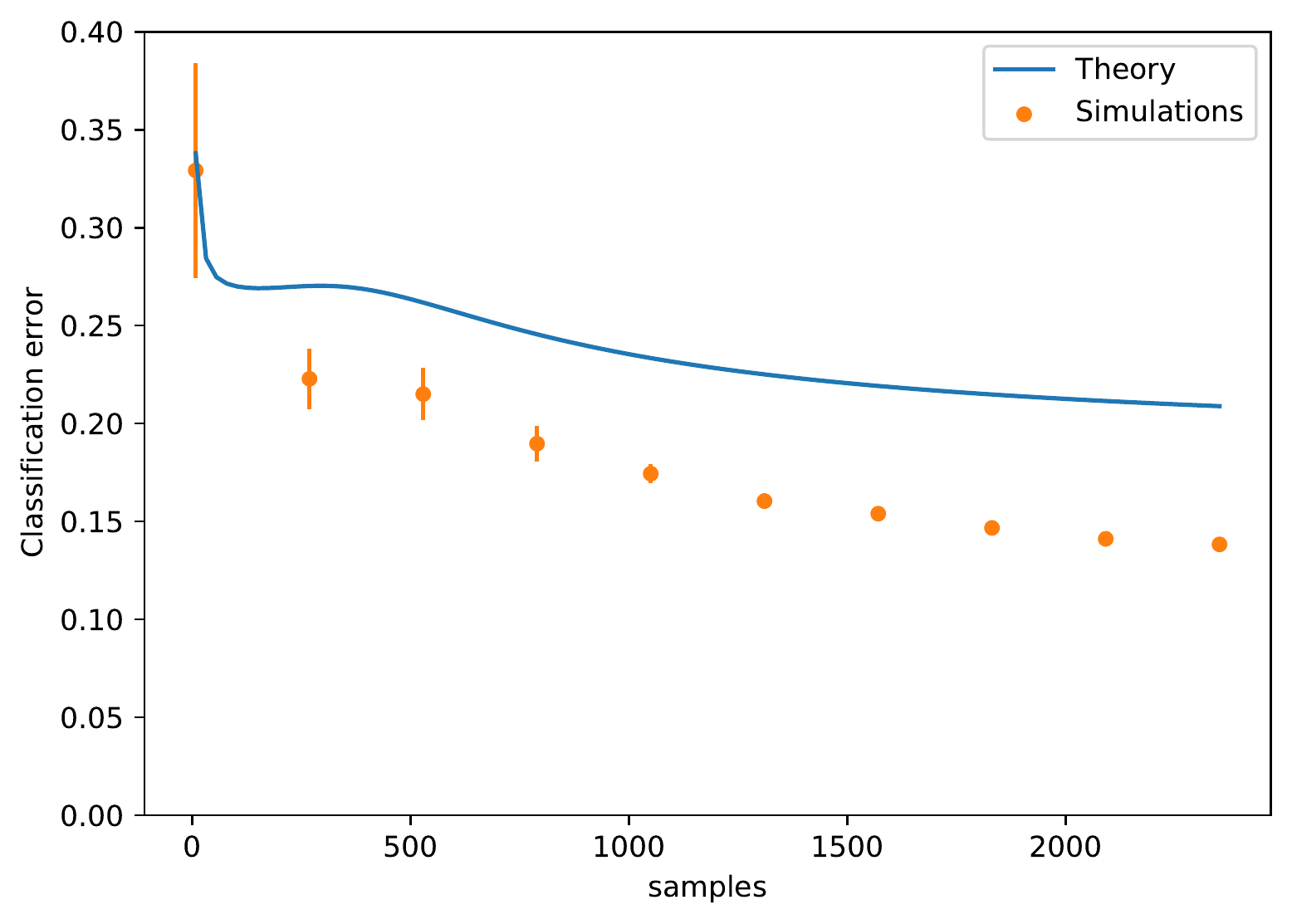}
\caption{Classification error for binary classification task with the square loss on MNIST odd vs. even task and $\lambda=0.01$}
\label{fig:app:l2class}
\end{center}
\end{figure}

As it was discussed in Sec.~\ref{sec:regression:real} of the main manuscript, the universality argument sketched above is only valid in the case of a linear student. For instance, applying the same construction to a binary classification task with $f_{0}(x) = \hat{f}(x) = \sign(x)$ lead to a mismatch between theory and experiments, as exemplified in Fig.~\ref{fig:logistic:break} of the main for a logistic regression task on CIFAR10 gray-scale images. Interestingly, this is even the case for binary classification with the square loss $\loss(x,y) = \frac{1}{2}(x-y)^2$, in which the estimator $\hat{\vec{w}}$ is the same as for ridge regression. In other words, by simply changing the predictor $\hat{f}(x)=\sign(x)$, we have a breakdown of universality, as shown in Fig.~\ref{fig:app:l2class}.

%%%%%%%%%%%%%%%%%%%%%%%%%%%%%%%%%%%%%%%%%%%%%%%%%%%%%%%%
\subsection{Binary classification on GAN generated data}
\label{sec:app:gan}
%%%%%%%%%%%%%%%%%%%%%%%%%%%%%%%%%%%%%%%%%%%%%%%%%%%%%%%%%
For our purposes, a generative adversarial network (GAN) is a pre-trained neural network defining a map $\mathcal{G}$ taking a Gaussian i.i.d. vector $\vec{z}\sim\mathcal{N}(\vec{0}, \mat{I})$ (a.k.a. the latent representation) into a realistic looking input image $\vec{x}\in\mathbb{R}^{\ddim}$. In both Figs.~\ref{fig:logistic:epochs} and \ref{fig:logistic:break}, we have used a deep convolutional GAN (dcGAN) \cite{radford2015unsupervised} with the following architecture
%\begin{minted}[
%framesep=2mm,
%baselinestretch=1.2,
%fontsize=\footnotesize,
%]{python}
%Generator(
%  (main): Sequential(
%    (0): ConvTranspose2d(100, 512, kernel_size=(4, 4), stride=(1, 1), bias=False)
%    (1): BatchNorm2d(512, eps=1e-05, momentum=0.1, affine=True, track_running_stats=True)
%    (2): ReLU(inplace=True)
%    (3): ConvTranspose2d(512, 256, kernel_size=(4, 4), stride=(2, 2), padding=(1, 1), bias=False)
%    (4): BatchNorm2d(256, eps=1e-05, momentum=0.1, affine=True, track_running_stats=True)
%    (5): ReLU(inplace=True)
%    (6): ConvTranspose2d(256, 128, kernel_size=(4, 4), stride=(2, 2), padding=(1, 1), bias=False)
%    (7): BatchNorm2d(128, eps=1e-05, momentum=0.1, affine=True, track_running_stats=True)
%    (8): ReLU(inplace=True)
%    (9): ConvTranspose2d(128, 64, kernel_size=(4, 4), stride=(2, 2), padding=(1, 1), bias=False)
%    (10): BatchNorm2d(64, eps=1e-05, momentum=0.1, affine=True, track_running_stats=True)
%    (11): ReLU(inplace=True)
%    (12): ConvTranspose2d(64, 3, kernel_size=(1, 1), stride=(1, 1), bias=False)
%    (13): Tanh()
%  )
%)
%\end{minted}

\noindent and which has been trained on the full CIFAR10 data set. It therefore takes a $100$-dimensional latent vector and returns a $\ddim = 32\times 32\times 3 = 3072$ CIFAR10-looking image. The GAN was trained on the original CIFAR10 data set without data augmentation for 50 epochs. Both the discriminator and the generator were trained using Adam, with Adam parameters $\beta_1=0.5$ and $\beta_2=0.999$. In practice, the advantage of working with a GAN is that we have a generative process to sample as many independent data points as we need, both for the simulations and for the estimation of the population covariances.

\begin{figure}[t]
  \begin{center}
  \includegraphics[width=0.7\textwidth]{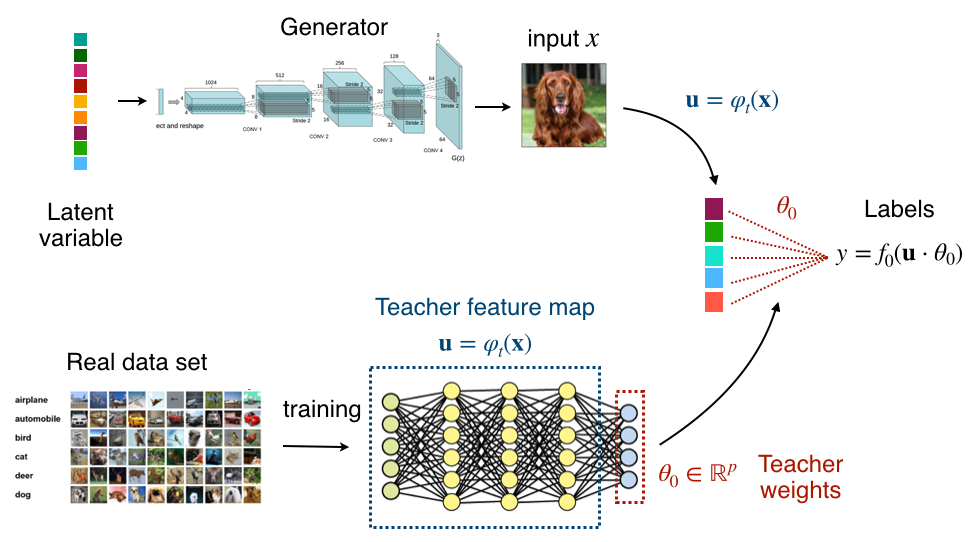}
\caption{Illustration of the pipeline to generate synthetic realistic data. A dcGAN is first trained to generate CIFAR10-looking images from i.i.d. Gaussian noise. Then, a teacher trained to classify real CIFAR10 images is used to assign labels to the dcGAN generated images.}
\label{fig:app:setupgan}
\end{center}
\end{figure}

\paragraph{Learning the teacher: } As discussed in Sec.~\ref{sec:main:gan} of the main manuscript, to label the GAN generated CIFAR10-looking images we learn a teacher feature map $\vec{\varphi}_{t}$ and weights $\vec{\theta}_{0}\in\mathbb{R}^{\tdim}$. For the experiments shown in Figs.~\ref{fig:logistic:epochs}, we have trained with a fully-connected neural network on the full CIFAR10 data set with the following squared architecture:
%\begin{minted}[
%framesep=2mm,
%baselinestretch=1.2,
%fontsize=\footnotesize,
%]{python}
%MLP(
%   (preprocess1): Linear(in_features=3072, out_features=3072, bias=False)
%   (g1): ReLU(inplace=True)
%   (preprocess2): Linear(in_features=3072, out_features=3072, bias=False)
%   (g2): ReLU(inplace=True)
%   (preprocess3): Linear(in_features=3072, out_features=3072, bias=False)
%   (g3): ReLU(inplace=True)
%   (bnz): BatchNorm1d(3072, eps=1e-05, momentum=0.1, affine=False, track_running_stats=False)
%   (fc): Linear(in_features=3072, out_features=1, bias=False)
%)
%\end{minted}
\begin{python}
Generator(
  (main): Sequential(
    (0): ConvTranspose2d(100, 512, kernel_size=(4, 4), stride=(1, 1), bias=False)
    (1): BatchNorm2d(512, eps=1e-05, momentum=0.1, affine=True, track_running_stats=True)
    (2): ReLU(inplace=True)
    (3): ConvTranspose2d(512, 256, kernel_size=(4, 4), stride=(2, 2), padding=(1, 1), bias=False)
    (4): BatchNorm2d(256, eps=1e-05, momentum=0.1, affine=True, track_running_stats=True)
    (5): ReLU(inplace=True)
    (6): ConvTranspose2d(256, 128, kernel_size=(4, 4), stride=(2, 2), padding=(1, 1), bias=False)
    (7): BatchNorm2d(128, eps=1e-05, momentum=0.1, affine=True, track_running_stats=True)
    (8): ReLU(inplace=True)
    (9): ConvTranspose2d(128, 64, kernel_size=(4, 4), stride=(2, 2), padding=(1, 1), bias=False)
    (10): BatchNorm2d(64, eps=1e-05, momentum=0.1, affine=True, track_running_stats=True)
    (11): ReLU(inplace=True)
    (12): ConvTranspose2d(64, 3, kernel_size=(1, 1), stride=(1, 1), bias=False)
    (13): Tanh()
  )
)
\end{python}
The teacher feature map $\vec{\varphi}_{t}:\mathbb{R}^{\ddim}\to\mathbb{R}^{\tdim}$ was then taken to be the first 2-layers, and the teacher weights $\vec{\theta}_{0}$ the weights of the last layer, where $\ddim = \tdim =32 \times 32 \times 3 = 3072$. We used the same architecture for the experiment in Fig.~\ref{fig:logistic:break}, but with $\ddim=\tdim=32\times 32 = 1024$ on gray-scale CIFAR10 images. Both teachers were trained on the odd-even discrimination task on CIFAR10 discussed above with the mean-squared error for 50 epochs, starting from \pyth{pyTorch}'s default Kaiming initialisation \cite{kaiminginit} . Optimisation was performed using SGD with momentum 0.9 and weight decay $5\cdot 10^{-4}$. We started with a learning rate of $0.05$, which decayed by a factor 0.1 after 25 and 40 epochs. The resulting trained teacher achieved a $78\%$ classification accuracy on this task. See Fig.~\ref{fig:app:setupgan} for an illustration of this pipeline.

\paragraph{Simulations: } The experiment shown in Fig.~\ref{fig:logistic:epochs} follow a similar pipeline as the one described in Sec.~\ref{sec:app:realdata}. The student feature maps $\vec{\varphi}_{s}^{t}$ are obtained by removing the last layer of a trained a 3-layer student network with architecture:
%\begin{minted}[
%framesep=2mm,
%baselinestretch=1.2,
%fontsize=\footnotesize,
%]{python}
%Sequential(
%  (0): Linear(in_features=1024, out_features=2304, bias=False)
%  (1): ReLU()
%  (2): Linear(in_features=2304, out_features=2304, bias=False)
%  (3): ReLU()
%  (4): Linear(in_features=2304, out_features=1, bias=False)
%)
%\end{minted}
\begin{python}
Sequential(
  (0): Linear(in_features=1024, out_features=2304, bias=False)
  (1): ReLU()
  (2): Linear(in_features=2304, out_features=2304, bias=False)
  (3): ReLU()
  (4): Linear(in_features=2304, out_features=1, bias=False)
)
\end{python}
Training was performed on a data set composed of $n = 30000$ independent samples drawn from the dcGAN described above, with labels $y^{\mu}\in\{+1, -1\}$ assigned by the learned teacher $y^{\mu} = \sign\left(\vec{u}^{\top}\vec{\theta}_{0}\right)$, $\vec{u}^{\mu}=\vec{\varphi}_{t}\left(\vec{x}^{\mu}\right)$. The network was trained for $300$ epochs using Adam optimiser on the MSE loss and \pyth{pyTorch}'s default Kaiming initialisation, and snapshops of the weights were extracted at epochs $t\in\{0, 5, 50, 200\}$. Finally, logistic regression was performed on the learned features $\vec{v}=\vec{\varphi}_{s}(\vec{x})$ on fresh pair of dcGAN generated samples and labels using the out-of-the-box \pyth{LogisticRegression} solver from \pyth{Scikit-learn}. The points and error bars in Fig.~\ref{fig:logistic:epochs} were computed by averaging over $10$ independent runs. The same pipeline was used for Fig.~\ref{fig:logistic:break}, but for $\vec{\varphi}_{s}=\rm{id}$ and on dcGAN generated CIFAR10 gray-scale images. 

\paragraph{Self-consistent equations: } As before, the self-consistent eqs.~\ref{eq:main:sp} require the population covariances $(\Omega, \Phi, \Psi)$ and the teacher weights $\vec{\theta}_{0}$. For synthetic GAN data, the population covariances of the feature maps $\left(\vec{\varphi}_{t}, \vec{\varphi}^{t}_{s}\right)$ used in the simulations can be estimated as well as needed with a Monte Carlo sampling algorithm. For the curves shown in Figs.~\ref{fig:logistic:epochs} and \ref{fig:logistic:break}, the covariances were estimated with $n=10^{6}$ samples with a precision of the order of $10^{-5}$. Together with the teacher weights $\vec{\theta}_{0}$ used to generate the labels, this provides everything needed to compute the theoretical learning curves from the self-consistent equations.

%% file: sections/appendix/Universality.tex
In this Appendix we discuss briefly random matrix theory, and consider heuristic reasons behind the validity of our asymptotic result beyond Gaussian covariates $(\bf{u},\bf{v})$ in the context of ridge regression, with linear teacher. As is well known, the computation of the training and test MSE for ridge regression can be written as a random matrix theory problem. We do not attempt a rigorous approach, but rather to motivate with simple arguments, many of them actually well known, the observed universality and its limits. 

First, let us remind the definition of the model and introduce some simplifications that arise in ridge regression task. We have Gaussian covariates vectors ${\bf u} \in \mathbb{R}^{\tdim}$ and ${\bf v} \in \mathbb{R}^{\sdim}$, with correlations matrices $\Psi,\Omega$ and $\Phi$, from which we draw $n$ independent samples:
\begin{align}
    \begin{bmatrix}
    \mathbf{u} \\
    \mathbf{v}
    \end{bmatrix} \in \mathbb{R}^{p+d} \sim \mathcal{N}\left(0, \begin{bmatrix}\Psi & \Phi \\ \Phi^{\top} & \Omega \end{bmatrix}\right)\,.
\end{align}
We assume the existence of a linear teacher generating the labels ${\bf y} = \mat{U} \vec{\theta}_0$, and recall the student performs ridge regression on the data matrix $\mathcal{V}$.

Note that since ridge regression can be performed in any basis, we might as well work in the basis where the population covariance $\Psi$ of the vector ${\bf u}$ is diagonal. Additionally, we shall use the fact that one can consider a $\vec{\theta}_0$ to be an i.i.d. Rademacher vector, i.e. a random vector of $\pm 1$ {\it without loss of generality}. Indeed, the statistical properties of the random variable ${\bf u} \cdot \vec{\theta}_0$, for a generic $\vec{\theta}_0$, and of the random variable ${\bf \tilde u}\cdot \vec{\theta}$, with $\vec{\theta}$ a Rademacher vector are identical provided a change in the (diagonal) covariance:
\begin{align}
\Psi =  \begin{pmatrix}
\Psi_1 {(\vec{\theta}_0)_1}^2 & 0 & \ldots & 0\\
0& \Psi_2 {(\vec{\theta}_0)_2}^2 & \ldots & 0\\
\vdots & \vdots & \ddots & \vdots\\
0& 0 & \ldots & \Psi_{\tdim} {(\vec{\theta}_0)_{\tdim}}^2 \, .
\end{pmatrix}
\end{align}
The Gaussian model we consider can therefore be rewritten with a Rademacher vector $\vec{\theta}$ provided we change the correlation matrix ${\Psi}$ (as well as the cross-correlation $\Phi$) accordingly. 

We now come back on the problem. Given the vector $\bf y$ and the data $\mathcal{V}\in\mathbb{R}^{\samples\times\sdim}$, the ridge estimator has the following closed-form solution:
\begin{equation} 
 \hat{\mathbf{w}} = \left(\frac{1}{n} \mathcal{V}^{\top}\mathcal{V} + \lambda \mat{I}_{\sdim}\right)^{-1}\mathcal{V}^{\top} {\bf y}= \left(\mat{S}_{v,v} + \lambda \mat{I}_{\sdim}\right) \mat{S}^{\top}_{u,v}\vec{\theta}
\end{equation}
where we have defined the {\it empirical} covariance matrices 
\begin{align} 
 \mat{S}_{u,u} \equiv \frac 1n \mat{U}^{\top}\mat{U}, && \mat{S}_{u,v} \equiv \frac 1n \mat{U}^{\top}\mathcal{V}\, && \mat{S}_{v,v} \equiv \frac 1n \mathcal{V}^{\top}\mathcal{V}.
\end{align}
Given this vector, one can now readily write the expected value of the training  and test losses as follows:
\begin{align}
\mathcal{E}_{\rm{train.}} &= \mathbb{E}_{\mat{U},\mathcal{V},\vec{\theta}}\left[ 
\frac 1n \| \mat{U} \vec{\theta} - \mathcal{V} \hat{\mathbf{w}}  \left(\mat{U},\mathcal{V}\right) \|_2^2
\right]  \notag\\
&=
\mathbb{E}_{\mat{U},\mathcal{V},\vec{\theta}}\left[ \frac 1n \vec{\theta}^{\top} \mat{U}^{\top}\mat{U}\vec{\theta} \right]
+ \mathbb{E}_{\mat{U},\mathcal{V}}\left[ \frac 1n \hat{\mathbf{w}}\left(\mat{U},\mathcal{V}\right)^{\top} \mathcal{V}^{\top} \mathcal{V}  \hat{\mathbf{w}} (\mat{U},\mathcal{V})\right]
- 2 \mathbb{E}_{\mat{U},\mathcal{V}}\left[ \frac 1n \vec{\theta}^{\top} \mat{U}^{\top} \mathcal{V}  \hat{\mathbf{w}} (\mat{U},\mathcal{V})\right]
\notag\\
&= \mathbb{E} \left[{\rm Tr~} \mat{S}_{u,u}  \right] + 
\mathbb{E} \left[
{\rm Tr~}
\mat{S}_{u,v} \left(\mat{S}_{v,v} + \lambda \mat{I}_{\sdim}\right)^{-1} 
 \mat{S}_{v,v} \left(\mat{S}_{v,v} + \lambda \mat{I}_{\sdim}\right)^{-1} \mat{S}_{u,v}^{\top} 
\right]
- 2 \mathbb{E} \left[
{\rm Tr~}
\mat{S}_{u,v} \left(\mat{S}_{v,v} + \lambda \mat{I}_{\sdim}\right)^{-1} \mat{S}_{u,v}^{\top} 
\right]\label{finalexp:tra_l2}
\end{align}
and
\begin{align}
\mathcal{E}_{\rm{gen.}} &= \mathbb{E}_{\mat{U},\mathcal{V},{\bf u,\bf v},\theta}\left[ \frac 1n \| {\bf u}^{\top} \vec{\theta} - {\bf v}^{\top}\hat{\mathbf{w}}(\mat{U},\mathcal{V})   \|_2^2\right]
\notag\\
&=
\mathbb{E}_{\mat{U},\mathcal{V},{\bf u,\bf v},\theta}\left[ \frac 1n \theta^{\top} {\bf u u^{\top}} \vec{\theta} \right]
+ \mathbb{E}_{\mat{U},\mathcal{V}}\left[ \frac 1n \hat{\mathbf{w}}(\mat{U},\mathcal{V})^{\top} \bf{v} \bf{v}^{\top}  \hat{\mathbf{w}} (\mat{U},\mathcal{V})\right]
- 2 \mathbb{E}_{\mat{U},\mathcal{V}}\left[ \frac 1n \vec{\theta}^{\top} {\bf uv^{\top}} \hat{\mathbf{w}} (\mat{U},\mathcal{V})\right]\notag\\
&= {\rm Tr~} \Sigma_{u,u}  + 
\mathbb{E} \left[
{\rm Tr~}
\mat{S}_{u,v} \left(\mat{S}_{v,v} + \lambda \mat{I}_{\sdim}\right)^{-1} 
 \Sigma_{v,v} \left(\mat{S}_{v,v} + \lambda \mat{I}_{\sdim}\right)^{-1} \mat{S}_{u,v} ^{\top} 
\right]
- 2 \mathbb{E} \left[
{\rm Tr~}
\Sigma_{u,v} \left(\mat{S}_{v,v} + \lambda \mat{I}_{\sdim}\right)^{-1} \mat{S}_{u,v} ^{\top} 
\right] \label{finalexp:gen_l2}
\end{align}
where we have denoted the {\it population} correlation matrices $\Psi \equiv \Sigma_{u,u},\Phi \equiv \Sigma_{u,v}, \Omega \equiv \Sigma_{v,v}$ for readability and a direct comparison with their empirical counterpart. The traces appears by the left and right multiplication by the random vector $\vec{\theta}$.

At this point, the entire problem has been mapped to a random matrix theory exercise: assuming data are indeed Gaussian, one can use RMT to compute the six traces that appears in (\ref{finalexp:tra_l2},\ref{finalexp:gen_l2}). Indeed, this is the canonical approach used in most rigorous works for the ridge regression task in the teacher-student framework, instance in ~\cite{dobriban2018high,hastie2019surprises,liao2020random,wu2020optimal}. Remarkably, the replica (and the rigorous Gordon counterpart) allow to find the same result without the explicit use of RMT. 

We now discuss, heuristically, why these results are valid even though the distribution of $[\bf u,v]$ is not actually Gaussian, and in some instances even for real data.  Indeed, that both $\left(\mathcal{E}_{\rm{train.}}, \mathcal{E}_{\rm{gen.}}\right)$ {\it do not depend} explicitly on the distribution of the data, but ---assuming some concentration (or self-averaging)--- only on:
\begin{enumerate}[leftmargin=*]
    \item The spectrum of the population covariances $\Sigma_{u,u},\Sigma_{u,v},\Sigma_{v,v}$.
    \item The spectrum of the empirical covariances $\mat{S}_{u,u},\mat{S}_{u,v},\mat{S}_{v,v}$.
     \item The expectation of the trace of products between empirical and population covariances.
\end{enumerate}
We expect that asymptotically the prediction from the theory will thus be valid for much more generic distributions $[{\bf u,\bf v}] \sim P_{{\bf u,v}}$, provided they share the same population covariances $\Sigma_{u,u},\Sigma_{u,v},\Sigma_{v,v}$ (which we call $\Psi,\Phi, \Omega$). To see this, we need to check how this change in distribution would affect points (1),(2) and (3). Fixing the population covariances, the first bullet point (1) is automatically taken into account. Point (2) and (3) are, however, less  trivial: in order to have universality we need that a) the spectrum of the {\it empirical} covariances of the non-Gaussian distribution to converge the one obtained with the Gaussian one; and b) the trace of products between the {\it empirical} and the {\it population} covariances also to converge to the universal values computed from Gaussians data.

These two last points have been investigated in RMT \cite{anderson2010introduction},  and it is a classical result that such quantities are universal and converge to the Gaussian-predicted values for many distribution, way beyond the Gaussian assumption (in which case the spectral densities are known as the Wigner and Wishart model, or Marcenko-Pastur distribution \cite{marchenko1967distribution}): this powerful universality of RMT is at the origin of the applicability of the model beyond Gaussian data. 
For instance, \cite{ledoit2011eigenvectors} showed that these assumptions are verified for any data generated as $\bf u = \Sigma_{u,u}^{1/2} \vec{\omega}$, assuming the components of the vector $\vec{\omega}$ are drawn i.i.d. from {\it any} distribution (with some assumption on the larger moments). While this is still restrictive, stronger results can be shown, and
\cite{bai2008large,el2009concentration,chafai2018convergence,louart2018concentration} extended them (also loosening the independence assumption) for a very generic class of distributions of correlated random vectors $\bf u$.

Let us give a concrete example. For simplicity, consider the restricted case where ${\bf u} = {\bf v}$, i.e. the teacher acts on the same space as the student. In this case, eqs.~ (\ref{finalexp:tra_l2},\ref{finalexp:gen_l2}) simplify (this is essentially the analysis in \cite{dobriban2018high}) to:
\begin{align}
\mathcal{E}_{\rm{train.}} &= \mathbb{E}_{\mat{U}, \vec{\theta}}\left[ 
\frac 1n \| \mat{U}\vec{\theta} - \mat{U}\hat{\mathbf{w}}(\mat{U}) \|_2^2
\right]  \notag\\
&=
\mathbb{E}_{\mat{U},\vec{\theta}}\left[ \frac 1n \vec{\theta}^{\top} \mat{U}^{\top}\mat{U} \vec{\theta} \right]
+ \mathbb{E}_{\mat{U}}\left[ \frac 1n \hat{\mathbf{w}}(\mat{U})^{\top} \mat{U}^{\top}\mat{U}  \hat{\mathbf{w}} (U)\right]
- 2 \mathbb{E}_{\mat{U}}\left[ \frac 1n \vec{\theta}^{\top}\mat{U}^{\top}\mat{U}  \hat{\mathbf{w}} (\mat{U})\right]
\notag\\
&= \mathbb{E} \left[{\rm Tr~} S  \right] + 
\mathbb{E} \left[
{\rm Tr~}
S \left(\mat{S} + \lambda\mat{I}_{\sdim}\right)^{-1} 
 S \left(\mat{S} + \lambda\mat{I}_{\sdim}\right)^{-1} \mat{S}^{\top} 
\right]
- 2 \mathbb{E} \left[
{\rm Tr~}
S \left(\mat{S} + \lambda\mat{I}_{\sdim}\right)^{-1} \mat{S}^{\top}  \right]\label{finalexp:tra_l2_simple}
\end{align}
and
\begin{align}
\mathcal{E}_{\rm{gen.}} &= \mathbb{E}_{U,{\bf u},\theta}\left[ \frac 1n \| {\bf u}^{\top} \vec{\theta} - {\bf u}^{\top} \hat{\mathbf{w}}(\mat{U})   \|_2^2\right]
\notag\\
&=
\mathbb{E}_{\mat{U},{\bf u},\vec{\theta}}\left[ \frac 1n \vec{\theta}^{\top} {\bf u u^{\top}}\vec{\theta} \right]
+ \mathbb{E}_{\mat{U}}\left[ \frac 1n \hat{\mathbf{w}}(\mat{U})^{\top} {\bf uu^{\top} }  \hat{\mathbf{w}} (U)\right]
- 2 \mathbb{E}_{\mat{U}}\left[ \frac 1n \vec{\theta}^{\top} {\bf uu^{\top} } \hat{\mathbf{w}} (\mat{U})\right]\notag\\
&= {\rm Tr~} \Sigma_{u,u}  + 
\mathbb{E} \left[
{\rm Tr~}
S \left(\mat{S} + \lambda\mat{I}_{\sdim}\right)^{-1} 
 \Sigma \left(\mat{S} + \lambda\mat{I}_{\sdim}\right)^{-1} \mat{S}^{\top} 
\right]
- 2 \mathbb{E} \left[
{\rm Tr~}
\Sigma \left(\mat{S} + \lambda\mat{I}_{\sdim}\right)^{-1} \mat{S}^{\top} 
\right] \label{finalexp:gen_l2_simple}
\end{align}
In the expression of the training loss eq.~\eqref{finalexp:tra_l2_simple}, we see terms such as
\begin{equation}
{\cal A} = \mathbb{E} \left[
{\rm Tr~}
S \left(\mat{S} + \lambda\mat{I}_{\sdim}\right)^{-1} \mat{S}^{\top} \right] \, ,
\end{equation}
depend only on the limiting distribution of eigenvalues of $\mat{S}\in\mathbb{R}^{\sdim\times\sdim}$. This is a very well known problem when the dimension $\sdim$ and the number of samples $\samples$ are send to infinity with fixed ratio $\alpha =\samples/\sdim$, and the limiting spectral density is known as the Marcenko-Pastur law. This is a very robust distribution that is valid way beyond the Gaussian hypothesis \cite{bai2008large,el2009concentration,chafai2018convergence,louart2018concentration}.

In the expression of the generalisation loss eq.~\eqref{finalexp:gen_l2_simple}, however, terms such as
\begin{equation}
{\cal B} = \mathbb{E} \left[
{\rm Tr~}
\Sigma \left(\mat{S} + \lambda\mat{I}_{\sdim}\right)^{-1} \mat{S}^{\top} \right] \, .
\end{equation}
appears. These can be computed using classical RMT results on the concentration of the inverse of the covariance~\cite{hachem2007deterministic,ledoit2011eigenvectors}. The strongest result we are aware of for such problems is from the remarkable work of  \cite{louart2018concentration}. This universality of random matrix theory is thus at the origin of the surprisingly successful application of our Gaussian theory to real data with arbitrary feature maps. Of course the discussion here is limited to the case where ${\bf u}= {\bf v}$ and a concrete mathematical statement would require the generalisation of these arguments to the more generic case of eqs.(\ref{finalexp:tra_l2},\ref{finalexp:gen_l2}), which are closer to the work of \cite{wu2020optimal}. We leave this discussion to future works. 

A similar universality has been discussed for kernel methods in very recent works, but  for the slightly different setting in which data is drawn from a Mixture of Gaussians \cite{liao2020random,seddik2020random} (in which case there is no teacher, the label depends on which Gaussian has been chosen). The universality observed here for ridge regression with linear student, albeit different, is of a similar nature, and it would be interesting to discuss the link between these two approaches.